\documentclass[preprint,12pt,authoryear]{elsarticle}

\usepackage{graphicx}
\usepackage{amsmath}
\usepackage{algorithm}
\usepackage{algorithmic}
\usepackage{array}
\usepackage{amssymb}
\usepackage{multirow}
\usepackage{float}
\usepackage{longtable}
\usepackage{hyperref}
\usepackage{cleveref}
\usepackage{graphicx}
\usepackage{caption}
\usepackage{subcaption}
\usepackage{booktabs}

\usepackage{lineno}

\journal{Journal of Air Transport Management} 

\begin{document}
\begin{frontmatter}

\title{Data-Driven Runway and Taxiway Exits Prediction of Landing Aircraft: A Case Study at Hartsfield-Jackson Atlanta International Airport}

\affiliation[inst1]{organization={Department of Aerospace Engineering and Engineering Mechanics},
            addressline={The University of Texas at Austin}, 
            city={Austin},
            postcode={78712}, 
            state={TX},
            country={USA}}
            
\author[inst1]{Alex Porcayo}
\author[inst1]{Yutian Pang\corref{mycorrespondingauthor}}
\cortext[mycorrespondingauthor]{Corresponding author.}
\ead{yutian.pang@austin.utexas.edu}
\author[inst1]{Maria Thomas}
\author[inst1]{John-Paul Clarke}

\begin{highlights}
\item A two-stage, data-driven decision aid predicts runway exit and taxiway crossing decisions for arriving aircraft at KATL north complex.
\item Nine classifiers are benchmarked and trained on ASDE-X trajectories, aircraft and ramp attributes, traffic rates, and airport weather.
\item Feature importance analysis shows that approach speed dominates exit choice, while departure and crossing rates together with ramp destination drive the crossing-versus-end-around decision.
\item Class overlap analysis via t-SNE and UMAP identifies feature-space inseparability as the primary prediction bottleneck; probability calibration confirms well-calibrated outputs.
\item Precision--recall and confusion-matrix analyses expose class-imbalance effects that motivate imbalance-aware training for deployment.
\item Operational insights identify conditions under which tactical crossings are favored over the end-around, and vice versa.
\end{highlights}

\begin{abstract}
Airport surface operations increasingly limit system-wide performance at high-throughput hubs. We study the airport surface arrival taxi-in problem at Hartsfield–Jackson Atlanta International Airport (KATL) and propose a two-stage, data-driven decision aid that emulates controller workflow: \emph{Stage~I} predicts the runway exit selected by an arriving aircraft; \emph{Stage~II} predicts whether, conditional on that exit, the aircraft will cross the active departure runway at a designated point or take the end-around taxiway. Models are trained on ASDE-X surface trajectories, aircraft characteristics, ramp destinations, short-horizon traffic rates (arrivals, departures, crossing usage), and airport weather across multiple look-back windows. We benchmark nine classification methods spanning linear models, neural networks, and tree-based ensembles, with Random Forest, XGBoost, LightGBM, and CatBoost as primary candidates, and report accuracy as well as imbalance-aware metrics (macro-F1, precision–recall curves, confusion matrices, Brier score, Expected Calibration Error). Across east and west flows, XGBoost and LightGBM outperform Random Forest. Stage~I achieves accuracies of \mbox{0.86–0.89} with macro-F1 of \mbox{0.40–0.50}; Stage~II achieves \mbox{0.70–0.74} accuracy with macro-F1 of \mbox{0.28–0.55}. Feature importance analyses indicate that approach speed is the dominant driver of exit choice, while departure rate, crossing rate, ramp destination, and (for west flow) the taken exit are the strongest predictors of crossing vs.\ end-around. Although overall accuracies are high, minority classes exhibit lower recall, which motivates explicit imbalance handling (e.g., class weighting, focal objectives) for operational deployment. A class overlap analysis using t-SNE and UMAP identifies feature-space inseparability as the primary bottleneck for minority-class prediction. The framework is positioned as a controller-support tool intended to enhance ATCO situational awareness through calibrated, explainable predictions while preserving human judgment and operational responsibility for the final routing decision. The results provide interpretable, airport-specific insight into when tactical runway crossings are favored relative to the end-around and illustrate how predictive models can support controller situational awareness and surface-flow efficiency within future human--machine collaboration environments.
\end{abstract}


\end{frontmatter}

\section{Introduction \label{sec: introduction}}
Airport surface operations are increasingly a binding constraint on the safety and efficiency of the air transportation system. Following the COVID-19 pandemic, demand has rebounded sharply across the thirty primary U.S.\ airports: total operations grew by 3.8\% from 2022 to 2023 \citep{FAA2024}, and NASA anticipates global traffic roughly doubling by the mid-2030s \citep{nasaNAH2016}. Every additional arrival must vacate the runway at an appropriate high-speed exit, negotiate potential runway crossings, and reach its ramp without impeding departures or causing surface congestion. When surface operations flow smoothly, travelers arrive sooner, fuel burn and emissions drop, and departure throughput improves; when they do not, delays cascade and are costly to recover. Many U.S.\ towers also continue to operate below staffing targets (some at roughly 60\% of desired levels) which further increases the value of reliable and interpretable decision support for Air Traffic Controllers (ATCOs) \citep{FAA2024workforce}. NASA’s Aeronautics Research Mission Directorate (ARMD) has accordingly prioritized decision-support capabilities that improve predictability and reduce workload while preserving safety margins \citep{NASA_ARMD_2023}.

The taxi-in phase for arrivals is a particularly difficult component of this problem. At large hub airports, a single landing can face multiple plausible runway exits and, depending on airport geometry and departure flow, must either cross an active departure runway or take a longer end-around taxiway. The appropriate choice is context dependent: it varies with approach speed, aircraft type and weight class, exit geometry, ramp destination, contemporaneous arrival and departure rates, and even micro-scale meteorology affecting braking effectiveness and visibility. Some decisions appear straightforward in hindsight yet are difficult to optimize in real time, and there is ongoing debate over whether tactical runway crossings are preferable to strategic use of end-around routes and to what extent those decisions should be automated versus left to controller judgment.

Motivated by these operational realities, this paper asks: \textit{To what extent can a data-driven machine-learning (ML) pipeline accurately predict the runway exit that an arriving aircraft will use, and whether that aircraft will subsequently cross an active departure runway or instead take the end-around taxiway, at a complex, high-throughput airport?} We target the north complex at Hartsfield–Jackson Atlanta International Airport (KATL), a canonical setting where arrivals use Runway 8L/26R and departures use Runway 8R/26L, and where ATCOs routinely balance runway crossings against the end-around option. 

\begin{figure}[H]
   \centering
   \includegraphics[scale=0.35]{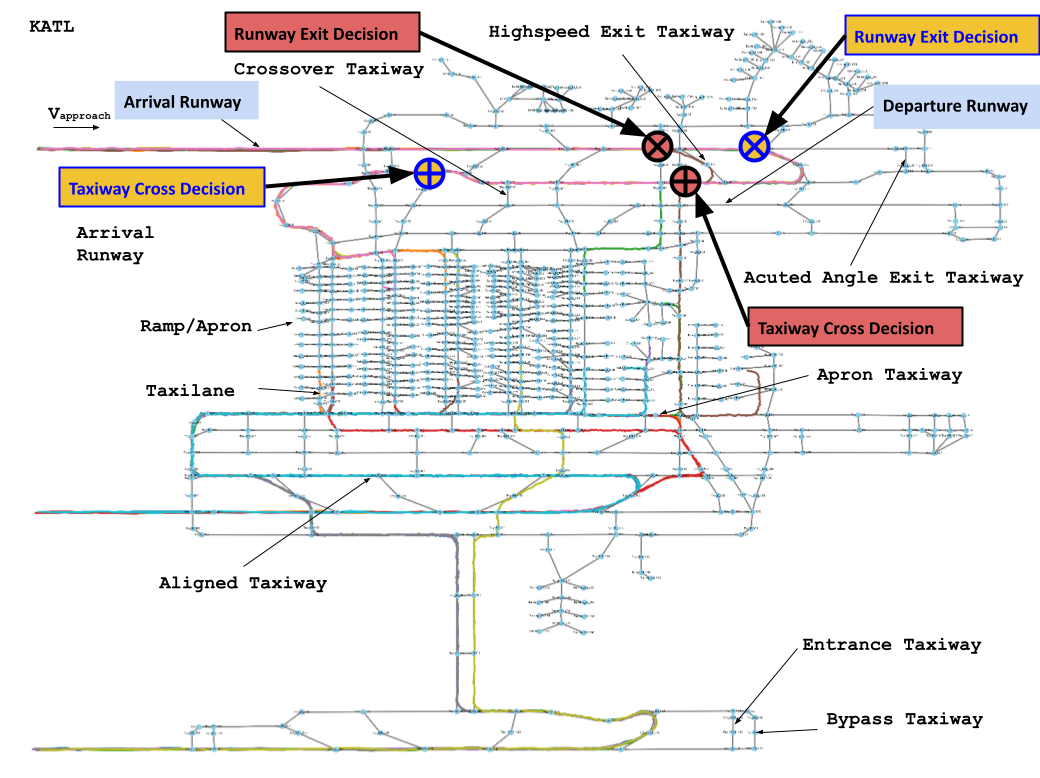}
   \caption{KATL’s north complex (arrivals on 8L/26R, departures on 8R/26L) concentrates the trade-offs that this paper targets: a high-throughput departure stream, multiple plausible high-speed exits, and a well-utilized end-around taxiway.}
   \label{fig:KATL_NORTH}
\end{figure}

A deliberate design choice in this work is to predict \emph{actual} operational decisions from historical data rather than prescribing optimal strategies. We position the framework as a controller-support tool, that is, an assistive aid that is intended to enhance ATCO situational awareness through calibrated, explainable forecasts while leaving the final tactical decision and operational responsibility with the human controller. This human-in-command framing aligns with current ATM modernization roadmaps that emphasize human--machine collaboration rather than autonomous replacement of controllers \citep{NASA_ARMD_2023, wang2022artificial}. This approach is well established in the runway exit prediction literature. \citet{herrema2019machine} train gradient-boosted models at Vienna Airport to predict actual exit choices, noting that efficiency considerations such as gate proximity and taxi-time minimization drive rational pilot decisions that would be difficult to formalize as explicit optimization objectives. \citet{martinez2018boosted} and \citet{meijers2019data} similarly learn from observed ASDE-X data at European and US airports; \citet{woo2022runway} develop exit prediction with explainability at Changi Airport; and \citet{mirmohammadsadeghi2016prediction} demonstrate that airline-specific gate assignments drive exit choices at multiple US airports. Accurate prediction of what \emph{will} happen is a prerequisite for evaluating what \emph{should} happen: it enables controllers to anticipate runway occupancy, identify likely deviations from expected paths, and assess probability distributions over outcomes.

\begin{figure}[H]
    \centering
    \includegraphics[scale=0.25]{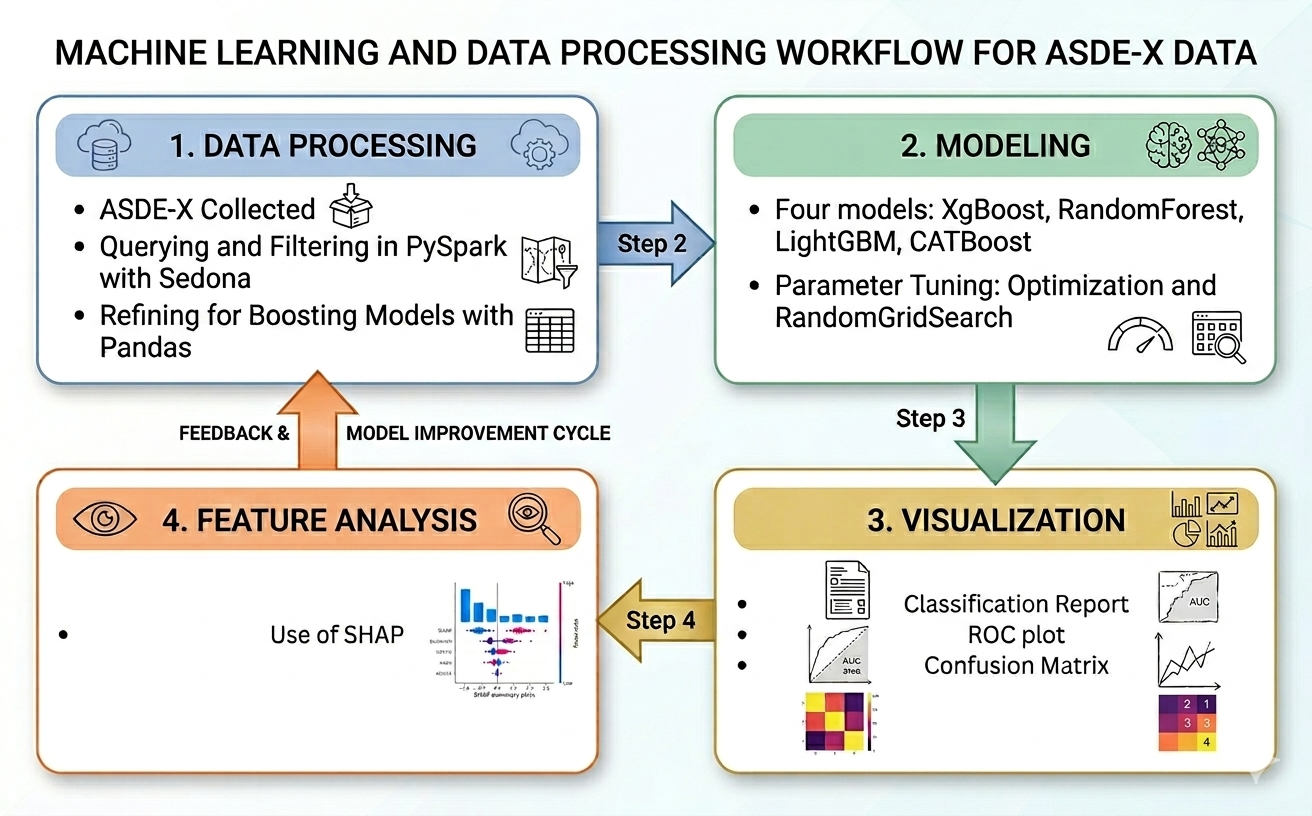}
    \caption{Schematic of study area and decision points on KATL's north complex.}
    \label{fig:Diagram}
\end{figure}

We focus on arrivals taxi-in at KATL’s north complex; departures, gate assignment, and collaborative surface scheduling are out of the scope except insofar as their rates affect the Stage~II decision. For east and west flows, arrivals face three candidate exits; post-exit, ATCOs must decide to cross the departure runway at one of several locations or to taxi around via the EAT. The overall data/feature flow is schematized in \Cref{fig:Diagram}. The layout and decision points are shown in \Cref{fig:KATL_NORTH} and the detailed naming of taxi/run-way exits is shown in \Cref{fig:mainfigure}. This setting yields sufficient data diversity to train and test models, while keeping the geometry and procedures consistent enough for meaningful feature attribution. Specifically, this paper proposes a two-stage classifier, trained on ASDE-X surface trajectories, weather forecasts, aircraft characteristics, and ramp destinations, to predict (Stage~I) the selected runway exit and (Stage~II) the crossing vs. end-around decision with sufficient accuracy and interpretability to serve as a practical decision aid for ATCOs at KATL. 

Classical contributions model taxi-out processes via queuing and fast-time simulation to improve pushback decisions and departure planning \citep{idris2001queuing}. These studies establish how runway configuration, downstream constraints, and takeoff queue size drive taxi-out variability and predictability \citep{keith2012optimization, lee2015taxi, jeong2020unimpeded}, and they demonstrate the sensitivity of surface flow to runway crossings and scheduling assumptions. Surface movement simulators and decision-support tools such as LINOS/SARDA-style approaches \citep{lee2015taxi} have been evaluated for real-time taxi time prediction and departure sequencing under uncertainty \citep{FAA2024, NASA_ARMD_2023}. In parallel, end-around taxiway (EAT) analyses indicate sizable environmental and throughput benefits when used under the right conditions; the magnitude of those benefits depends on departure rate, ramp destination, and geometry, so decision policies must weigh longer distances against fewer stops and independence from the departure stream \citep{fala2016surface}.

Several studies predict runway exit choices for arrivals using boosting models and rich feature sets. At Vienna, \citet{martinez2018boosted} and \citet{herrema2019machine} modeled runway occupancy time (ROT) \citep{meijers2019data} and procedural vs.\ non-procedural exit usage, achieving strong accuracies but reducing a multiclass exit choice into a binary classification at the runway-set level. \citet{woo2022runway} extended boosted classification at Changi's Runway 02L to predict one of three exits (with 80--90\% accuracy) using trajectory, METAR, and operational features, though aircraft type coverage was limited. Beyond exit selection, airport-level studies have predicted unimpeded taxi times with node-link models and routing algorithms such as A* \citep{zhao2021research, jeong2020unimpeded}, optimized departure and arrival schedules with mixed-integer programs \citep{keith2012optimization, kang2024dynamic}, and fit detailed taxi speed profiles with 4D surface trajectories for better timing prediction \citep{xu2021fourd, pham2021generative}. Taken together, these results indicate that exit choice is predictable from a small set of salient features and that surface timing and conflicts are tightly coupled to routing and crossing decisions.

A persistent barrier to operational use of ML in air traffic management is transparency. Recent work integrates eXplainable AI (XAI) to improve interpretability and calibrate human--machine teaming in safety-critical contexts \citep{xie2021explanation, sutthithatip2022explainable, wang2022artificial, pang2024machine}. This concern is especially acute for decision aids used during peak workload periods, where controllers and supervisors must understand the rationale behind a recommended exit or crossing alternative.

Despite this progress, several gaps motivate our study. Arrival runway-exit prediction has often been posed as a binary problem (procedural vs.\ non-procedural), eliding operationally distinct high-speed exits and limiting tactical fidelity \citep{martinez2018boosted, herrema2019machine}. While prior work quantifies end-around benefits at a policy level \citep{fala2016surface}, relatively few \emph{predictive} models forecast, for each arrival, whether ATCOs will choose to cross the departure runway or take the EAT, conditional on the runway exit and evolving departure rate. Across both subproblems, class imbalance, where dominant exits and crossing points versus rare alternatives, can inflate accuracy while masking poor recall on minority classes, a concern if the goal is to support nuanced minority decisions in real time. Finally, recent work in the imbalanced classification literature has established that class overlap, where instances of different classes share common regions in the feature space, can degrade classifier performance more severely than class imbalance alone \citep{vuttipittayamongkol2021class}, and that boosting methods (particularly CatBoost) are well suited for multiclass imbalanced tasks \citep{tanha2020boosting}. Whether class overlap limits runway exit and crossing prediction has not been investigated in the ATM literature, yet the answer has direct implications for the achievable performance ceiling of any data-driven approach.

Our contributions are as follows. We model the specific runway exit chosen (three candidate exits per flow direction, Stage~I) and the subsequent decision to cross the departure runway at a designated taxiway point or to use the end-around taxiway (Stage~II). This two-stage design mirrors actual controller workflow: exit choice first, then crossing versus end-around conditioned on that exit. The feature set integrates ASDE-X trajectories, aircraft characteristics, ramp destinations, short-horizon operational rates (arrivals, departures, crossing usage) over look-back windows from 5 to 60~minutes, and airport weather (wind, visibility). Consistent with related work, we find that approach speed, ramp destination, and contemporaneous departure and crossing rates are among the most informative predictors, with differences by flow direction.

We benchmark nine classification methods---four tree-based ensembles (Random Forest, XGBoost, LightGBM, CatBoost) and five baselines (Logistic Regression, SVM, KNN, MLP, Decision Tree), and report macro-averaged F1, ROC--AUC, precision--recall curves, confusion matrices, per-class metrics, Brier score, and Expected Calibration Error (ECE). SHAP-based global importance connects predictions to operationally meaningful variables, aligning with the explainability practices recommended for ATM decision support \citep{NASA_ARMD_2023}.

We further conduct a systematic ablation of imbalance mitigation strategies (SMOTE, cost-sensitive learning, calibrated one-vs-rest classifiers) and a class overlap analysis using t-SNE \citep{van2008visualizing} and UMAP \citep{mcinnes2018umap}. The analysis identifies feature-space inseparability as the primary bottleneck for minority-class classification. Probability calibration metrics (Brier score \citep{redelmeier1991assessing}, ECE \citep{nixon2019measuring}) confirm well-calibrated probabilistic outputs suitable for controller decision support.

Finally, we quantify how ramp destination and exit geometry interact with departure rate and crossing usage to drive crossing versus end-around choices at KATL, yielding empirical thresholds that are consistent with, but more granular than, prior EAT policy analyses \citep{fala2016surface}. These insights can inform local standard operating procedures, training, and what-if fast-time studies.

The remainder of the paper proceeds as follows. \Cref{sec: literature review} reviews related work on optimization-based routing, unimpeded taxi-time modeling, 4D trajectory fitting, end-around taxiway analysis, and ML-based exit prediction. \Cref{sec: empirical} describes the KATL north-side data and exploratory analyses that informed feature engineering. \Cref{sec: methodologies} details the modeling pipeline, including problem formulation, model families, metrics, hyperparameter tuning, and class imbalance handling. \Cref{sec: discussion} presents results and interprets them in operational terms, including limitations related to imbalance effects, minority-class recall, and generalizability beyond KATL. \Cref{sec: conclusion} concludes with implications for ATCO decision support and directions for integration with fast-time simulation and optimization.

\section{Literature Review  \label{sec: literature review}}
Research on airport surface operations has progressed from analytical queuing and network optimization to data-driven prediction and XAI, with increasingly realistic agent-based simulation. Foundational queuing work showed that macroscopic surface-state variables dominate taxi-out time variability and are therefore useful features for predictive models and advisory tools \citep{idris2001queuing}. Subsequent studies refined these models and compared them to running-average baselines, demonstrating meaningful reductions in prediction error \citep{lee2015taxi}. These results informed fast-time simulation environments and probabilistic planners that explicitly represent uncertainty in link times and traffic evolution, enabling \textit{what-if} comparisons between stochastic scheduling and deterministic heuristics \citep{balakrishna2008estimating, yin2024prediction}.

At a finer scale, 4D trajectory prediction methods fit velocity profiles to historical surveillance tracks, learning route-conditioned, aircraft-aware nominal speed profiles that map to instantaneous velocities for time-of-arrival prediction at intersections. Case studies report reduced timing error relative to purely dynamic-model baselines \citep{xu2021fourd}. Related work frames taxi-speed control as an optimal control or reinforcement-learning problem and uses generative-adversarial imitation learning to reproduce human-like taxi-speed patterns \citep{balakrishna2008estimating, pham2021generative, ali2022deep}. These approaches reduce dependence on hard-to-identify aerodynamic parameters and instead leverage data-driven warping and averaging to capture realistic taxi behavior across fleet mix and pavement conditions \citep{xiang2023application, sui2023conflict}.

A large parallel stream optimizes movements at the network level. Mixed-integer programming (MIP) and multi-agent path finding have been used to co-optimize pushbacks, taxi routes, and runway usage subject to separation, conflicts, and evolving traffic states \citep{roling2008optimal, atkin2010airport, clare2011optimization}. Under uncertainty, planners integrate probabilistic link times and re-planning, with fast-time simulation harnessed to test robust schedules and controller advisories \citep{ma2016optimal, li2019departure}. Recent frameworks automatically derive node–link models from airport maps, embed aircraft-dynamics–aware speed models, and generate itineraries consistent with runway and gate constraints, creating an end-to-end loop for evaluation and benchmarking of routing/scheduling strategies \citep{zhao2021research}.

\subsection{Runway Exit Prediction}

Runway exit prediction involves forecasting which taxiway exit a landed aircraft will take, given factors such as aircraft type and weight class, approach speed, wind conditions, runway and exit geometry, and terminal ramp location.

\cite{martinez2018boosted} developed a boosted-tree framework to predict runway occupancy time (ROT) and runway exit decisions at Vienna International Airport using radar trajectories, meteorology, and flight-plan details. The study suggests that runway throughput can increase while maintaining safety margins if ROT and exit choice are predicted accurately. A binary exit-set formulation (procedural vs.\ non-procedural) achieved 77\% accuracy overall with lower precision on minority behaviors, highlighting class-imbalance effects \citep{martinez2018boosted, barua2021gradient}. \cite{herrema2019machine} extended this work with weight-class-specific gradient boosting at Vienna, reporting very high accuracy for procedural exits (95–98\%) and lower but still strong results for non-procedural exits (73–79\%), again reflecting the difficulty of rare behaviors. At Singapore Changi (Runway 02L), \cite{woo2022runway} used scalable boosting on trajectories, METAR, and operational features to attain 80–90\% accuracy across three exits, though coverage was limited to three aircraft types, constraining transferability.

Broader data-driven literature connects exit choice to ROT variability and approach dynamics. Using large-sample surface-surveillance data, \cite{meijers2019data} quantify how exit used, aircraft type, airline, final approach speed, and the presence of a follower on approach jointly explain most ROT variance, empirically linking approach speed and exit location to runway clear times. In practice, exit-selection models benefit from combining trajectory-derived speed features, ramp destinations, and near-term traffic pressure, which sharpen predictions while also exposing class-imbalance effects \citep{weiszer2015real, meijers2019data, woo2022runway}. Across these studies, boosted trees (e.g., XGBoost, LightGBM) provide a favorable accuracy--interpretability trade-off, especially when paired with feature-attribution methods to explain predictions to operators \citep{xie2021explanation}.

\subsection{End-Around Prediction}
Most existing work on airport surface movement prediction focuses on taxi-time prediction or on optimizing routing and scheduling. For unimpeded taxi time, \cite{jeong2020unimpeded} use a node--link model with link travel times conditioned on flight type, weight class, and leg type, yielding approximately 1-minute RMSE for both departures and arrivals. Mixed-integer models that account for separation constraints, continuous time, and routing conflicts have been applied at airports with complex holding-point structures (e.g., Heathrow), producing average time savings of roughly 20\% relative to sequential or ad-hoc procedures (Keith and Richards, 2012). Recent predict-then-optimize systems combine short-horizon time prediction with conflict-aware routing and scheduling, reporting 30--35\% reductions in additional taxi time over sequential baselines at large Asian hubs \citep{kang2025dynamic}. Conflict detection and resolution is increasingly integrated with routing to ensure feasible, deconflicted itineraries \citep{atkin2010airport, clare2011optimization, zhao2021research}.

There have also been studies on end-around taxiway (EAT) usage. \cite{fala2014surface} analyze three airports with EATs and find environmental benefits (fuel/emissions) of 8–25\% under rule-based EAT usage in Monte Carlo experiments. Complementing that perspective, discrete-event simulation and surveillance-data–driven analyses at ATL’s north complex quantify the cross vs.\ end-around trade-off as a function of departure rate, crossing rate, fleet mix, and permissible hold-for-crossing delays. Departure rate is often the dominant driver of whether crossing is time-competitive with the end-around. While the end-around can be longer in distance, it avoids intersection stops and keeps departure streams independent, thereby mitigating stop–go fuel penalties and reducing runway incursion risk \cite{fala2014surface, fala2016surface, wang2024simulation, zhao2025flexible}. These results motivate predictive tools that, conditional on the runway exit taken, anticipate whether a given arrival will cross or route via the end-around—precisely the second stage modeled in our study.

Queue-aware predictors \citep{idris2001queuing, sui2023conflict} and robust schedulers \citep{zhao2021research, keith2012optimization, kang2024dynamic, atkin2010airport, roling2008optimal} establish macroscopic levers such as holding policies and route choices, while 4D trajectory and speed-profile methods capture microscopic behavior at exits and intersections \citep{xu2021fourd, pham2021generative}. Data-driven analyses connecting approach speed, exit choice, and ROT \citep{meijers2019data} justify using approach kinematics, ramp destinations, and short-horizon traffic rates as features for predicting both exit selection and the crossing-versus-end-around decision. Our work discretizes the controller decision-making process into a first-stage runway-exit classifier followed by a second-stage crossing-point or end-around classifier, aligning predictive outputs with controller workflow at KATL’s north complex. Explainable-ML tooling (e.g., SHAP) supports operator trust by attributing predictions to interpretable features such as approach speed, ramp destination, and departure and crossing rates \citep{xie2021explanation}.

\section{Empirical Data Analysis \label{sec: empirical}}
This section summarizes empirical regularities in the KATL north-side dataset and motivates the features used in the prediction models. We define labels and strata (flow direction, ramp, aircraft class), describe how raw sources were synchronized and filtered, construct operational and environmental covariates over multiple lookback windows, and report exploratory findings that justify the two-stage classification design and imbalance-aware evaluation.

\begin{figure}[H]
    \centering
    
    \begin{subfigure}[b]{0.48\textwidth}
        \centering
        \includegraphics[width=\linewidth]{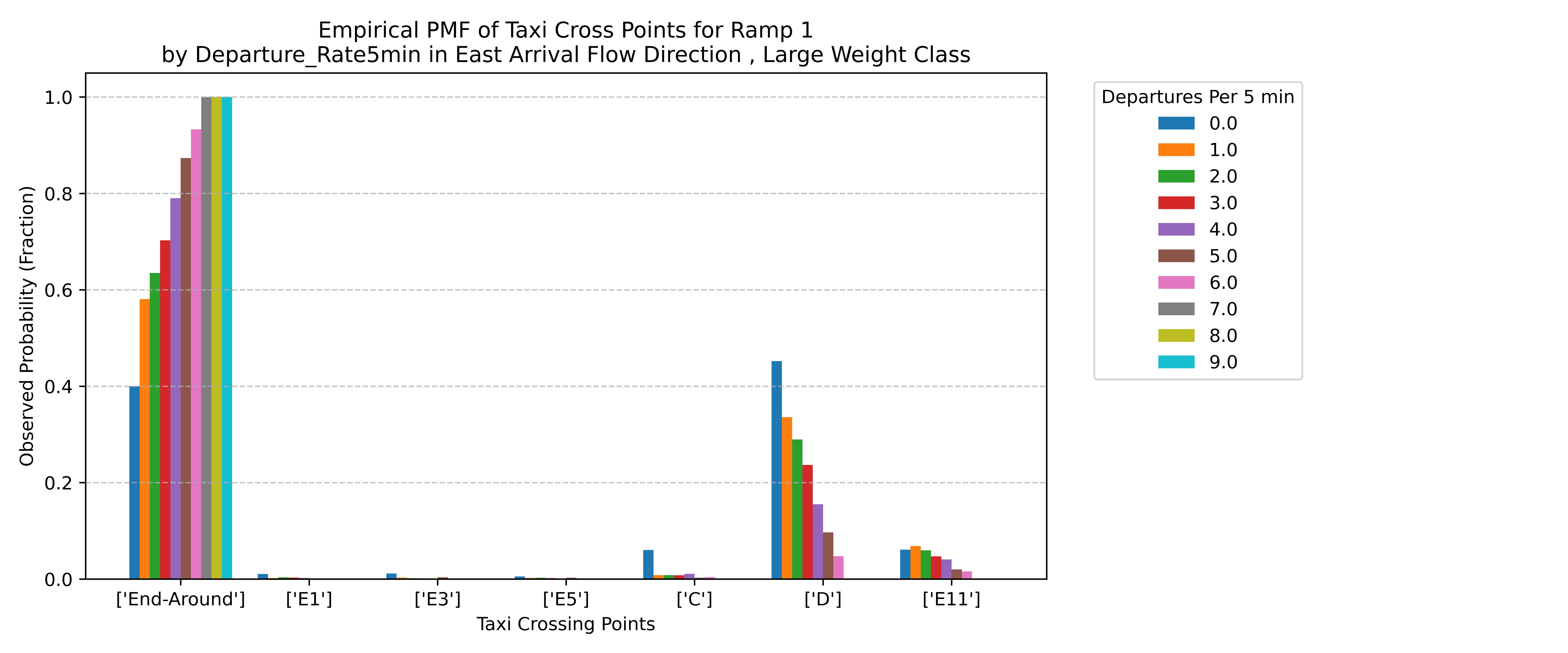}
        \caption{Ramp 1 — Large Aircraft Weight Class}
        \label{fig:Ramp1Elarge}
    \end{subfigure}
    \hfill
    \begin{subfigure}[b]{0.48\textwidth}
        \centering
        \includegraphics[width=\linewidth]{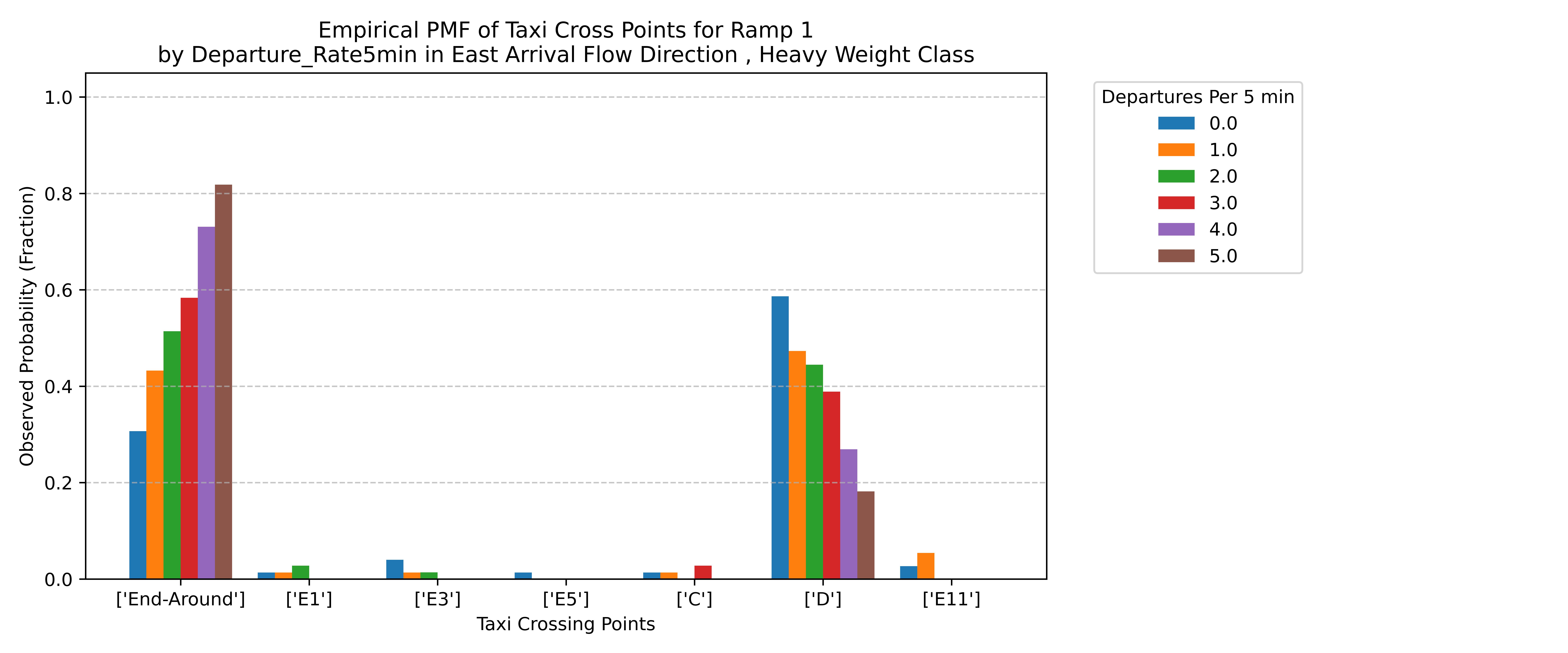}
        \caption{Ramp 1 — Heavy Aircraft Weight Class}
        \label{fig:Ramp1Eheavy}
    \end{subfigure}

    \vspace{0.5em}

    \begin{subfigure}[b]{0.48\textwidth}
        \centering
        \includegraphics[width=\linewidth]{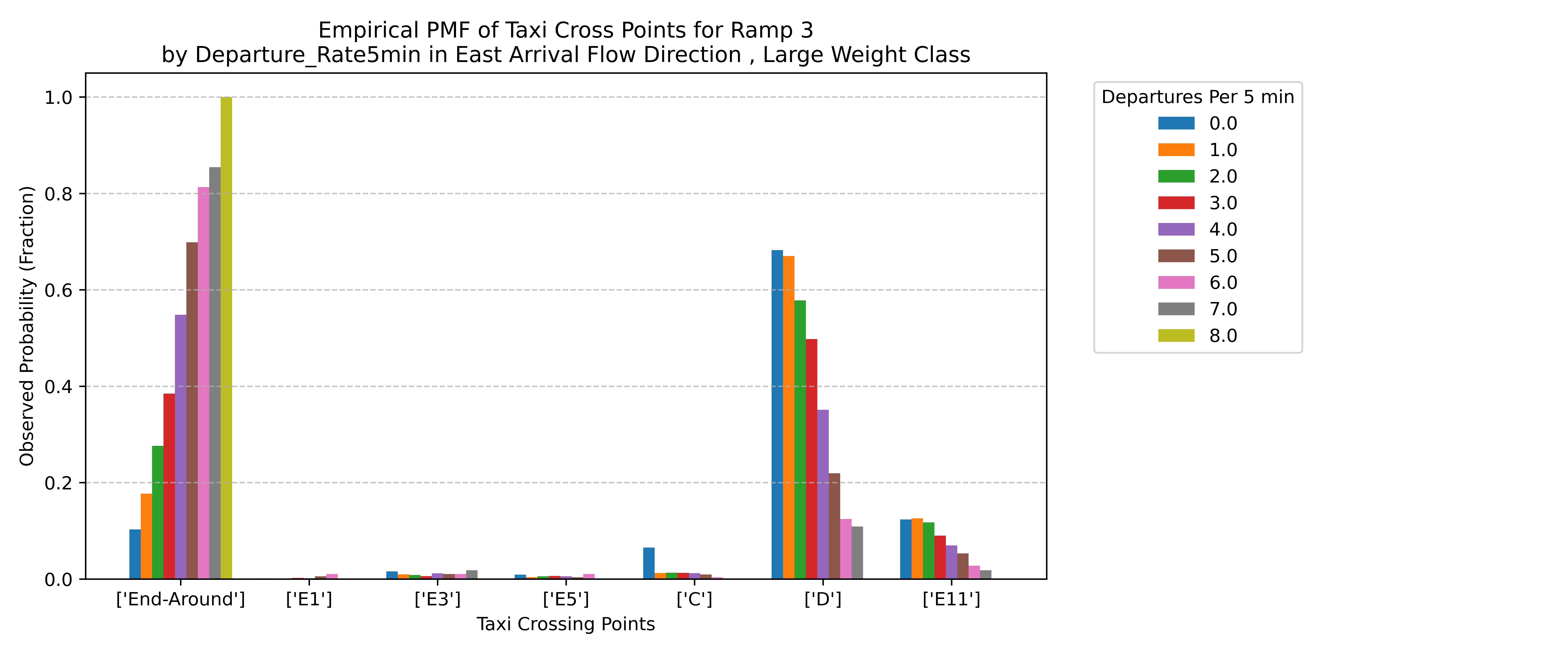}
        \caption{Ramp 3 — Large Aircraft Weight Class}
        \label{fig:Ramp3Elarge}
    \end{subfigure}
    \hfill
    \begin{subfigure}[b]{0.48\textwidth}
        \centering
        \includegraphics[width=\linewidth]{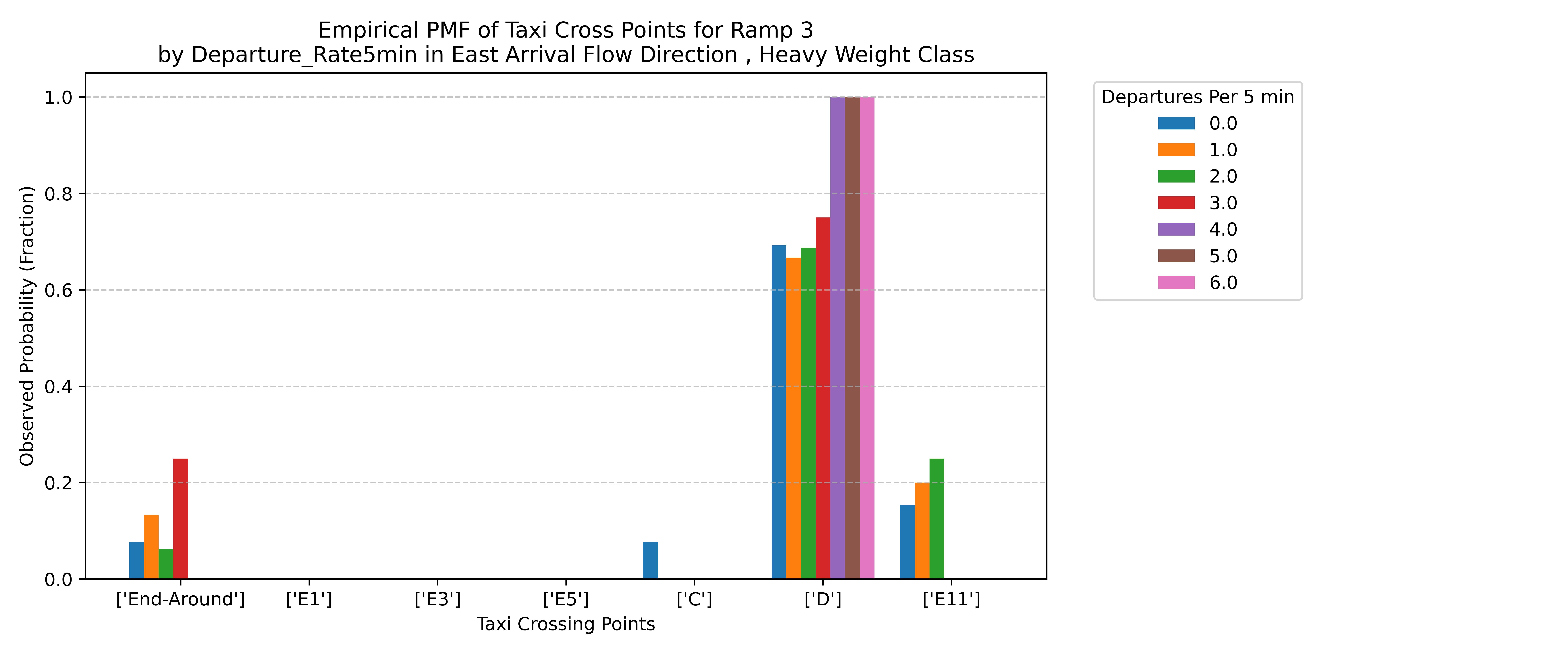}
        \caption{Ramp 3 — Heavy Aircraft Weight Class}
        \label{fig:Ramp3Eheavy}
    \end{subfigure}

    \vspace{0.5em}

    \begin{subfigure}[b]{0.48\textwidth}
        \centering
        \includegraphics[width=\linewidth]{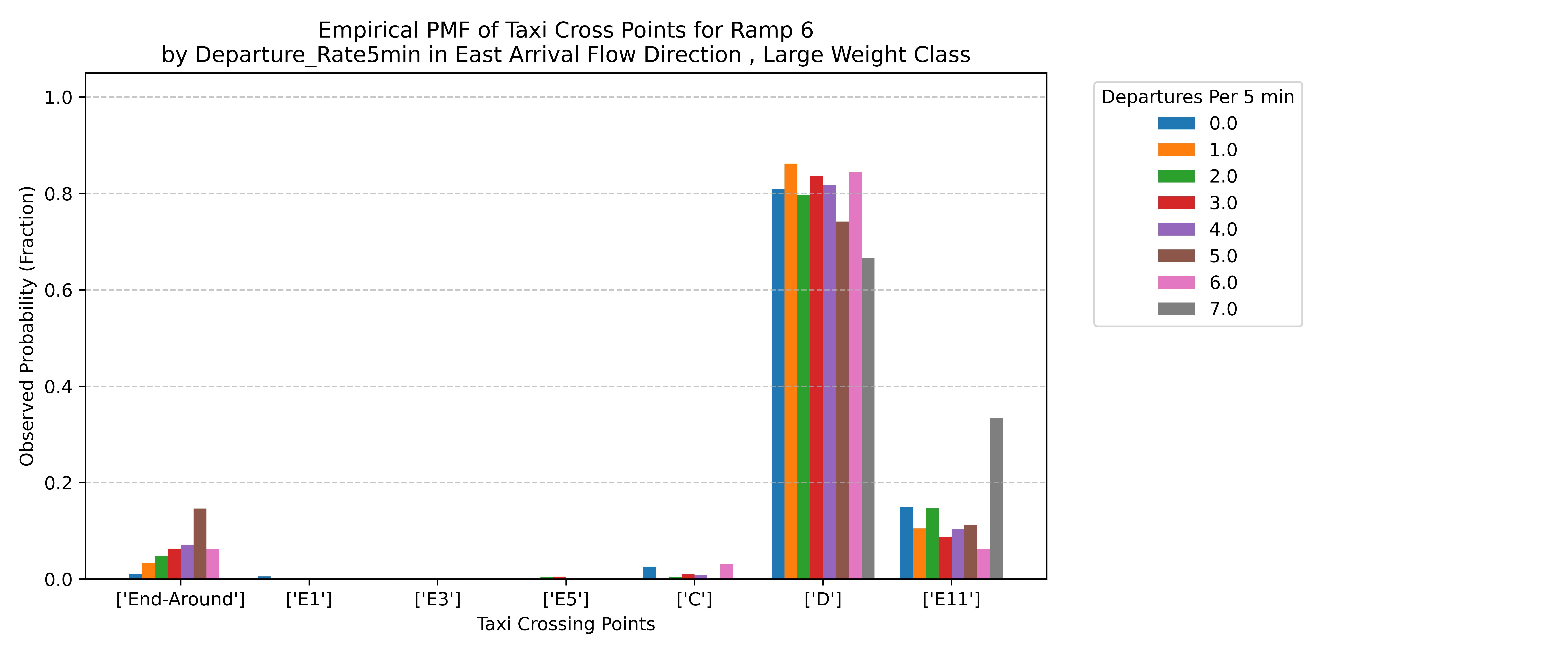}
        \caption{Ramp 6 — Large Aircraft Weight Class}
        \label{fig:Ramp6Elarge}
    \end{subfigure}
    \hfill
    \begin{subfigure}[b]{0.48\textwidth}
        \centering
        \includegraphics[width=\linewidth]{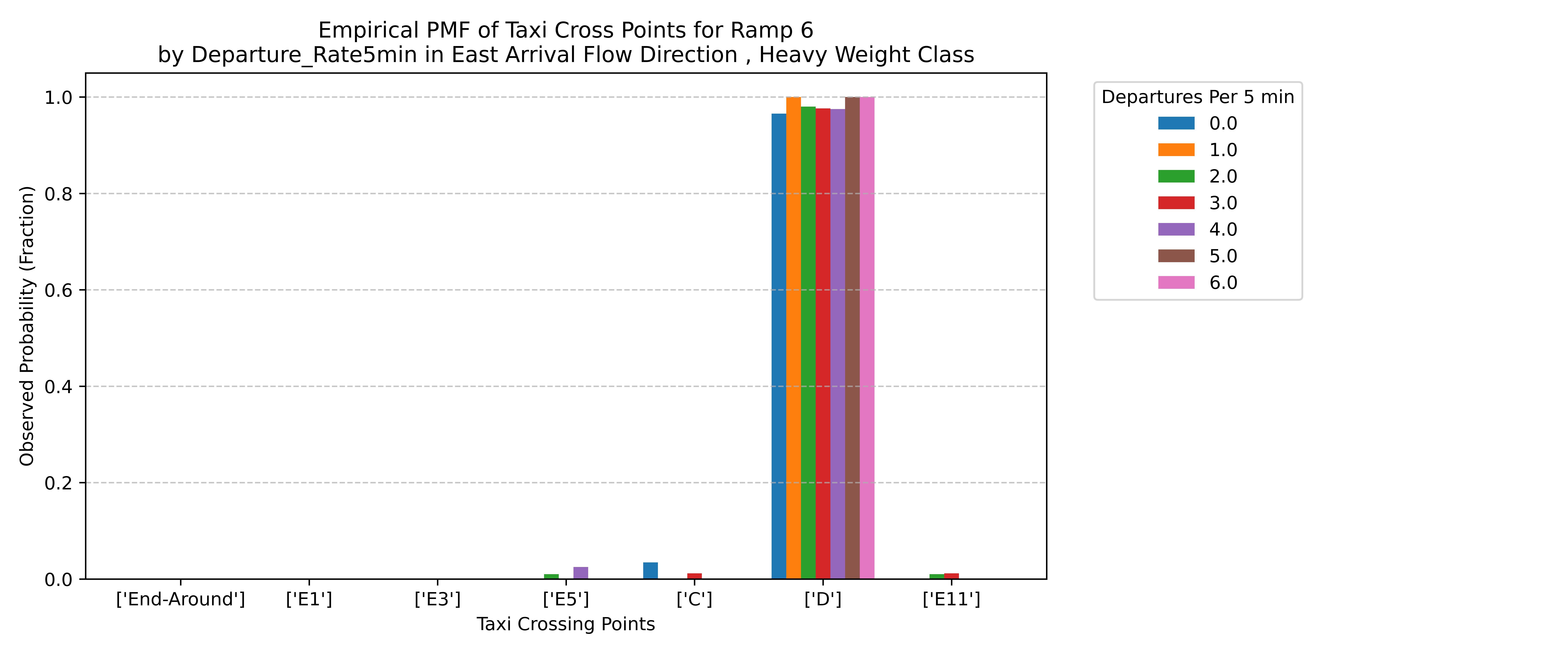}
        \caption{Ramp 6 — Heavy Aircraft Weight Class}
        \label{fig:Ramp6Eheavy}
    \end{subfigure}
    \caption{PMF histograms of the taxiways taken right before reaching the destination ramp.}
    \label{fig:NorthSideKATL}
\end{figure}

Figure~\ref{fig:NorthSideKATL} visualizes, via probability mass functions (PMFs), the distribution of last taxiway segments used immediately prior to ramp entry, stratified by flow direction (east) and by ramp destination (Ramps 1, 3, and 6), with panels separated by aircraft weight class (\emph{Large} vs.\ \emph{Heavy}). Two qualitative regularities emerge. First, within a fixed flow and ramp, the mass shifts across taxiway alternatives as departure rate changes, indicating that short-horizon traffic pressure is informative for route choice. Second, the \emph{Heavy} class exhibits a distinct selection pattern relative to \emph{Large}, consistent with differences in stopping distances, turning radii, and standard operating procedures. These separations suggest that weight class, ramp destination, and contemporaneous traffic rates are predictive of pre-ramp routing and, of crossing versus end-around usage, making classification modeling appropriate.

\subsection{Data Sources and Feature Engineering}
The study draws on two primary data sources. Airport surface movement data (ASDE-X) were obtained from the Sherlock Data Warehouse \citep{pang2021data}. ASDE-X provides per-second surface tracks with position, ground speed, and identifiers (callsign, transponder code, aircraft type). We extracted all movements within spatial polygons covering the KATL north complex (8L/26R arrivals, 8R/26L departures, associated high-speed exits, crossing points, and the end-around taxiway). Track segments were snapped to a cleaned node--link map of the north-side taxiway network to recover event times at touchdown, exit nose crossing, runway hold-short lines, departure-runway centerline crossing (if any), and ramp entry points. Flow direction (east vs.\ west) was determined by runway assignment (26R vs.\ 8L) and verified by touchdown ground track. Weather data were obtained from VisualCrossing as in \cite{pang2024machine} at 5-minute resolution and time-aligned to ASDE-X via nearest-neighbor matching with a maximum tolerance of $\pm$2.5 minutes. Weather variables include surface wind speed and direction, visibility, precipitation flags, and temperature. Winds were decomposed into \emph{headwind} and \emph{crosswind} components relative to runway heading and, for Stage~II, relative to end-around versus crossing routes, to capture braking and turning implications.

The study period spans January~2022 through December~2023. Arrivals were included if touchdown occurred on 8L (west flow) or 26R (east flow), the post-exit route remained on the north complex, and a valid ramp destination was recorded. Taxiway segments external to the north complex and special cases (e.g., tows, diversions) were filtered out using speed and track heuristics and operational flags. For each arrival, the Stage~I label is the first high-speed exit used to vacate the landing runway. Following facility conventions, exits are grouped into three per flow (B7/B11/B13 for east; B1/B3/B5 for west). The exit event time is the nose crossing of the exit centerline, which ensures temporal ordering relative to touchdown. Conditional on the Stage~I outcome, Stage~II labels whether the aircraft crosses the active departure runway (8R/26L) at a designated taxiway (e.g., C, D, E1, E3, E5, E11) or uses the end-around taxiway. The crossing event is detected by intersection of the track with the departure-runway polygon; end-around usage is detected by traversals through the end-around link set. The feature set also includes aircraft type and weight class, the class and type of the preceding and following arrivals in sequence, and the presence of departures queued at the relevant crossing points at exit time.

Airport surface conditions vary at short horizons; we therefore compute rolling rates over multiple windows $\Delta \in \{5,10,15,30,60\}$ minutes prior to the decision point:
\begin{itemize}
    \item Arrival rate $\lambda^\text{arr}_\Delta$: landings per minute on the arrival runway.
    \item Departure rate $\lambda^\text{dep}_\Delta$: takeoffs per minute on the departure runway.
    \item Crossing rate $\lambda^\text{cross}_\Delta$: number of aircraft that crossed the departure runway per minute (proxy for recent controller willingness/availability of crossing gaps).
\end{itemize}
For Stage~I, rates are computed with respect to touchdown time; for Stage~II, with respect to the exit time, to avoid look-ahead bias.

Several derived features supplement the raw data. Final approach speed is the ground speed averaged over the last $d$ meters before threshold; touchdown ground speed and deceleration proxies (slope of speed versus distance in the first $s$ meters post-touchdown) capture stopping capability and the likelihood of making earlier exits. Ramp destination is encoded as categorical indicators. Graph distances from each exit to each ramp are computed separately for crossing routes and end-around routes; the differential distance and estimated travel time between these alternatives serve as Stage~II covariates. For Stage~I, exit geometry (exit angle, paved length to first turn) is included as static attributes.

All features for Stage~I are computed strictly from data available up to touchdown; Stage~II features are computed up to exit time. Any variables determinable only after the decision (e.g., observed crossing time) are excluded to avoid target leakage. Missing values in categorical fields (e.g., aircraft type) are imputed to an \texttt{UNKNOWN} category; continuous weather/kinematic gaps are forward-filled within a maximum tolerance and otherwise flagged. Outliers in speed (e.g., spikes due to multilateration swaps) are smoothed via median filters with robust clipping. Categorical variables (aircraft type, ramp, exit) are one-hot encoded; ordinal bins are used only where physically justified (e.g., discretized wind components). All rate variables are standardized within flow direction to mitigate configuration effects.

\subsection{Empirical Studies}
Within each flow, PMFs in Figure~\ref{fig:NorthSideKATL} indicate non-uniform taxiway selection by ramp and weight class. The pattern is consistent with the intuition that higher $\lambda^\text{dep}$ reduces the attractiveness of routes that require a crossing stop, particularly for Heavies. The conditional PMFs show earlier exits are less likely at higher approach/touchdown speeds and under tailwind components; conversely, stronger headwinds and lower touchdown speeds increase early-exit probability. These relationships support using approach speed and wind as leading predictors in Stage~I. Both stages exhibit skewed label distributions (cf.\ Section~\ref{sec: discussion}): one exit/class often dominates within a flow, and specific crossing points are rare relative to end-around usage (or vice versa) depending on configuration and demand. Consequently, we report macro-averaged metrics and precision–recall curves and use stratified splits to preserve prevalence.

Decision points are temporally and causally ordered. ATCOs (and pilots) first decide on the runway exit, then choose to \emph{cross} or use end-around based on the first stage decision. A single multiclass formulation mixing exits and crossing points would confound these mechanisms and dilute minority classes. The two-stage design preserves structure and yields interpretable feature attributions aligned with controller workflow.

\subsection{Summary}
In summary, approach and kinematic variables, short-horizon operational rates, ramp destination together with route geometry, and aircraft weight class are the principal correlates of exit and routing choices on KATL’s north complex. These findings justify the feature set and the staged framing adopted in \Cref{sec: methodologies} and motivate the imbalance-aware training and evaluation strategies reported in \Cref{sec: discussion}.

Technologically, all ASDE-X extraction, spatial joins, and time-window aggregations were implemented in PySpark with Sedona for geospatial indexing; results were materialized to columnar files and post-processed in \textsf{pandas}. Spatial filtering used polygon masks for the north-side complex; network distances were computed on a cleaned graph representation of taxiways. Feature stores for Stage~I and Stage~II were generated separately to enforce the temporal cutoffs described above.

\section{Methodologies \label{sec: methodologies}}

\subsection{Problem Formulation}

\begin{figure}[H]
    \centering
    \includegraphics[scale=0.4]{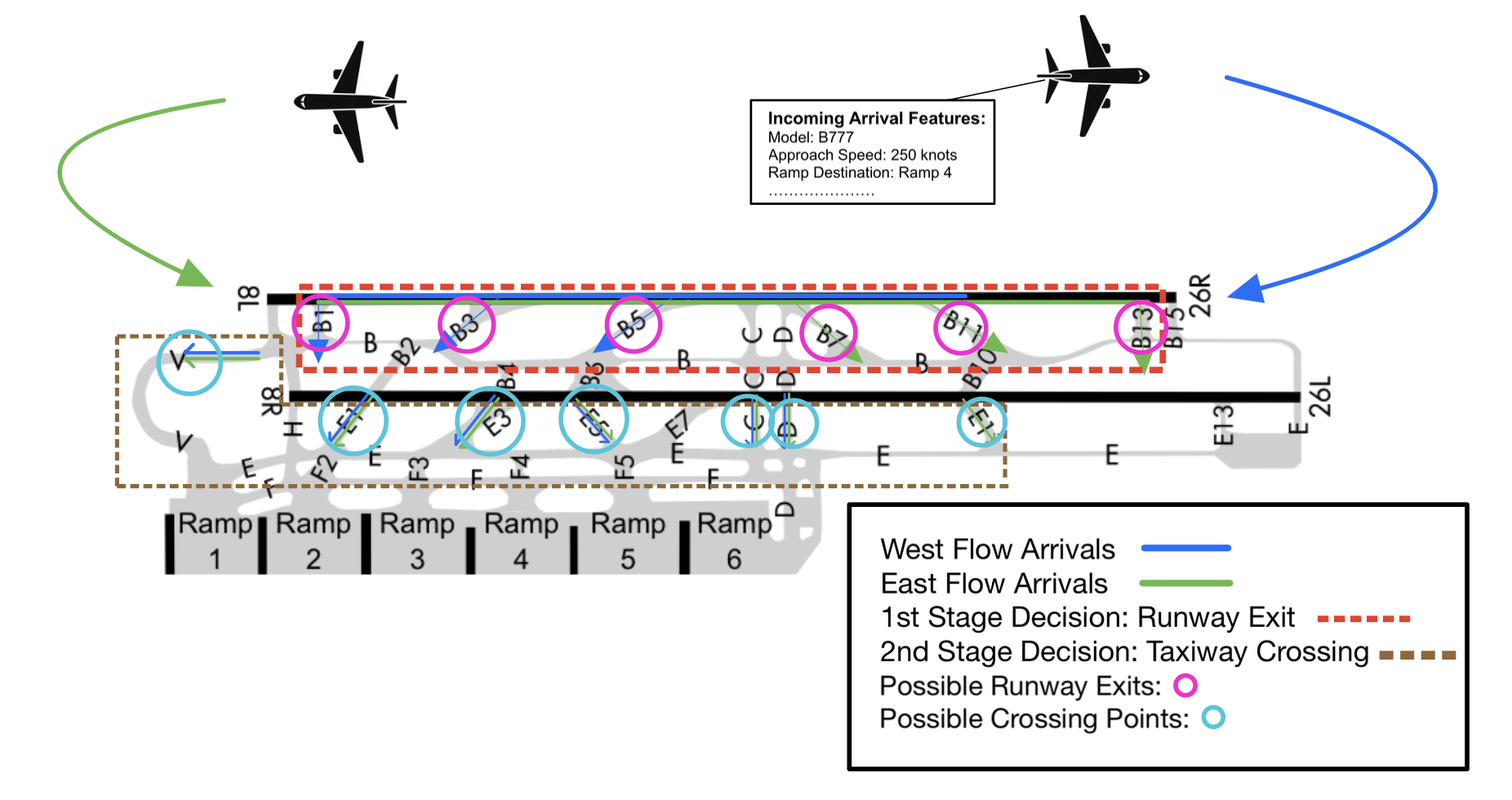}
    \caption{Layout of runway 8L/26R (arrivals) and 8R/26L (departures) and stage-wise decision spots. Stage~I predicts runway exit; Stage~II predicts crossing vs.\ end-around and, if crossing, the specific crossing point.}
    \label{fig:mainfigure}
\end{figure}

We formulate arrival taxi routing as a two-stage supervised classification problem aligned with controller workflow (exit selection $\rightarrow$ routing to ramp). Let $\mathcal{D}=\{(\mathbf{x}_i,y^{(1)}_i,y^{(2)}_i)\}_{i=1}^{N}$ be the dataset of arrivals, where $\mathbf{x}_i$ is the feature vector available up to the relevant decision time, $y^{(1)}_i\!\in\!\{1,\dots,K_1\}$ is the Stage~I label (runway exit) and $y^{(2)}_i\!\in\!\{1,\dots,K_2\}$ is the Stage~II label (crossing class vs.\ end-around), defined only after a valid Stage~I event. In our setting, $K_1=3$ per flow (east: B7/B11/B13; west: B1/B3/B5), and $K_2=7$ including end-around (east/west sets differ by available cross points).

\subsection{Prediction Objective}
Stage~I predicts $y^{(1)}$ at touchdown using only pre-touchdown or instantaneous features (approach/landing kinematics, head/cross-wind, traffic rates up to touchdown, ramp destination, aircraft class/type, geometry). Stage~II predicts $y^{(2)}$ at exit time using the realized exit (as a categorical feature), updated traffic rates and queue state, route geometry (graph distance/time via crossing vs.\ end-around), winds at exit time, and the same static attributes.

For each stage $s\in\{1,2\}$, a classifier $f^{(s)}:\mathbb{R}^p\!\rightarrow\!\Delta^{K_s-1}$ returns class probabilities via softmax scoring function,
\begin{equation}
p^{(s)}_{ik} \;=\; \frac{\exp\{f^{(s)}_k(\mathbf{x}_i)\}}{\sum_{\ell=1}^{K_s}\exp\{f^{(s)}_\ell(\mathbf{x}_i)\}},\quad k=1,\dots,K_s.
\end{equation}

Predicted labels are $\hat{y}^{(s)}_i=\arg\max_k p^{(s)}_{ik}$ unless otherwise stated (e.g., calibrated decision rules). Both stages exhibit skewed prevalence. We therefore minimize a class-weighted cross-entropy as,
\begin{equation}
\mathcal{L}^{(s)} \;=\; - \frac{1}{N_s}\sum_{i=1}^{N_s} w^{(s)}_{y^{(s)}_i}\,\log p^{(s)}_{i,y^{(s)}_i}.
\end{equation}

Let $t_\text{td}$ and $t_\text{ex}$ denote touchdown and exit times. For windows $\Delta\in\{5,10,15,30,60\}$ minutes we compute short-horizon rates as,
\begin{equation}
\begin{aligned}
\lambda^\text{arr}_\Delta(t) &= \frac{\#\text{arrivals in }[t-\Delta,t]}{\Delta} \\ 
\lambda^\text{dep}_\Delta(t) &= \frac{\#\text{departures in }[t-\Delta,t]}{\Delta} \\ 
\lambda^\text{cross}_\Delta(t) &= \frac{\#\text{crossings in }[t-\Delta,t]}{\Delta}
\end{aligned}
\end{equation}

For Stage~I we evaluate rates at $t_\text{td}$, for Stage~II at $t_\text{ex}$. Kinematic features include final-approach speed (mean ground speed over the last $d$~m before threshold), touchdown speed, and post-touchdown deceleration proxies. Weather features (wind speed/direction, visibility, precipitation) are aligned at 5-min granularity; wind is decomposed into headwind and crosswind components relative to runway heading. Geometry features include exit angle and paved length; Stage~II adds graph distances and estimated travel times from the realized exit to the destination ramp via crossing and end-around alternatives, as well as their difference. Categorical variables (aircraft type, weight class, ramp, exit) are one-hot encoded.


\subsection{Modeling and Training}
The problem is formulated into a two-part prediction task that aims to forecast the taxi path of the arriving aircraft on the north side of KATL, starting from the runway exit point and ending at the ramp destination (\Cref{fig:mainfigure}). The first part is to predict which one of the three runway exits the incoming arrival would like to take. The east flow uses east runway exits, while the west flow uses western exits. By incorporating features of flight dynamics and characteristics, operational information, and environmental conditions into the boost model, a runway exit decision can be predicted. Immediately after arrival, exit the landing runway; the runway exit will be identified and used as a feature for the second step. The second step is to decide whether the taxi aircraft will take an end-around or cross the departure runway at specific points. The boosting models use a similar set of features to the first step, exchaging the approach speed feature with the taken runway exit feature. Since each step prediction requires more than two options to predict, both problems are multiclass classification problems. Another consideration for this problem is that the labels and target values to expect, are not evenly distributed. With unbalanced labels, the model can be biased toward predicting in favour of the majority class. However, with synthetic data, a large training data size, and hyperparameter tuning, this challenge can be overcome. We evaluate four tree-based ensembles as primary models:
\begin{itemize}
    \item Random Forest (RF) Bagged decision trees with bootstrap sampling and random feature subsets; robust to noise and minimal preprocessing.
    \item XGBoost Gradient-boosted trees with second-order optimization, shrinkage, column subsampling, and built-in handling of class weights; early stopping on a validation set.
    \item LightGBM Histogram-based gradient boosting with GOSS/leaf-wise growth, efficient on large, sparse one-hot features; strong performance on tabular data.
    \item CatBoost Gradient-boosted trees with ordered boosting to reduce prediction shift, symmetric tree structure, and native support for categorical features and automatic class weighting; early stopping on a validation set.
\end{itemize}
We additionally benchmark five baseline methods to contextualize ensemble performance: Logistic Regression, Support Vector Machine (Linear SVM), $K$-Nearest Neighbors (KNN, $k=5$), Multilayer Perceptron (MLP with two hidden layers of 128 and 64 neurons), and Decision Tree. Features are standardized for methods that require it (Logistic Regression, SVM, KNN, MLP); tree-based methods operate on raw features. The choice of gradient-boosted tree ensembles is motivated by their strong empirical performance on structured tabular data \citep{grinsztajn2022tree, qin2021neural}. Boosting methods add base learners sequentially, each trained on the residual of the current ensemble, which reduces bias more effectively than bagging. They also yield feature importance rankings that are useful for understanding input--output relationships, and they can capture nonlinear decision boundaries that simple trees cannot. Gradient Boosting Machine (GBM) is a widely used instantiation \citep{friedman2000additive, friedman2001greedy, friedman2002stochastic} that connects boosting with numerical optimization by performing gradient descent on the loss function with respect to the base learners.

Default preprocessing avoids scaling (trees are scale-invariant). We standardize rate features only for interpretability in sensitivity plots. We use the earliest 80\% (by time) for training, the next 10\% for validation (hyperparameter tuning/early stopping), and the final 10\% for testing. Within each block we stratify by class to preserve prevalence. Stage~II folds are constructed from the subset with realized Stage~I labels to avoid dependency leakage. We combine randomized search over broad ranges (e.g., \texttt{max\_depth}, \texttt{n\_estimators}, \texttt{min\_child\_samples}/\texttt{min\_child\_weight}, \texttt{subsample}, \texttt{colsample\_bytree}, \texttt{learning\_rate}) with Bayesian optimization on the top configurations. Early stopping uses validation macro-F1. We use inverse-frequency class weights in the loss. For sensitivity checks we considered minority upsampling, SMOTE-NC for mixed categorical, numerical data, and synthetic tabular generation (SDV) with conditional sampling. Given the risk of implausible one-hot patterns, we report results from class-weighted training as the primary configuration.

Two additional training strategies were employed. For hyperparameter tuning, we combined randomized search (efficient in high-dimensional spaces but without learning across iterations) with Bayesian optimization (which learns from previous evaluations to focus on promising regions at higher computational cost). To address class imbalance, we experimented with SMOTE, which generates synthetic minority samples via interpolation, and with Synthetic Data Vault (SDV), which generates complete synthetic datasets conditioned on the statistical properties of the real data. SDV’s advantage over SMOTE is that it permits conditional constraints on the generated samples, avoiding implausible feature combinations in one-hot encoded spaces.

\subsection{Evaluation Metrics}
\subsubsection{Accuracy}
Overall accuracy is defined in \Cref{eq:acc}, where $N_{c}$ is the number of correctly predicted labels and $N_{t}$ is the total number of predictions.

\begin{equation}
\label{eq:acc}
\text{Accuracy} \;=\; \frac{N_c}{N_t},
\end{equation}

\subsubsection{ROC and AUC}
We compute one-vs-rest ROC curves and macro-averaged AUCs. Because precision--recall (PR) curves are more informative under class skew, we also report macro PR--AUC and plot per-class PR curves. The ROC curve plots the False Positive Rate (FPR, \Cref{eq:FPR}) against the True Positive Rate (TPR, \Cref{eq:TPR}) as the classification threshold varies from 0 to 1 (\Cref{eq:DT}). For multiclass problems we adopt the one-vs-rest formulation, treating each class in turn as positive and all others as negative. The area under the ROC curve (AUC) ranges from 0.5 (random) to 1 (perfect ranking).

\begin{equation}
\label{eq:FPR}
\text{FPR} = \frac{FP}{FP+TN},\qquad
\end{equation}

\begin{equation}
\label{eq:TPR}
\text{TPR}=\text{Recall}=\frac{TP}{TP+FN}.
\end{equation}

\begin{equation}
\label{eq:DT}
\begin{aligned}
P(N_{\text{sample}}) &\geq DT, \\
P(N_{\text{sample}}) &< DT
\end{aligned}
\end{equation}

\begin{figure}[H]
    \centering
    \includegraphics[scale=0.5]{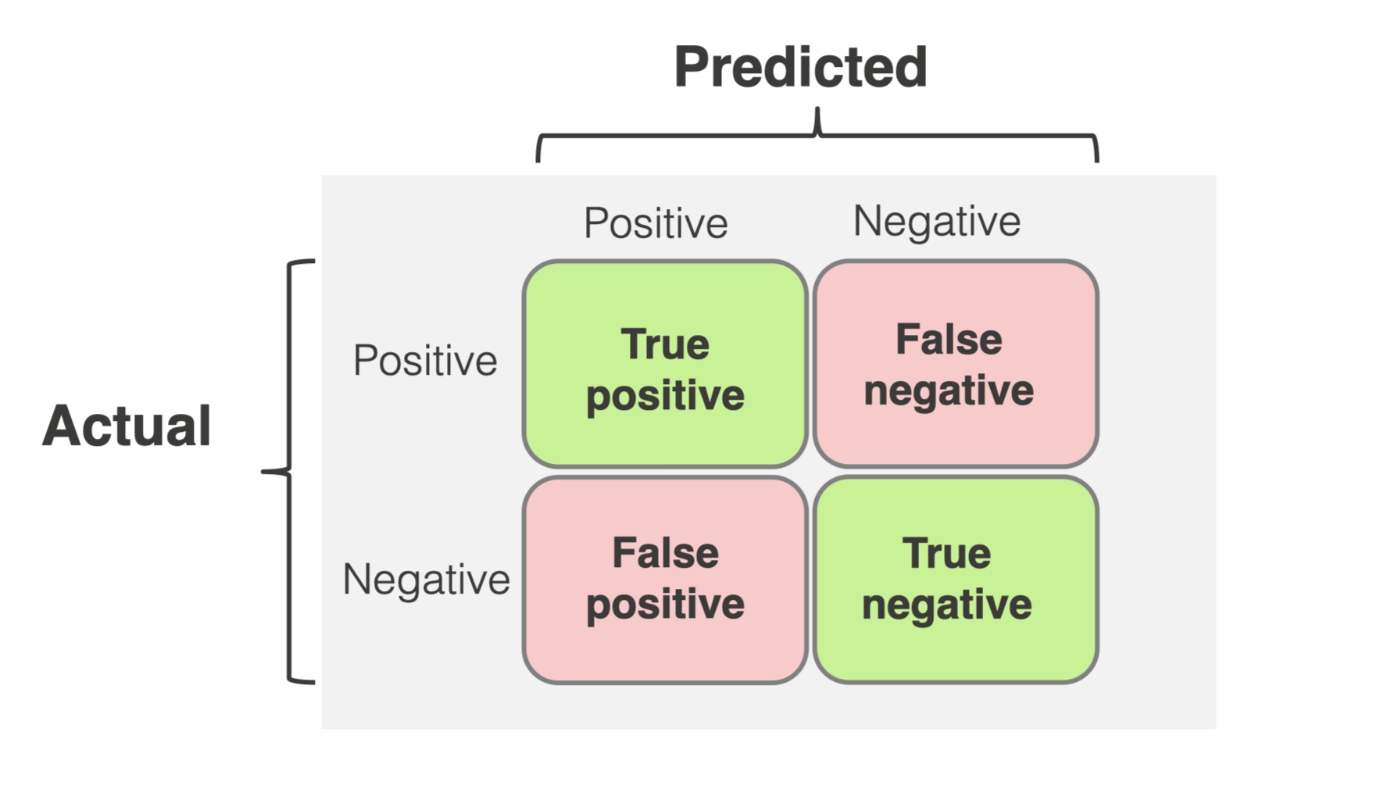}
     \caption{Simple Confusion Matrix}
    \label{fig:SIMPLECM}
\end{figure}

\begin{figure}[H]
    \centering
    \includegraphics[scale=0.5]{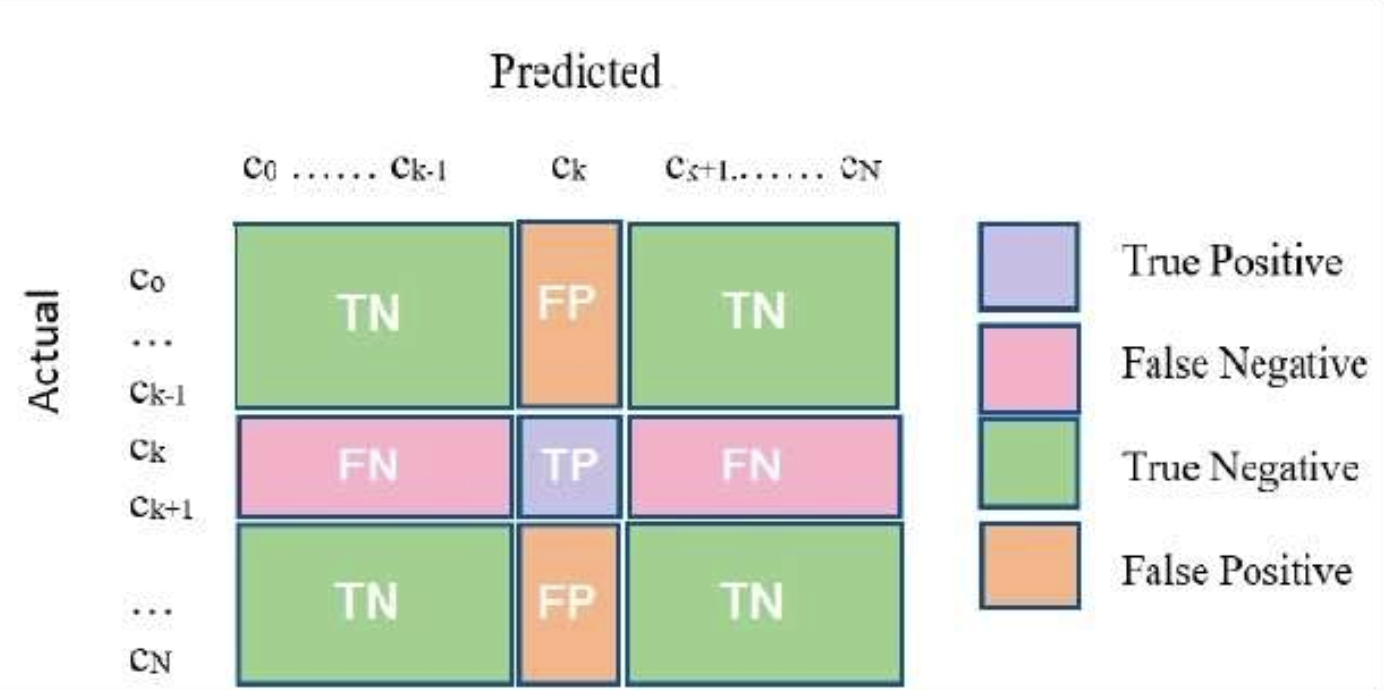}
     \caption{Multiclass Confusion Matrix}
    \label{fig:MULTI_CM_EX}
\end{figure}

\subsubsection{Precision--Recall}
The precision--recall curve plots precision (\Cref{eq:precision}) against recall (identical to TPR) as the classification threshold varies. Unlike the ROC curve, the PR curve is sensitive to class prevalence, making it the more informative diagnostic when classes are heavily imbalanced.
\begin{equation}
\label{eq:precision}
Precision =  \frac{TP}{TP+FP}
\end{equation}

\subsubsection{Confusion Matrix}
The confusion matrix tabulates true labels against predicted labels, making it straightforward to identify which classes are confused with one another. Binary and multiclass examples are shown in \Cref{fig:SIMPLECM} and \Cref{fig:MULTI_CM_EX}.

\subsection{Feature Importance}
We quantify feature influence using SHAP values. For each stage we report global importance as the mean absolute SHAP value per feature, SHAP dependence plots for the top covariates (e.g., approach speed, departure and crossing rates, ramp destination), and, where informative, SHAP interaction values (e.g., exit $\times$ ramp in Stage~II). To ensure stability, importance is averaged over 20 bootstrap refits. As complementary checks, we compute permutation importance on the test set and partial-dependence/ICE curves for selected features.

\section{Results and Discussions\label{sec: discussion}}
After preprocessing and filtering, the final dataset comprised 72,439 east-flow samples and 109,389 west-flow samples. Both subsets exhibit substantial class imbalance (\Cref{tab:stage1datasplit}-\Cref{tab:stage2datasplit}). For Stage~I (runway exit), a single exit dominates in each flow (B11 in east; B3 in west), and for Stage~II (cross vs.\ specific crossing points vs.\ end-around) several classes are rare (e.g., east E1/E3/E5; west C). To preserve temporal realism and prevalence, the data were split 80\%/10\%/10\% (train/validation/test) with stratification. Class weights were employed during training to mitigate skew without risking implausible one-hot combinations in the feature space.

\subsection{Stage I Results}

\begin{table}[H]
    \centering
    \caption{Class split of Stage~I for West Flow and East Flow arrivals.}
    \label{tab:stage1datasplit}
    \small
    \begin{tabular}{lc|lc}
        \toprule
        \multicolumn{2}{c|}{West Flow} & \multicolumn{2}{c}{East Flow} \\
        \cmidrule(r){1-2} \cmidrule(l){3-4}
        Class & \% & Class & \% \\
        \midrule
        \textbf{B1}  & 0.021 & \textbf{B7}  & 0.096 \\
        \textbf{B3}  & 0.833 & \textbf{B11} & 0.884 \\
        \textbf{B5}  & 0.146 & \textbf{B13} & 0.018 \\
        \bottomrule
    \end{tabular}
\end{table}

\begin{figure}[H]
    \centering
    \begin{subfigure}[b]{0.45\textwidth}
        \centering
        \includegraphics[scale=0.45]{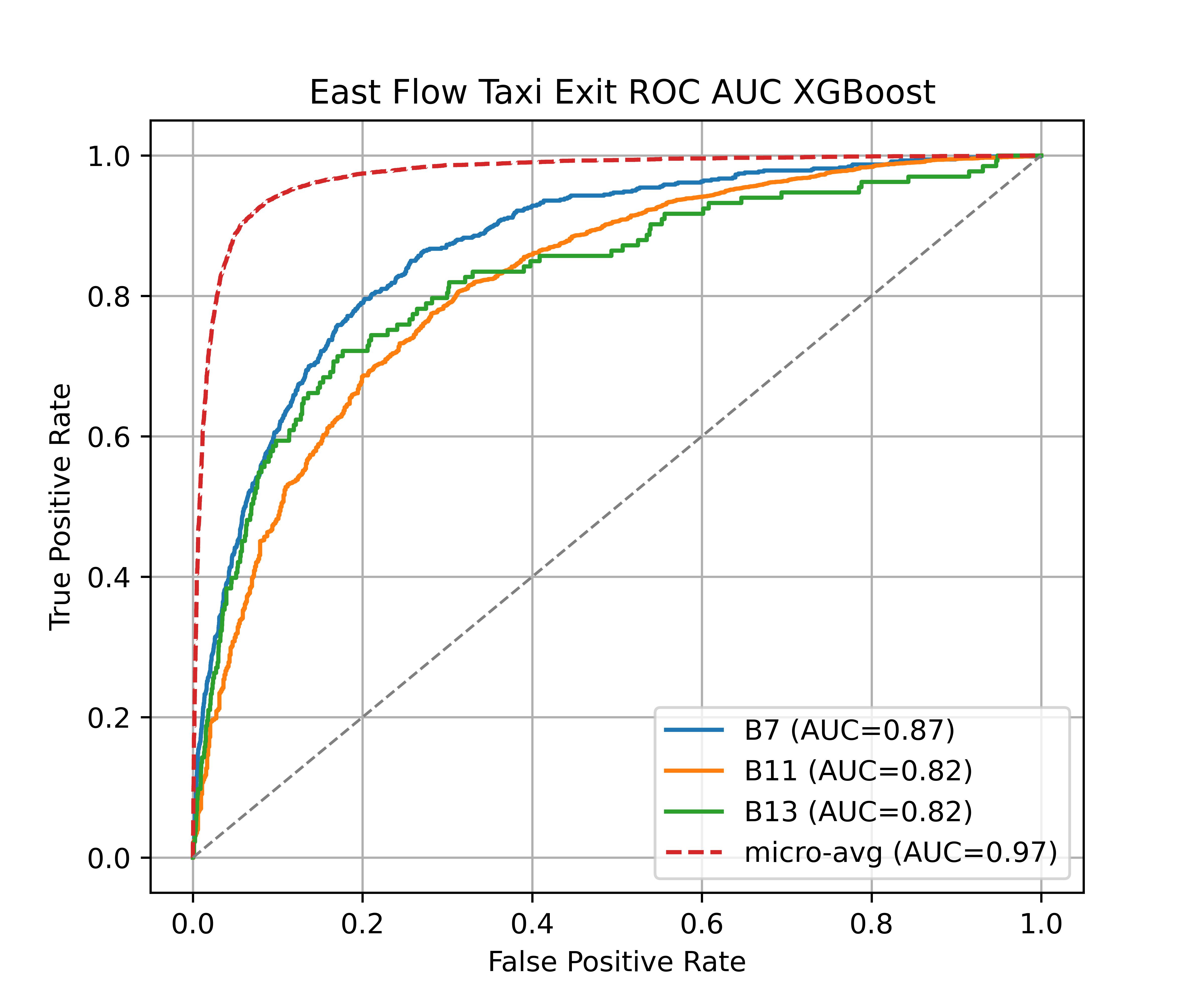}
        \caption{East XGBoost}
        \label{fig:EASTXGSTEP1}
    \end{subfigure}
    \begin{subfigure}[b]{0.45\textwidth}
        \centering
        \includegraphics[scale=0.45]{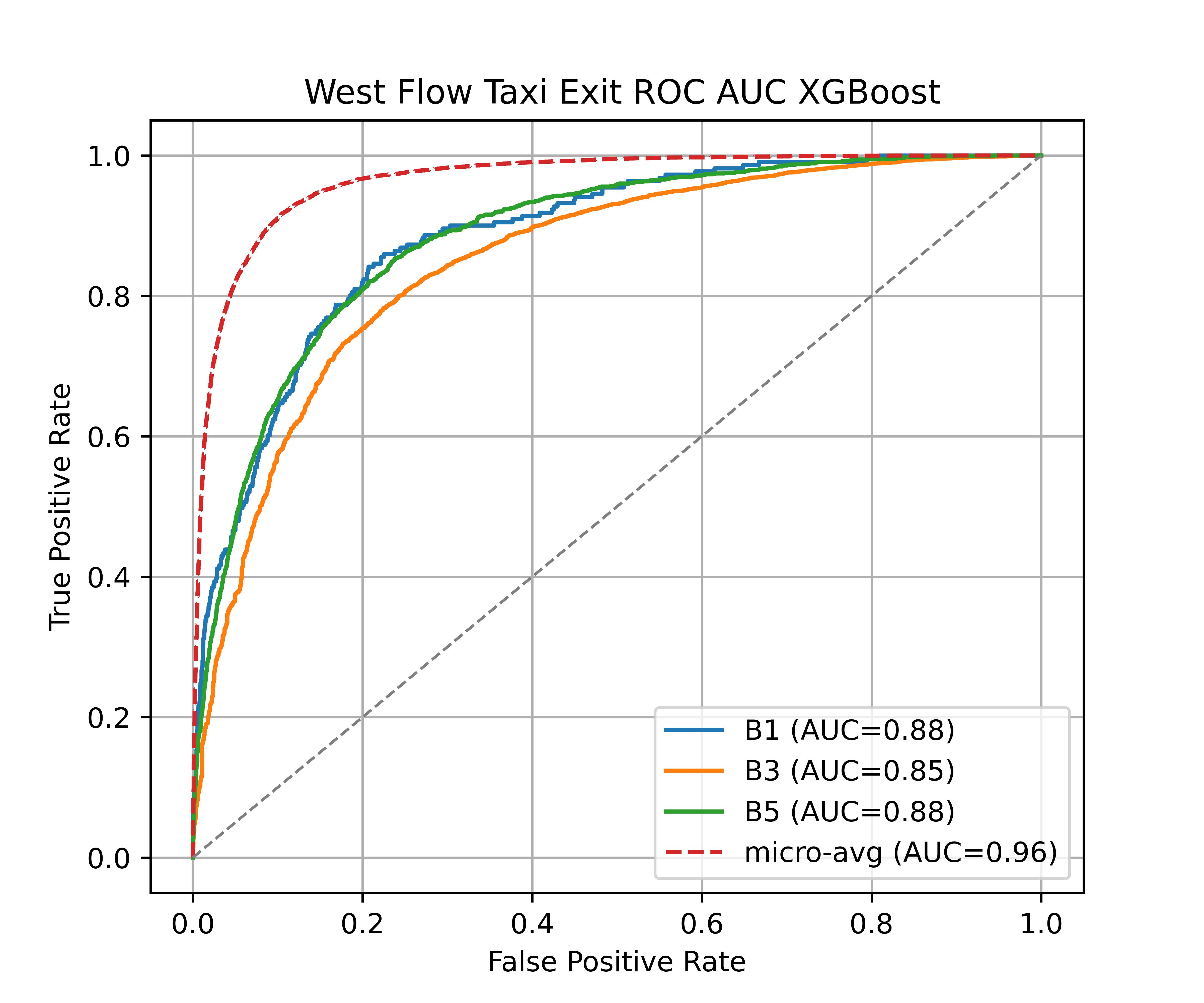}
        \caption{West XGBoost}
        \label{fig:WESTXGSTEP1}
    \end{subfigure}
    
    \begin{subfigure}[b]{0.45\textwidth}
        \centering
        \includegraphics[scale=0.45]{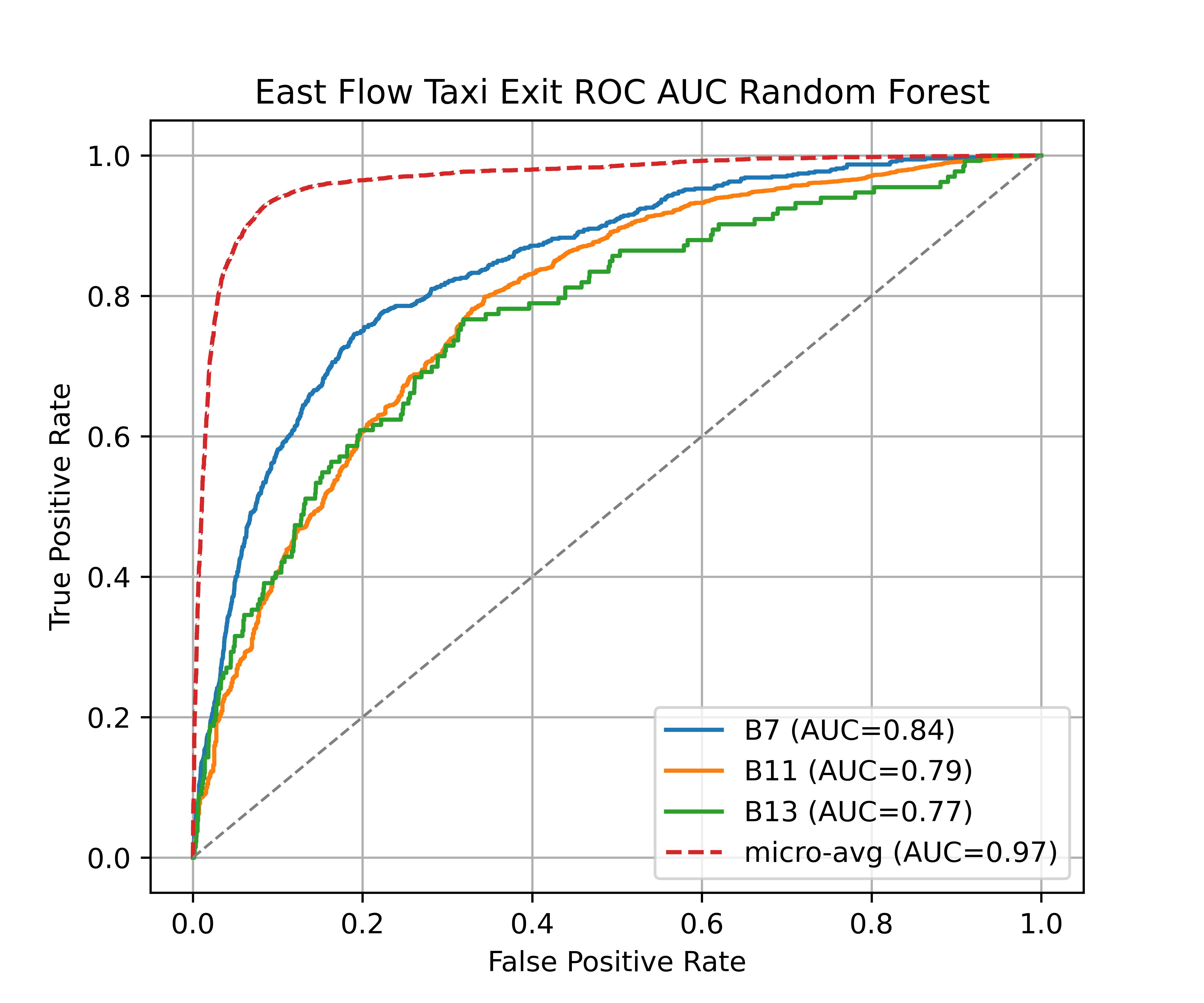}
        \caption{East Random Forest}
        \label{fig:EASTRFSTEP1}
    \end{subfigure}
    \begin{subfigure}[b]{0.45\textwidth}
        \centering
        \includegraphics[scale=0.45]{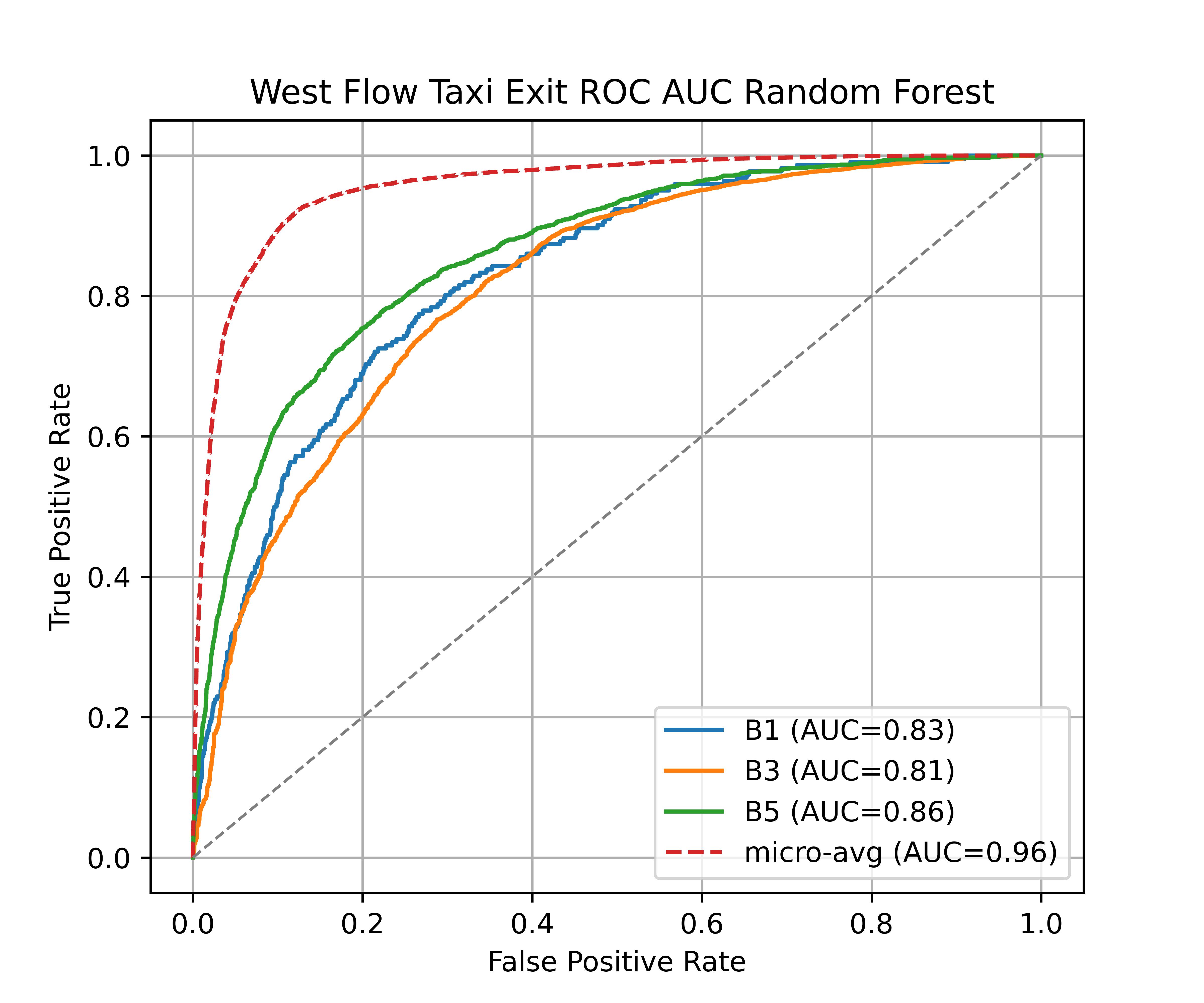}
        \caption{West Random Forest}
        \label{fig:westrfstep1}
    \end{subfigure}
    
    \begin{subfigure}[b]{0.45\textwidth}
        \centering
        \includegraphics[scale=0.45]{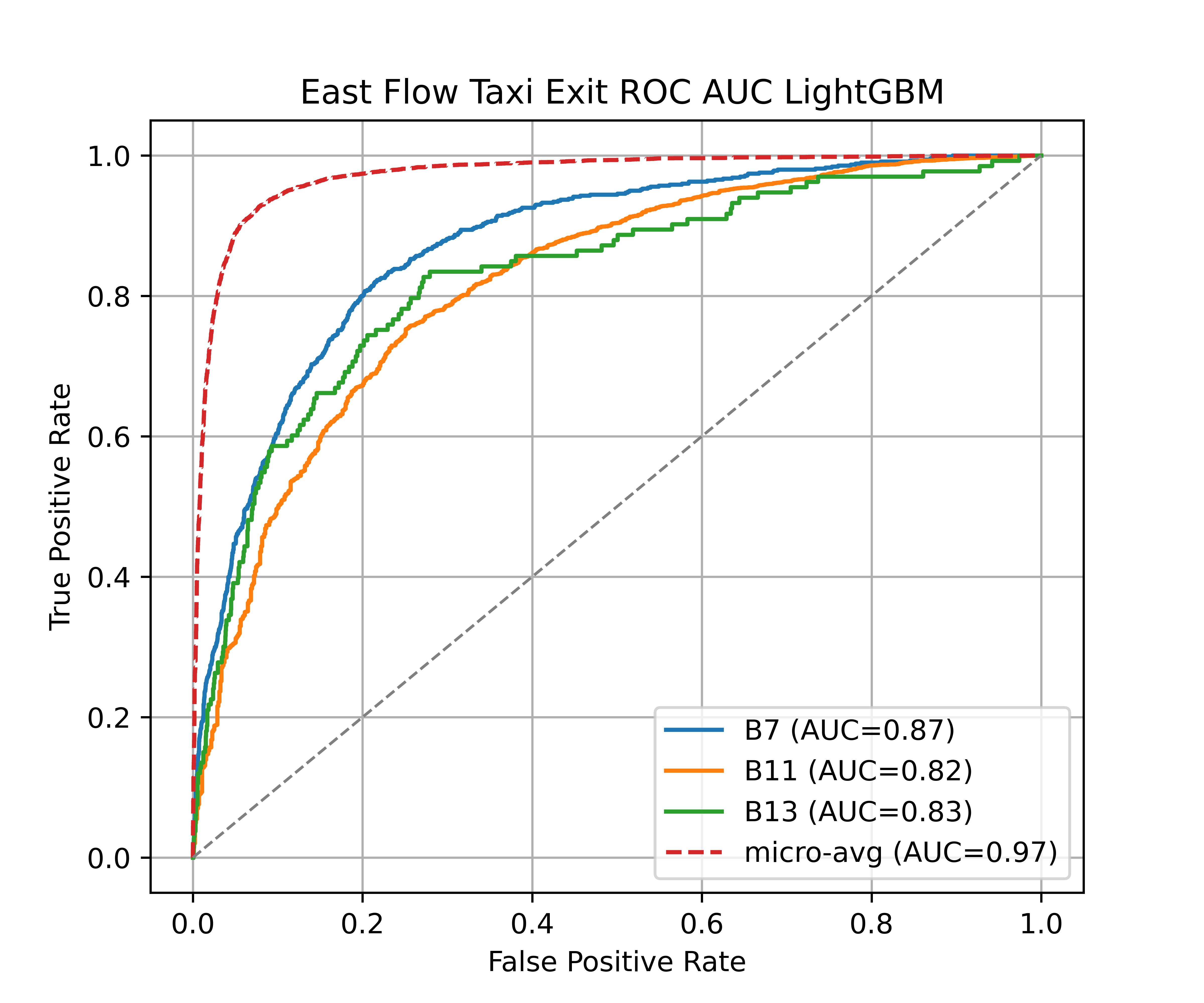}
        \caption{East LightGBM}
        \label{fig:eastgbmstep1}
    \end{subfigure}
    \begin{subfigure}[b]{0.45\textwidth}
        \centering
        \includegraphics[scale=0.45]{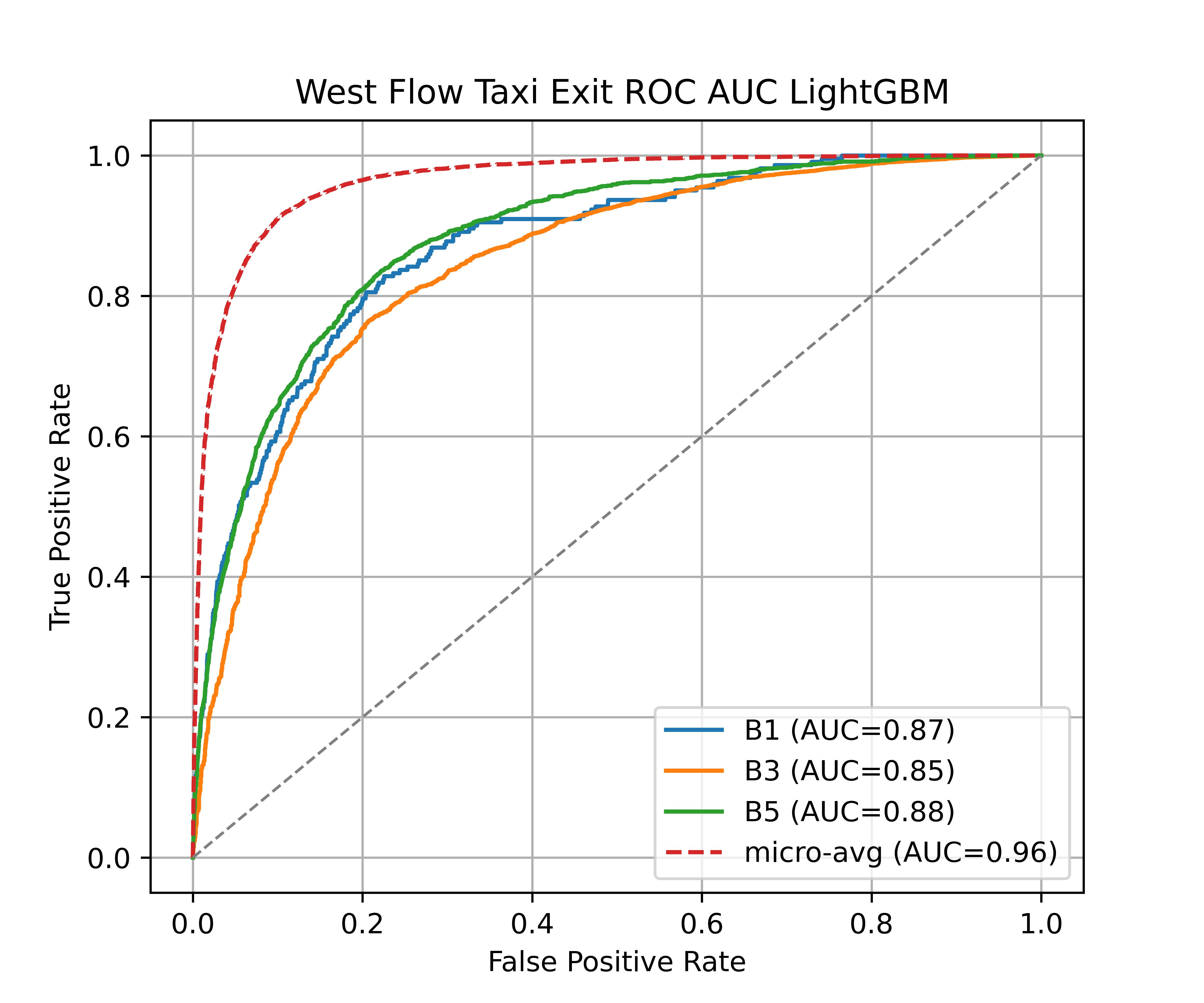}
        \caption{West LightGBM}
        \label{fig:westgbmstep1}
    \end{subfigure}

    \caption{ROC AUC Plots Stage I}
    \label{fig:ROCAUCSTAGE1}
\end{figure}

\begin{figure}[H]
    \centering
    \begin{subfigure}[b]{0.45\textwidth}
        \centering
        \includegraphics[scale=0.45]{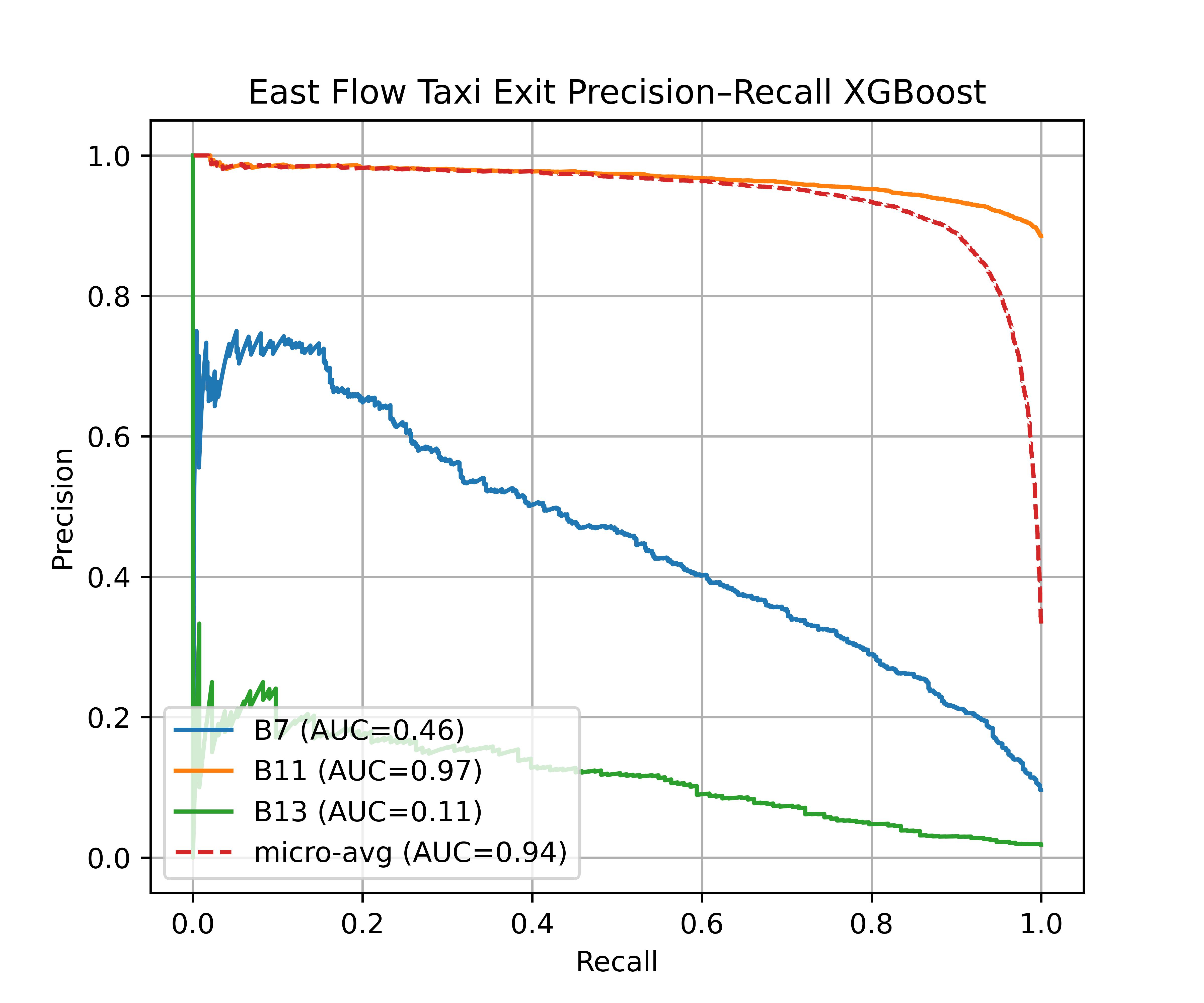}
        \caption{East XGBoost}
        \label{fig:EASTXGSTEP1PR}
    \end{subfigure}
    \begin{subfigure}[b]{0.45\textwidth}
        \centering
        \includegraphics[scale=0.45]{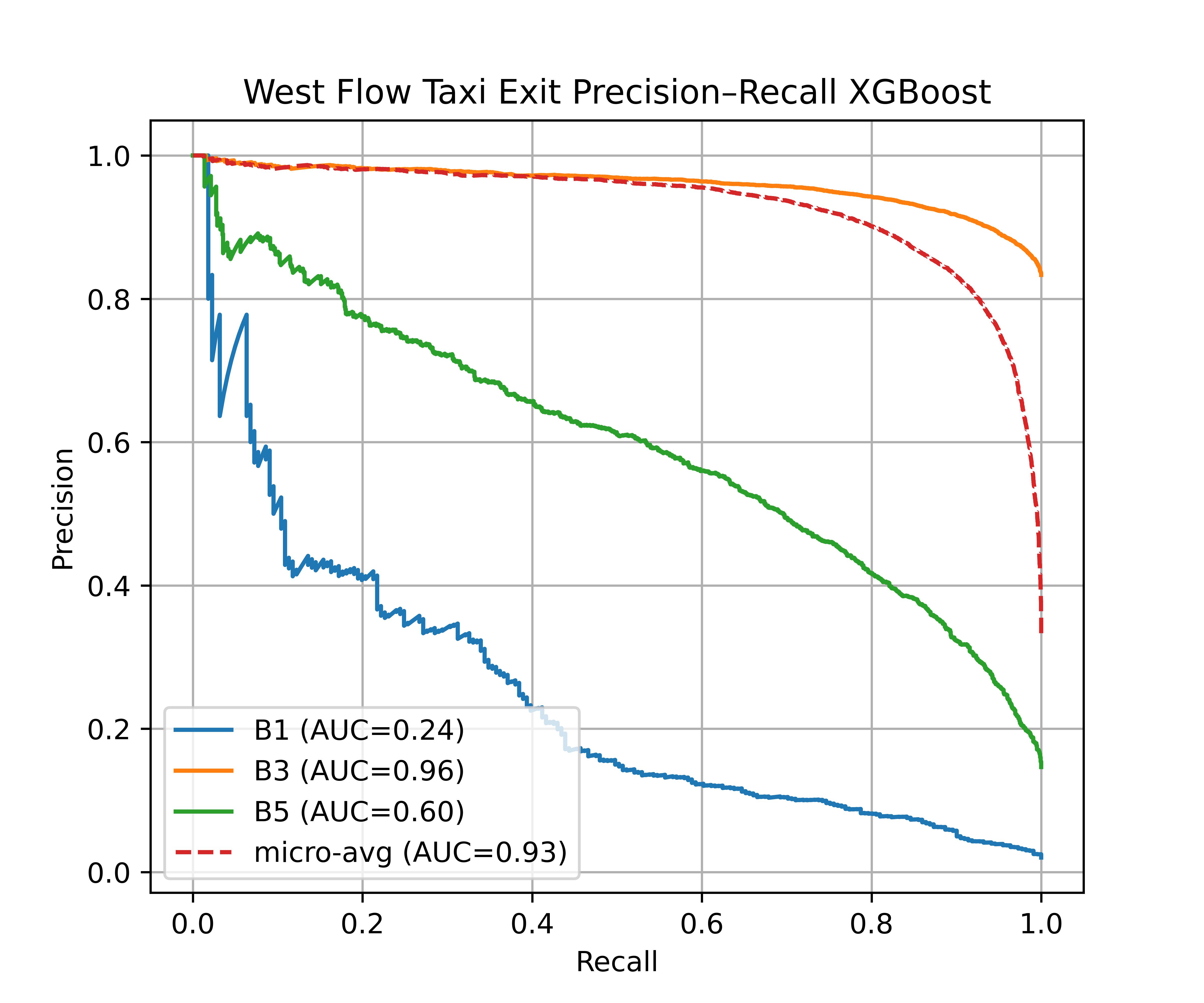}
        \caption{West XGBoost}
        \label{fig:WESTXGSTEP1PR}
    \end{subfigure}
    
    \begin{subfigure}[b]{0.45\textwidth}
        \centering
        \includegraphics[scale=0.45]{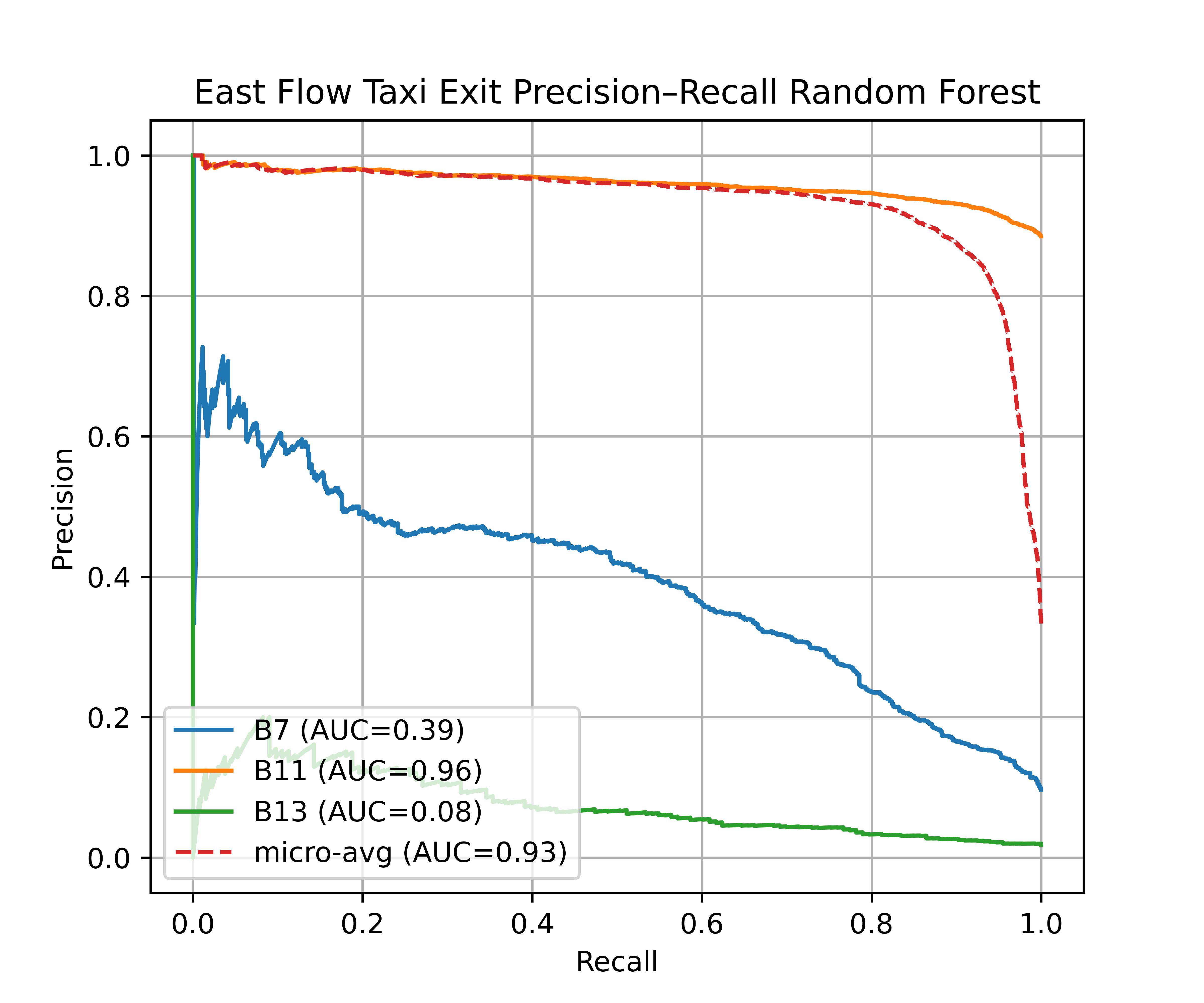}
        \caption{East Random Forest}
        \label{fig:EASTRFSTEP1PR}
    \end{subfigure}
    \begin{subfigure}[b]{0.45\textwidth}
        \centering
        \includegraphics[scale=0.45]{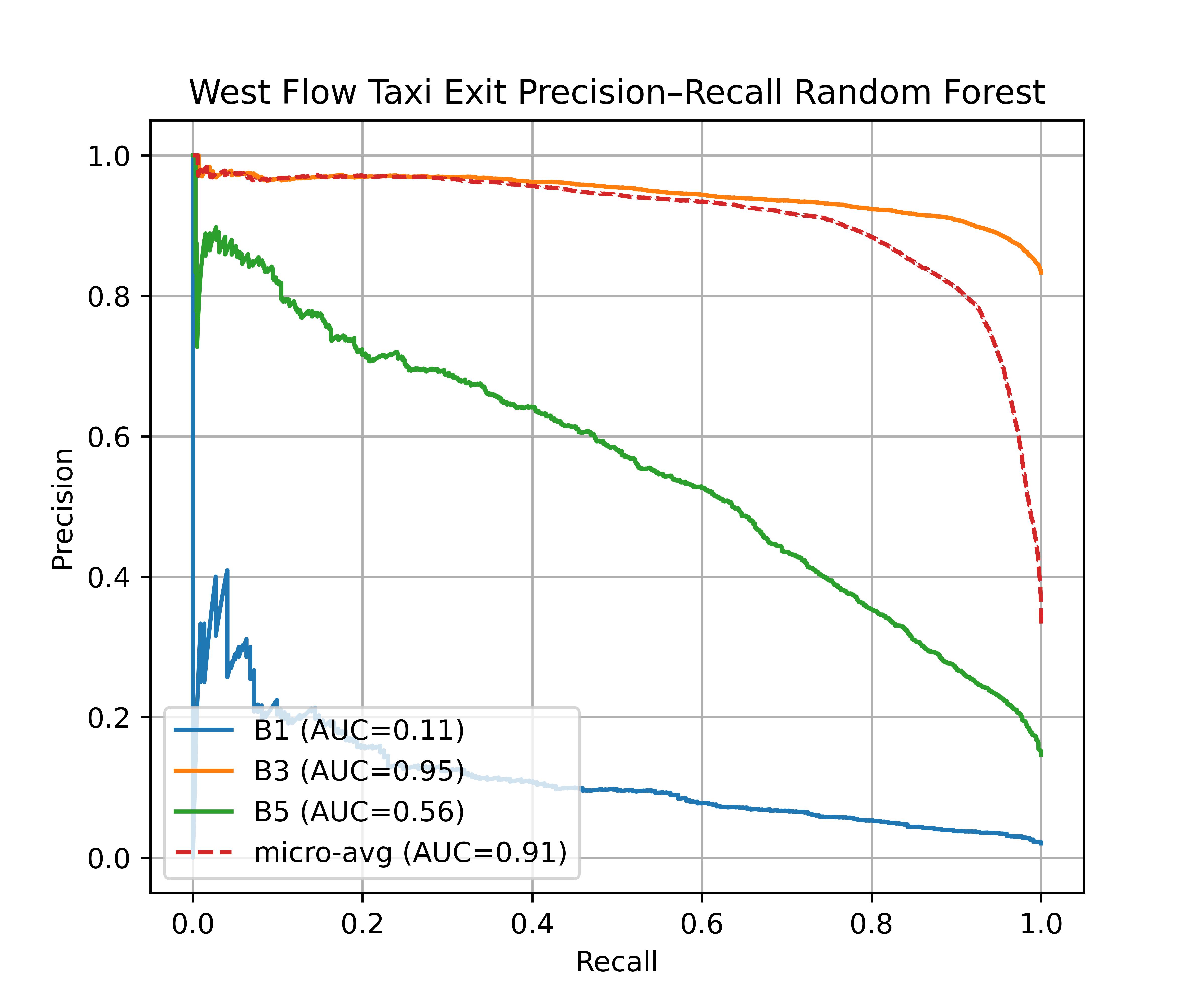}
        \caption{West Random Forest}
        \label{fig:westrfstep1PR}
    \end{subfigure}
    
    \begin{subfigure}[b]{0.45\textwidth}
        \centering
        \includegraphics[scale=0.45]{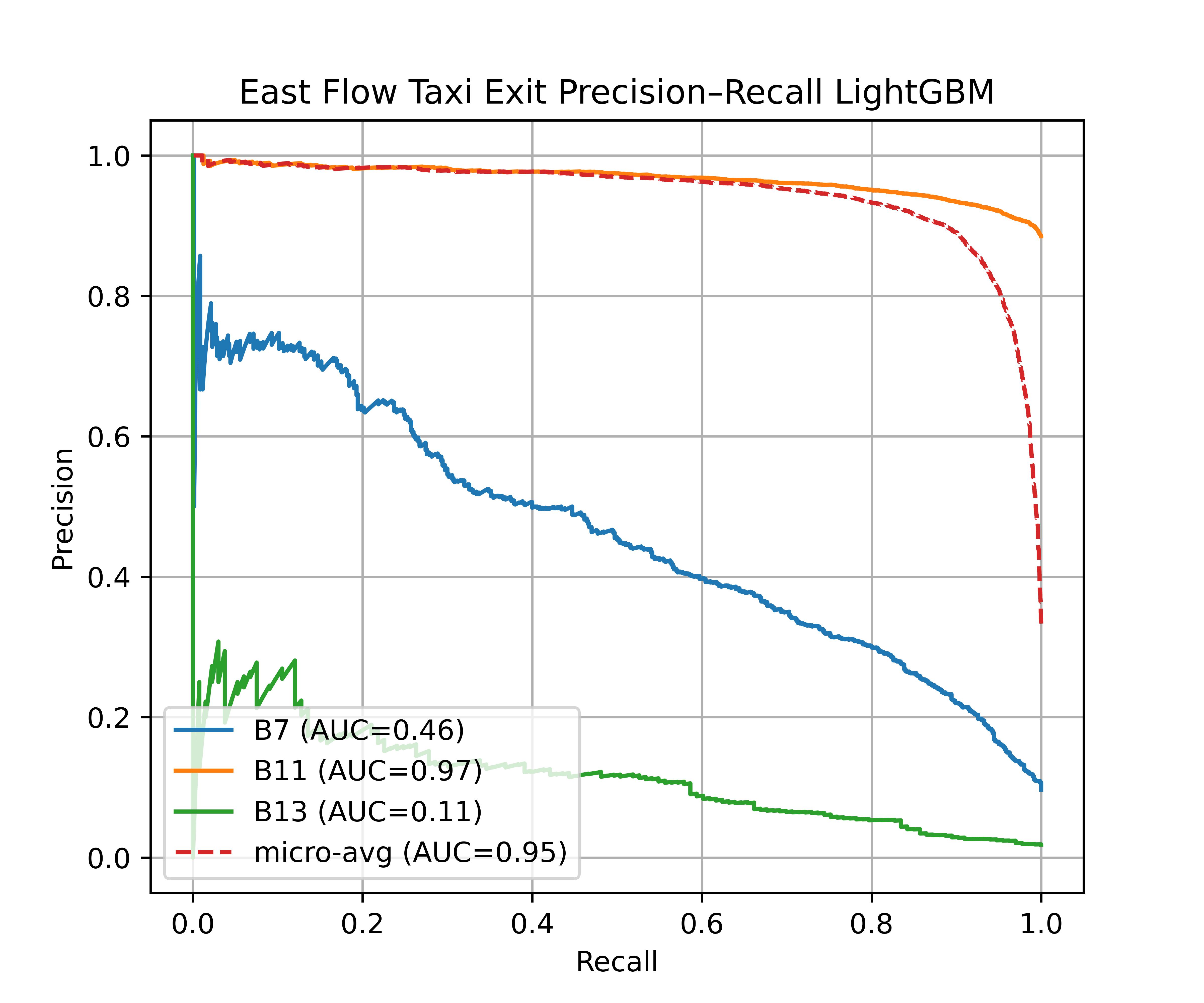}
        \caption{East LightGBM}
        \label{fig:eastgbmstep1PR}
    \end{subfigure}
    \begin{subfigure}[b]{0.45\textwidth}
        \centering
        \includegraphics[scale=0.45]{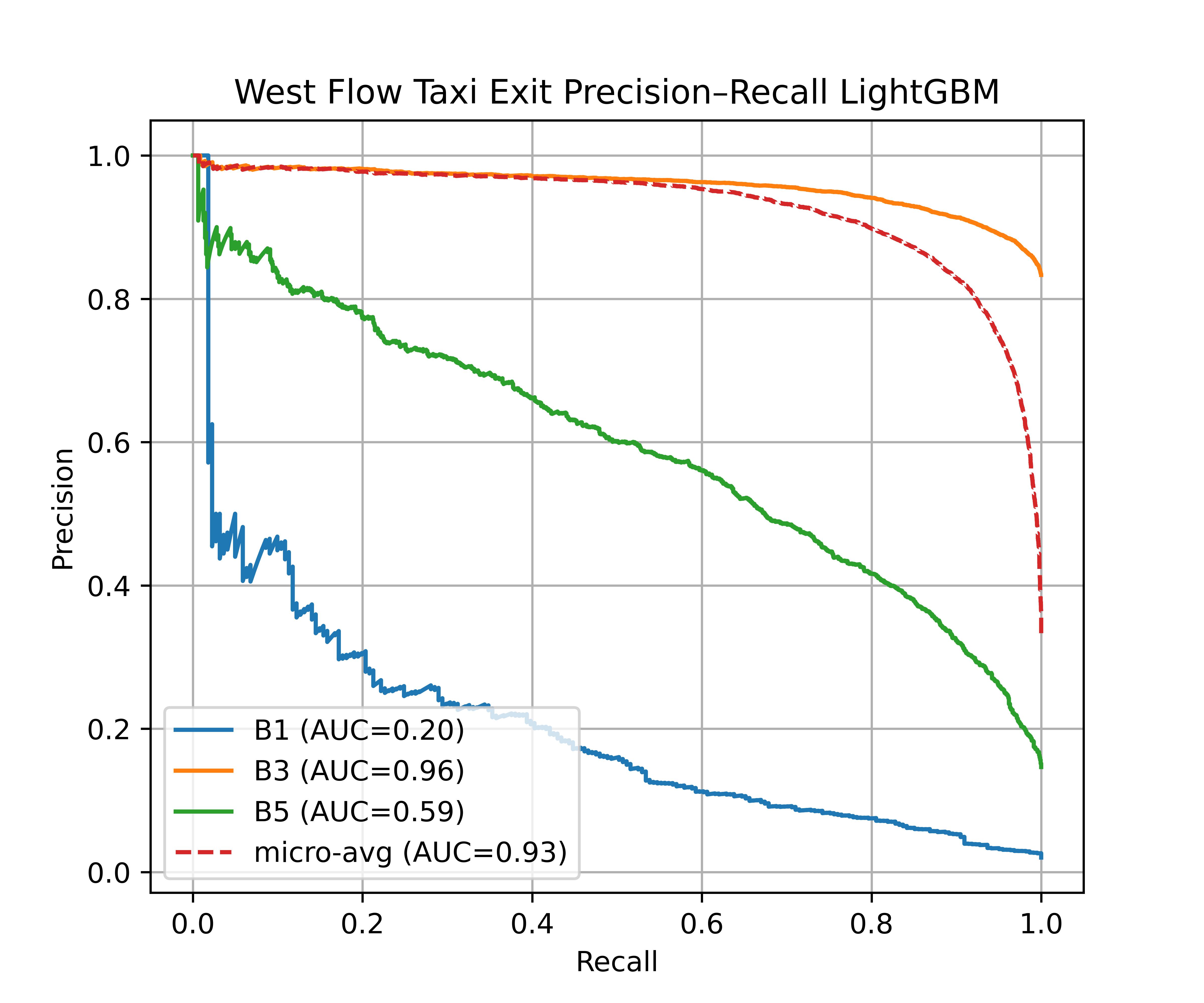}
        \caption{West LightGBM}
        \label{fig:westgbmstep1PR}
    \end{subfigure}

    \caption{Precision Recall Plots Stage I}
    \label{fig:PRSTAGE1}
\end{figure}

\begin{figure}[H]
    \centering
    \begin{subfigure}[b]{0.45\textwidth}
        \centering
        \includegraphics[scale=0.15]{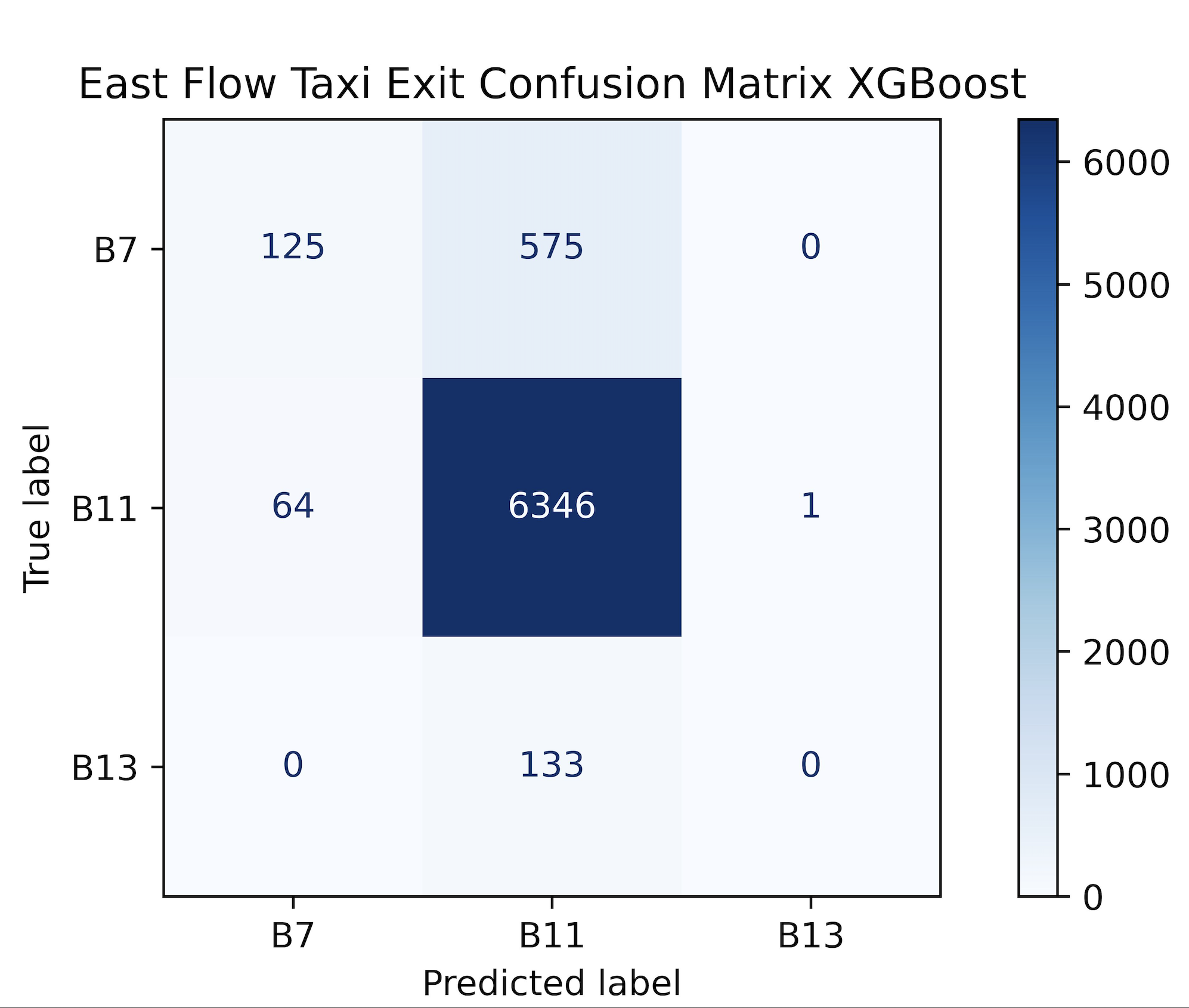}
        \caption{East XGBoost}
        \label{fig:EASTXGSTEP1CM}
    \end{subfigure}
    \begin{subfigure}[b]{0.45\textwidth}
        \centering
        \includegraphics[scale=0.15]{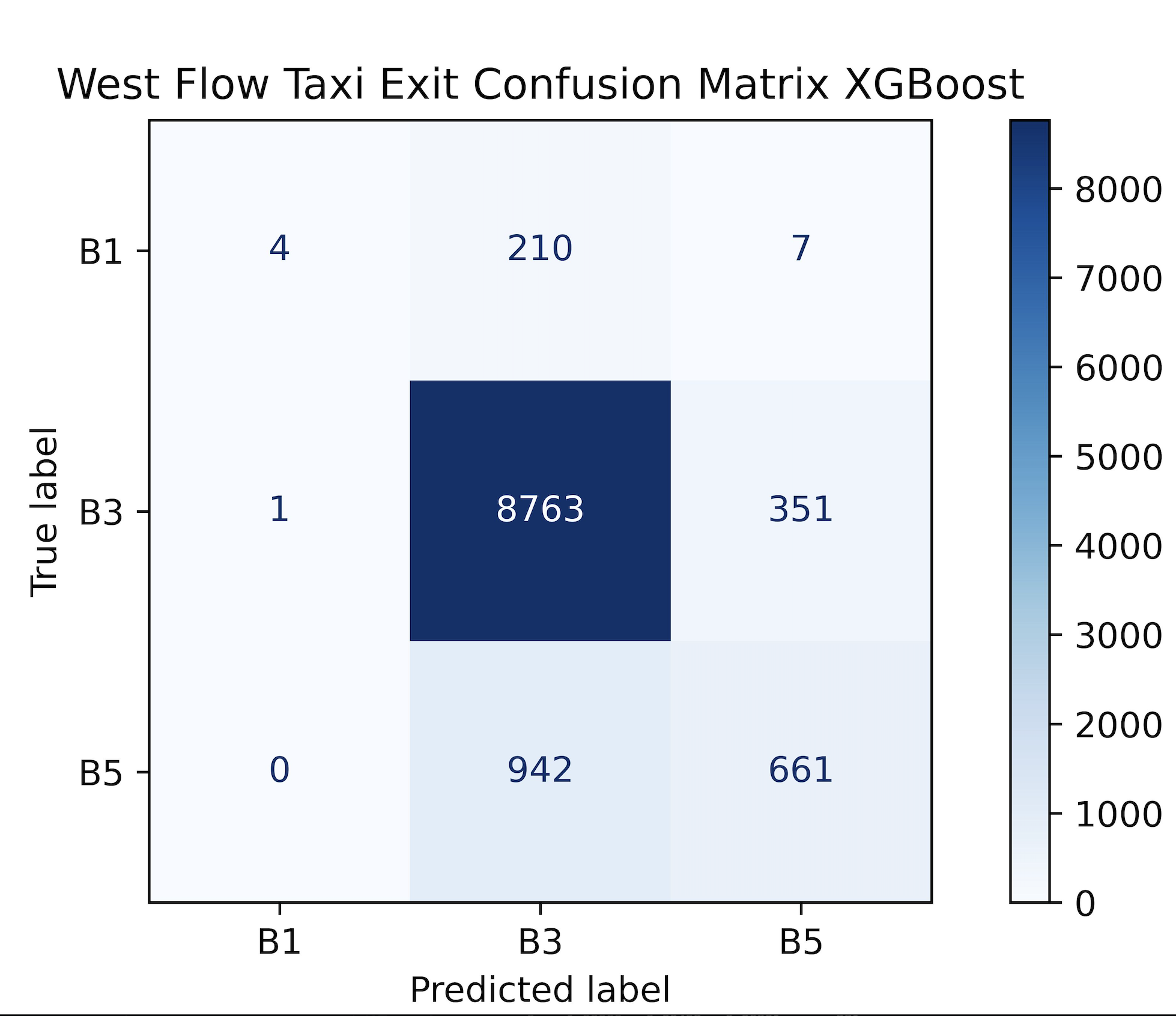}
        \caption{West XGBoost}
        \label{fig:WESTXGSTEP1CM}
    \end{subfigure}
    
    \begin{subfigure}[b]{0.45\textwidth}
        \centering
        \includegraphics[scale=0.15]{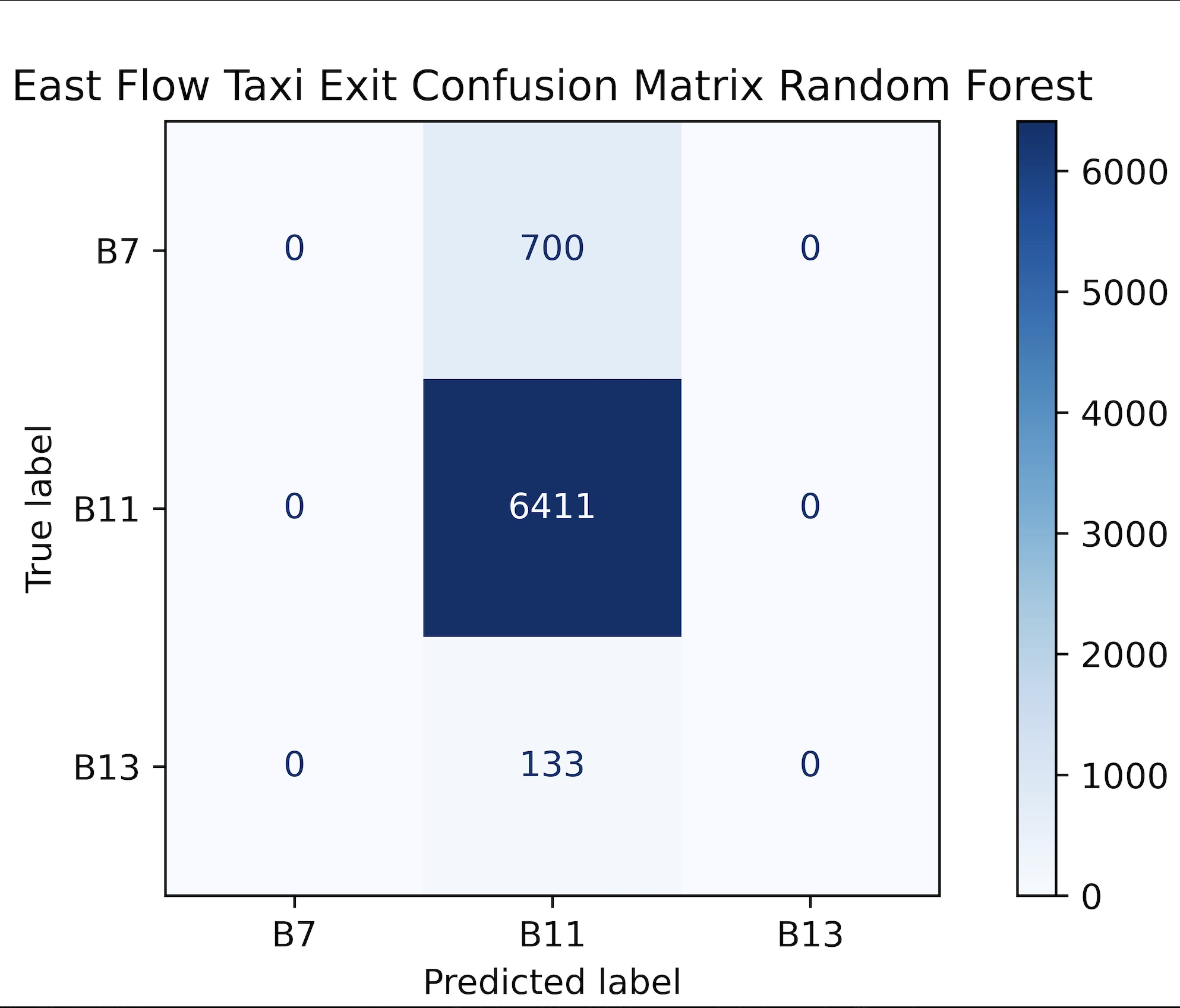}
        \caption{East Random Forest}
        \label{fig:EASTRFSTEP1CM}
    \end{subfigure}
    \begin{subfigure}[b]{0.45\textwidth}
        \centering
        \includegraphics[scale=0.15]{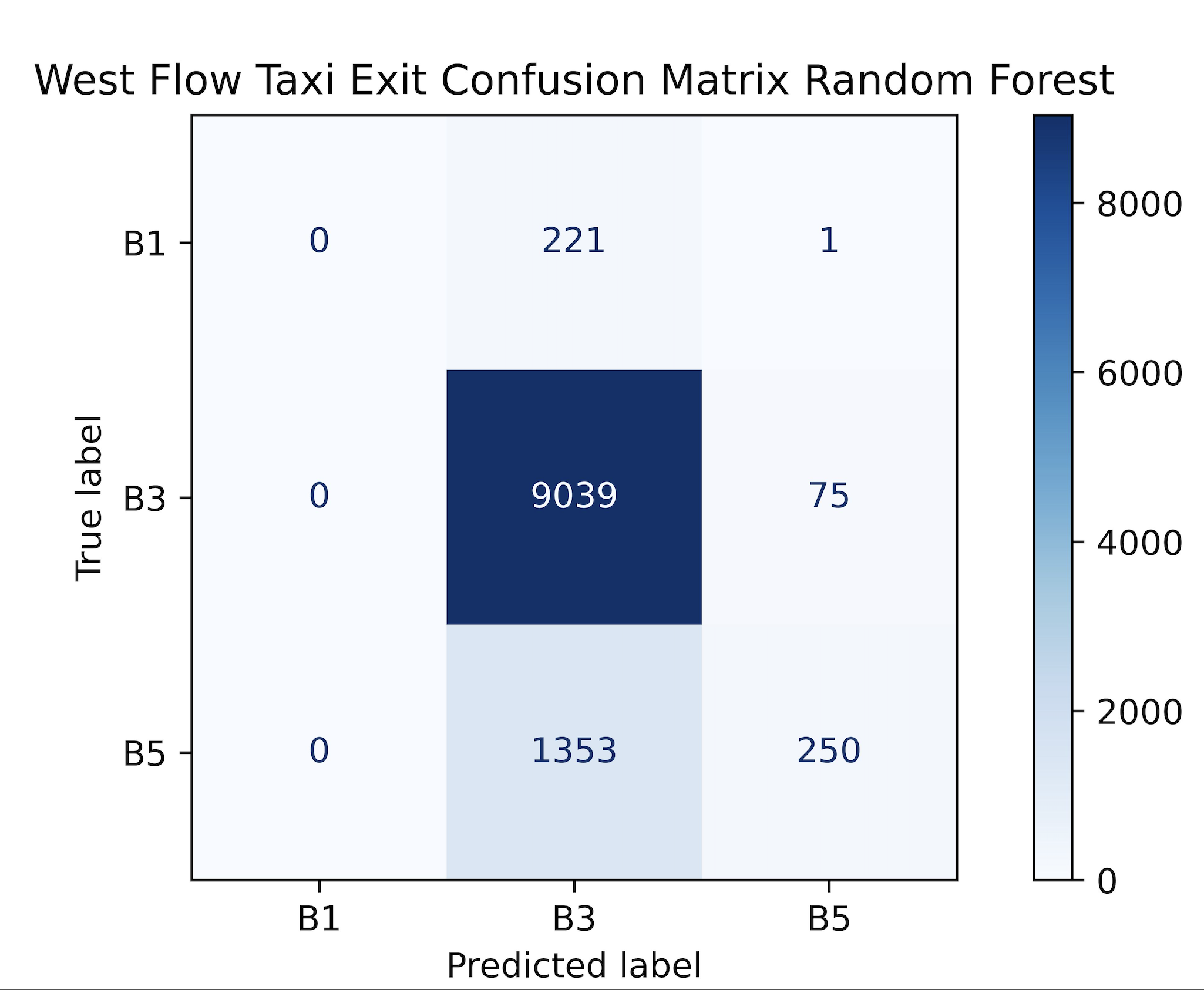}
        \caption{West Random Forest}
        \label{fig:westrfstep1CM}
    \end{subfigure}
    
    \begin{subfigure}[b]{0.45\textwidth}
        \centering
        \includegraphics[scale=0.15]{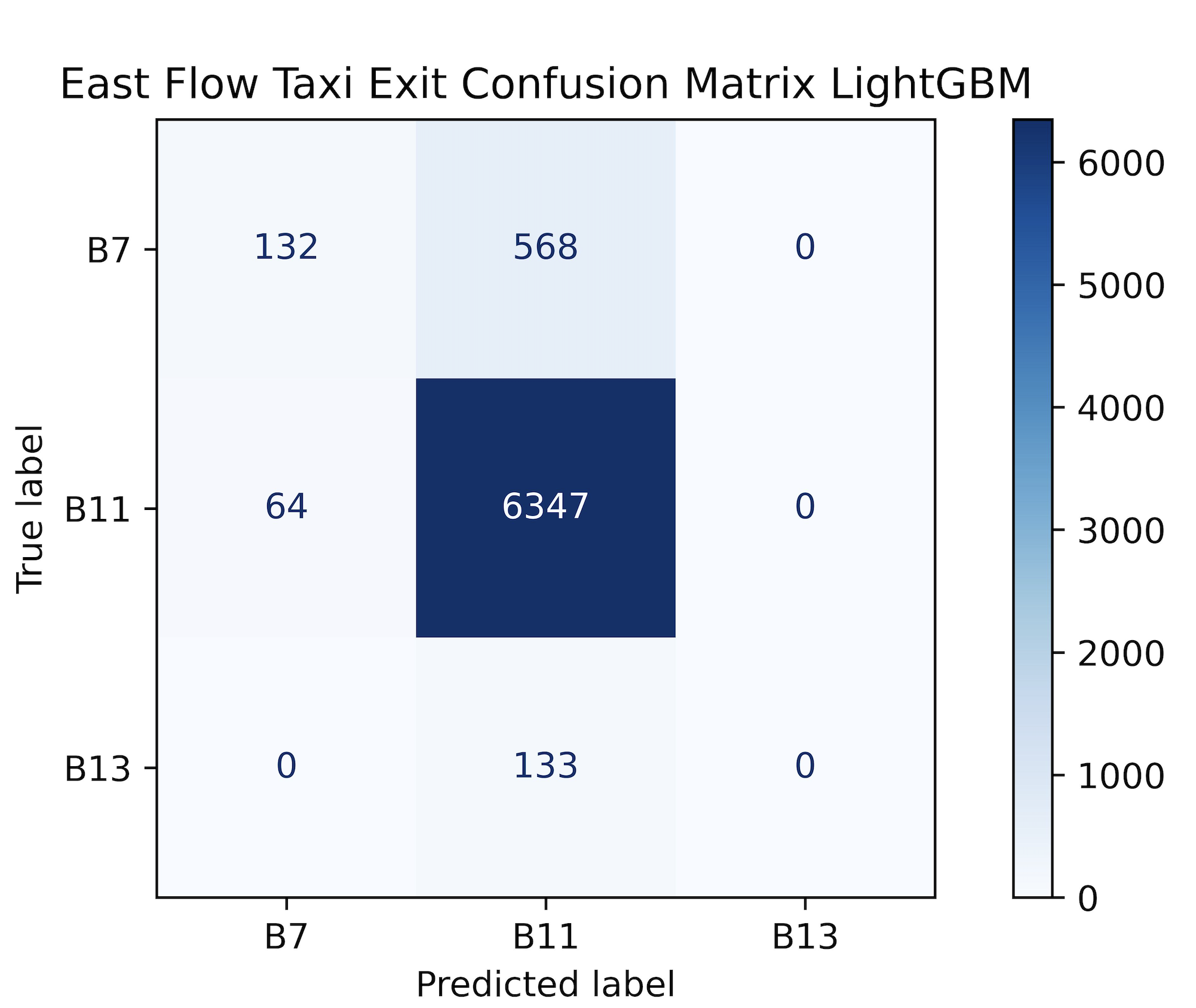}
        \caption{East LightGBM}
        \label{fig:eastgbmstep1CM}
    \end{subfigure}
    \begin{subfigure}[b]{0.45\textwidth}
        \centering
        \includegraphics[scale=0.15]{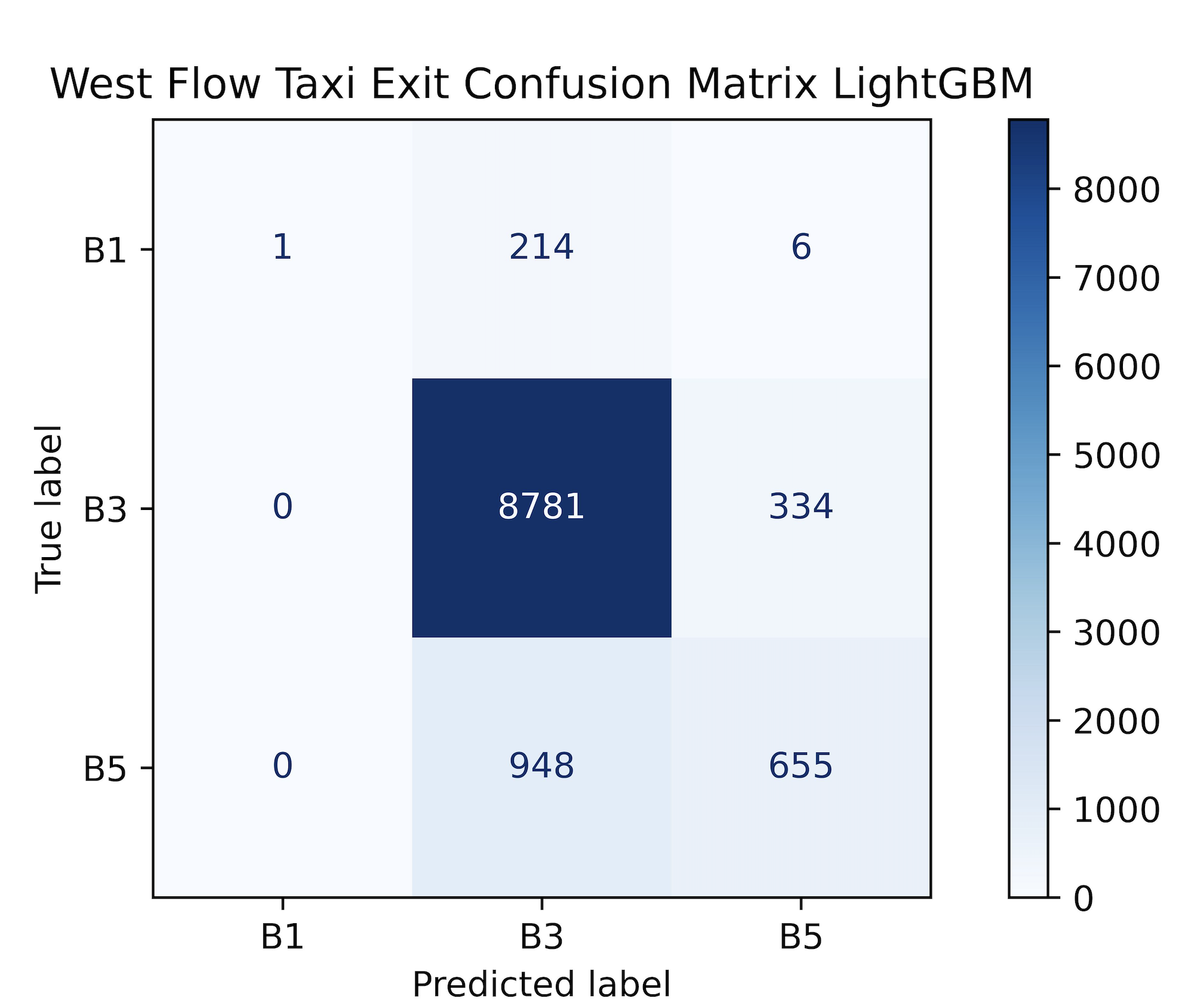}
        \caption{West LightGBM}
        \label{fig:westgbmstep1CM}
    \end{subfigure}

    \caption{Confusion Matrices Stage I}
    \label{fig:CMSTAGE1}
\end{figure}

\Cref{tab:STAGE1_MODEL COMPARISON} summarizes the results for the runway exit prediction stage. Gradient-boosted ensembles (LightGBM, XGBoost) achieve the highest accuracies (0.86--0.89) and macro-F1 (0.41--0.50), outperforming Random Forest across flows. However, macro-averaged recall and F1 remain below 0.5 for all models, reflecting a concentration of correct predictions for majority exits and difficulty predicting for minority exits. ROC curves exceed 0.70 AUC for most classes (\Cref{fig:ROCAUCSTAGE1}), indicating good ranking ability. The precision--recall (PR) curves in \Cref{fig:PRSTAGE1} reveal the more operationally relevant story under skew: high PR-AUC for the dominant exit, markedly lower PR-AUC for rare exits. The confusion matrices (\Cref{fig:CMSTAGE1}) visualize these error modes as mass shifting toward the majority exit, with near-miss errors typically impacting adjacent exits rather than arbitrary labels.

\begin{table}[H]
    \centering
    \caption{Stage I: Taxi Exit Model Comparison, each evaluation metric is the average of the multiclass result}
    \label{tab:STAGE1_MODEL COMPARISON}
    \begin{tabular}{llcccc}
        \toprule
        & & Precision & Recall & F1-Score & Accuracy \\
        \midrule
        \multirow{3}{*}{West Flow}
            & \textbf{XGBoost}  & 0.777 & 0.464 & 0.487 & 0.862  \\
            & \textbf{Random Forest}  & 0.540 & 0.382 & 0.392 & 0.849 \\
            & \textbf{LightGBM}  & 0.847 & 0.409 & 0.504 & 0.863  \\
        \addlinespace
        \multirow{3}{*}{East Flow}
            & \textbf{XGBoost}  & 0.520 & 0.390 & 0.408 & 0.893  \\
            & \textbf{Random Forest} & 0.295 & 0.333 & 0.313 & 0.885 \\
            & \textbf{LightGBM} & 0.525 & 0.393 & 0.413 & 0.894  \\
        \bottomrule
    \end{tabular}
\end{table}

LightGBM leads by a small macro-F1 margin. SHAP analyses (\Cref{fig:SHAP}, top row) identify approach speed as the dominant global driver in both flows: higher approach/touchdown speeds decrease the likelihood of earlier exits, matching braking-distance intuition. In the east flow, ramp destination, aircraft type/weight, and wind components (head/crosswind) contribute additional signals — consistent with geometry that places some ramps farther from the end-around and increases the value of later exits under certain winds. In the west flow, short-horizon arrival and crossing rates are more prominent, plausibly because west arrivals are geographically closer to the end-around; contemporaneous crossing activity and inbound density provide cues about controller sequencing and the viability of earlier exits.

\subsection{Stage II Results}
\begin{figure}[H]
    \centering
    \begin{subfigure}[b]{0.45\textwidth}
        \centering
        \includegraphics[scale=0.45]{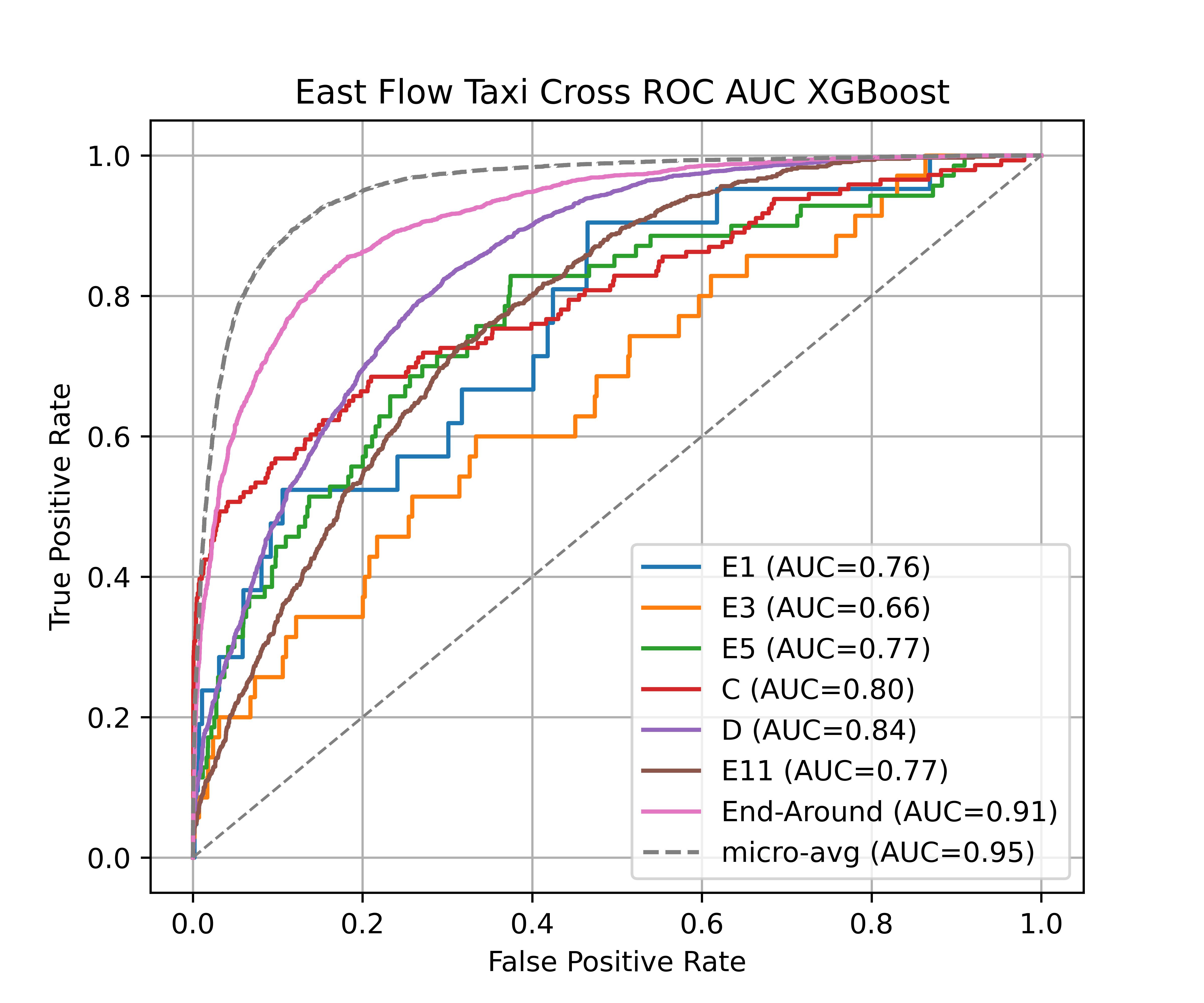}
        \caption{East XGBoost}
        \label{fig:EASTXGSTEP2}
    \end{subfigure}
    \begin{subfigure}[b]{0.45\textwidth}
        \centering
        \includegraphics[scale=0.45]{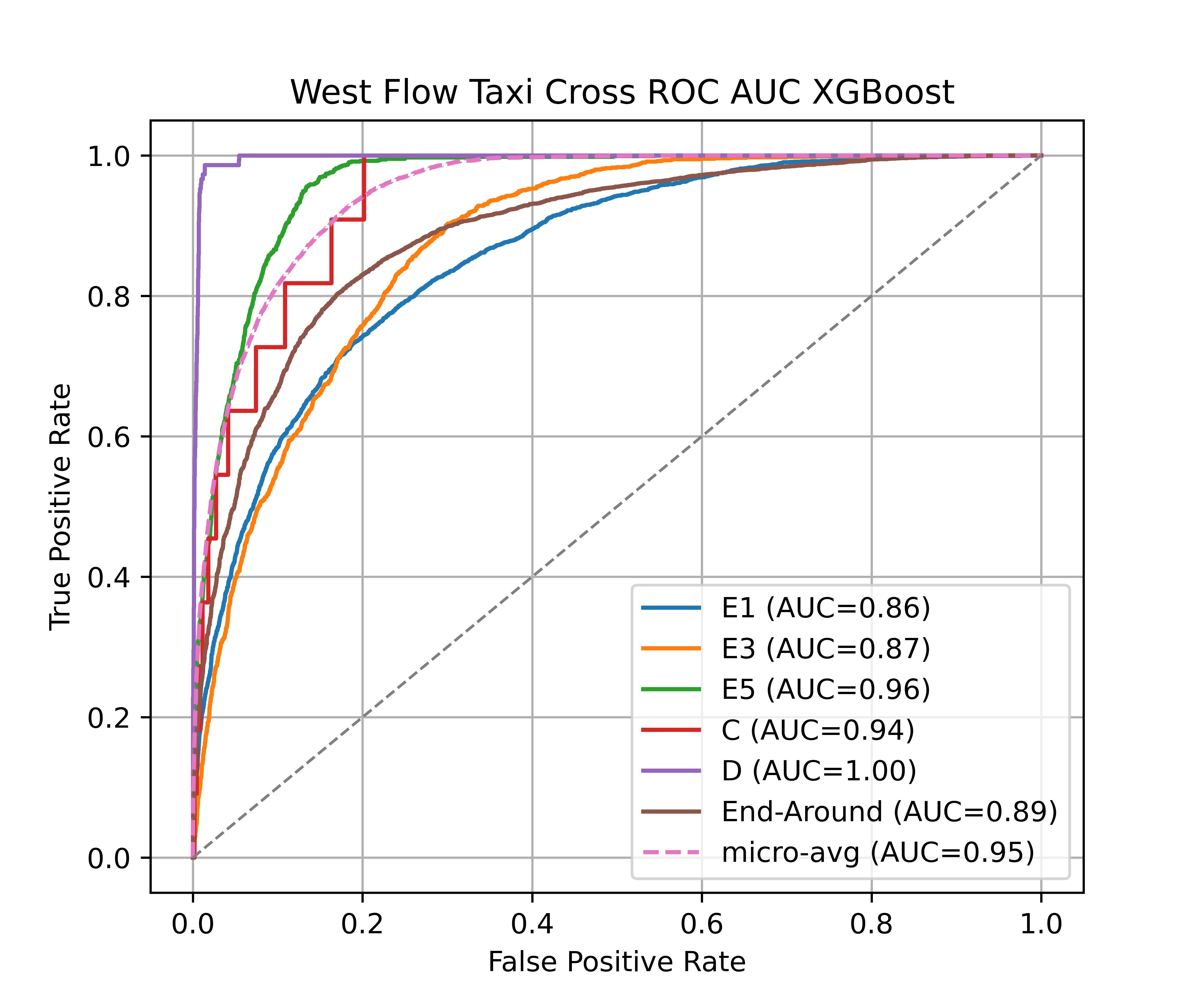}
        \caption{West XGBoost}
        \label{fig:WESTXGSTEP2}
    \end{subfigure}
    
    \begin{subfigure}[b]{0.45\textwidth}
        \centering
        \includegraphics[scale=0.45]{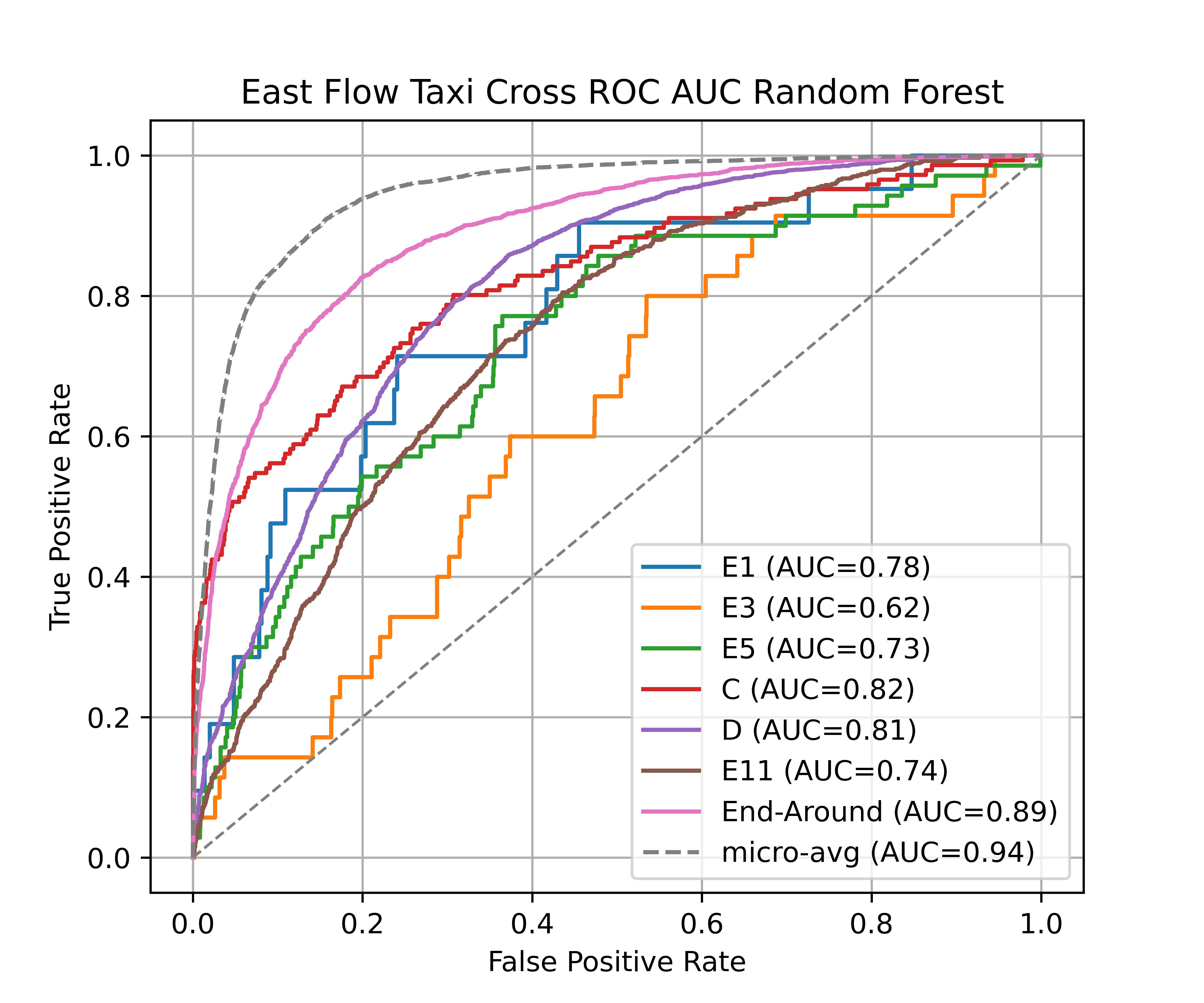}
        \caption{East Random Forest}
        \label{fig:EASTRFSTEP2}
    \end{subfigure}
    \begin{subfigure}[b]{0.45\textwidth}
        \centering
        \includegraphics[scale=0.45]{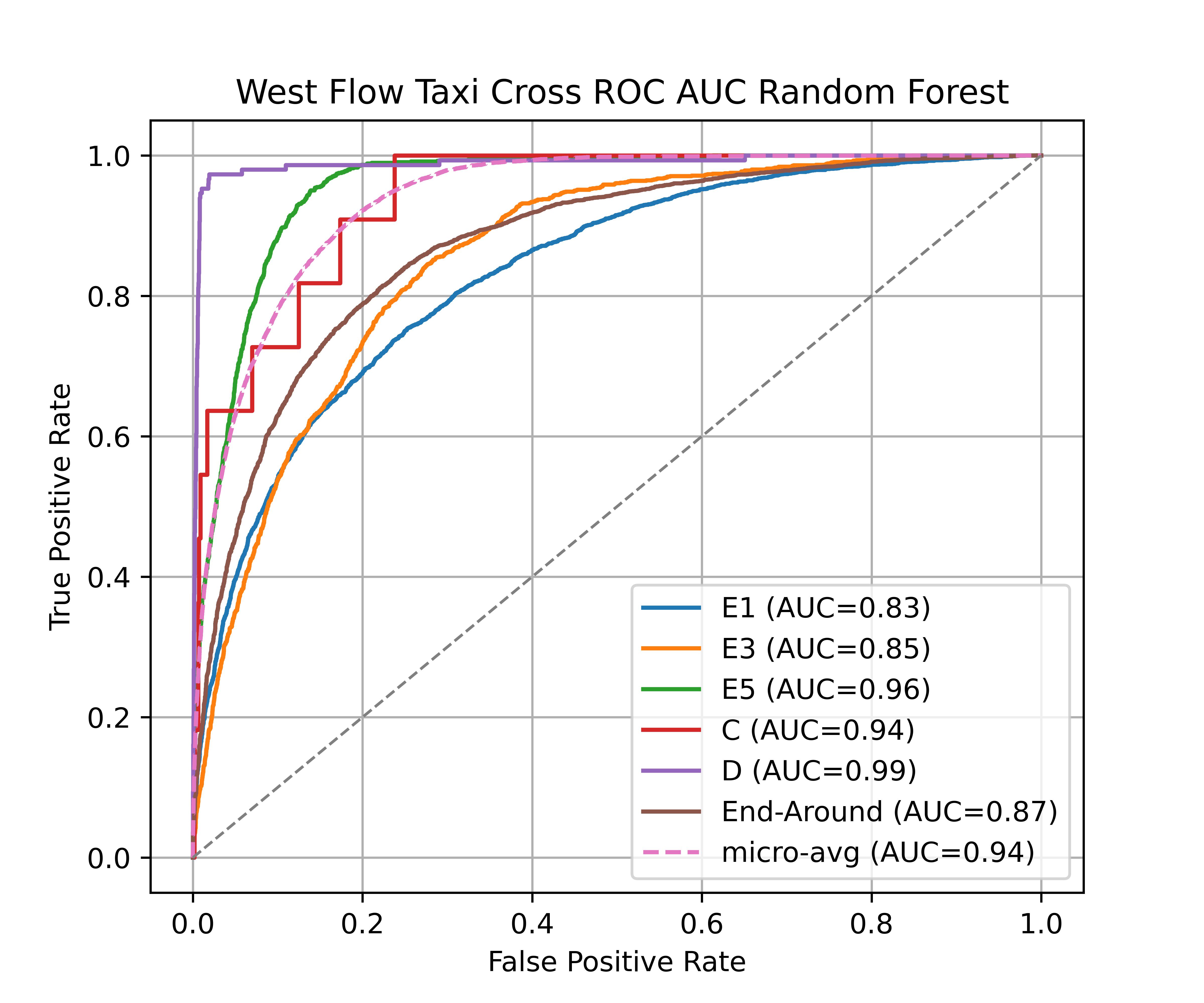}
        \caption{West Random Forest}
        \label{fig:westrfstep2}
    \end{subfigure}
    
    \begin{subfigure}[b]{0.45\textwidth}
        \centering
        \includegraphics[scale=0.45]{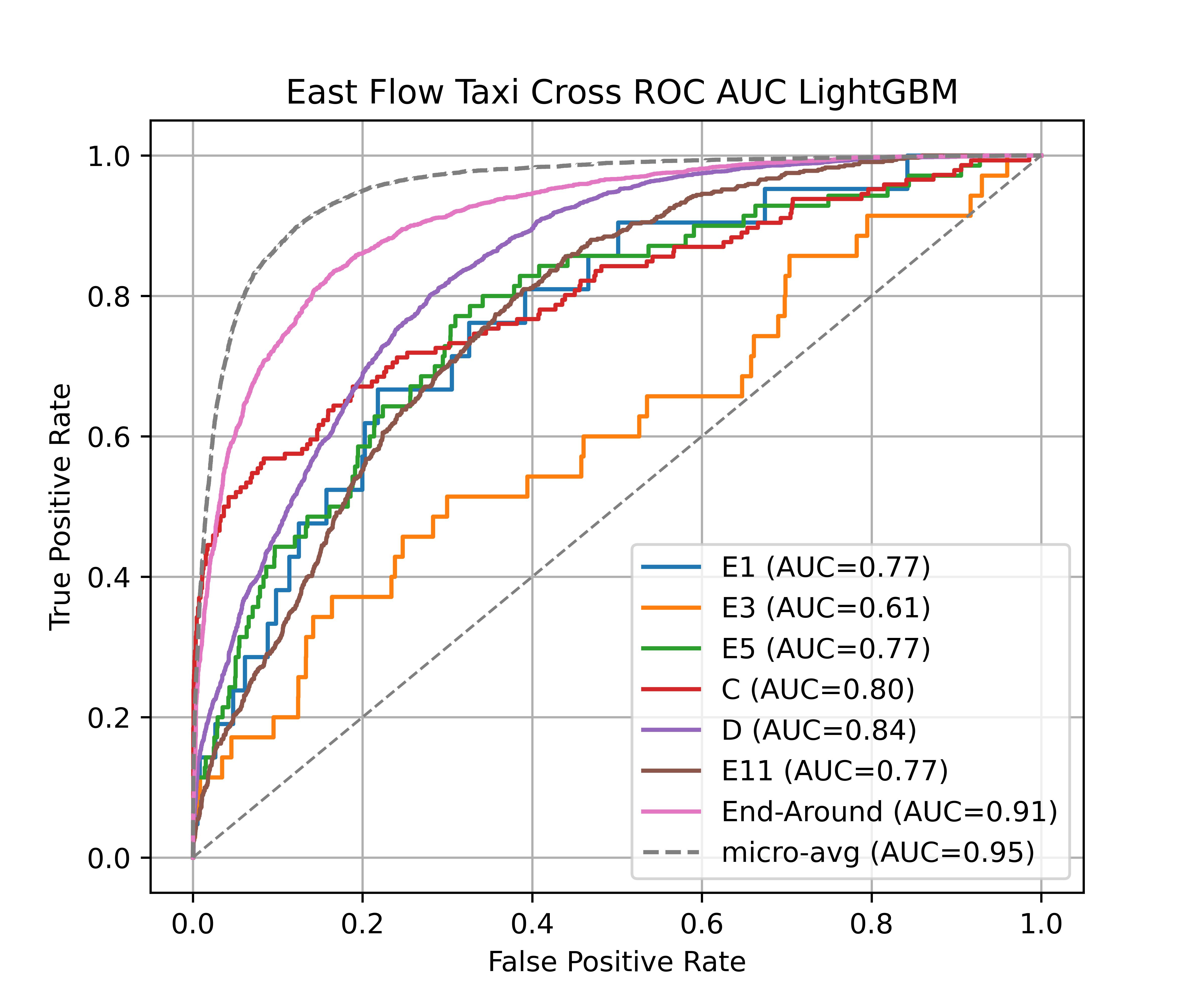}
        \caption{East LightGBM}
        \label{fig:eastgbmstep2}
    \end{subfigure}
    \begin{subfigure}[b]{0.45\textwidth}
        \centering
        \includegraphics[scale=0.45]{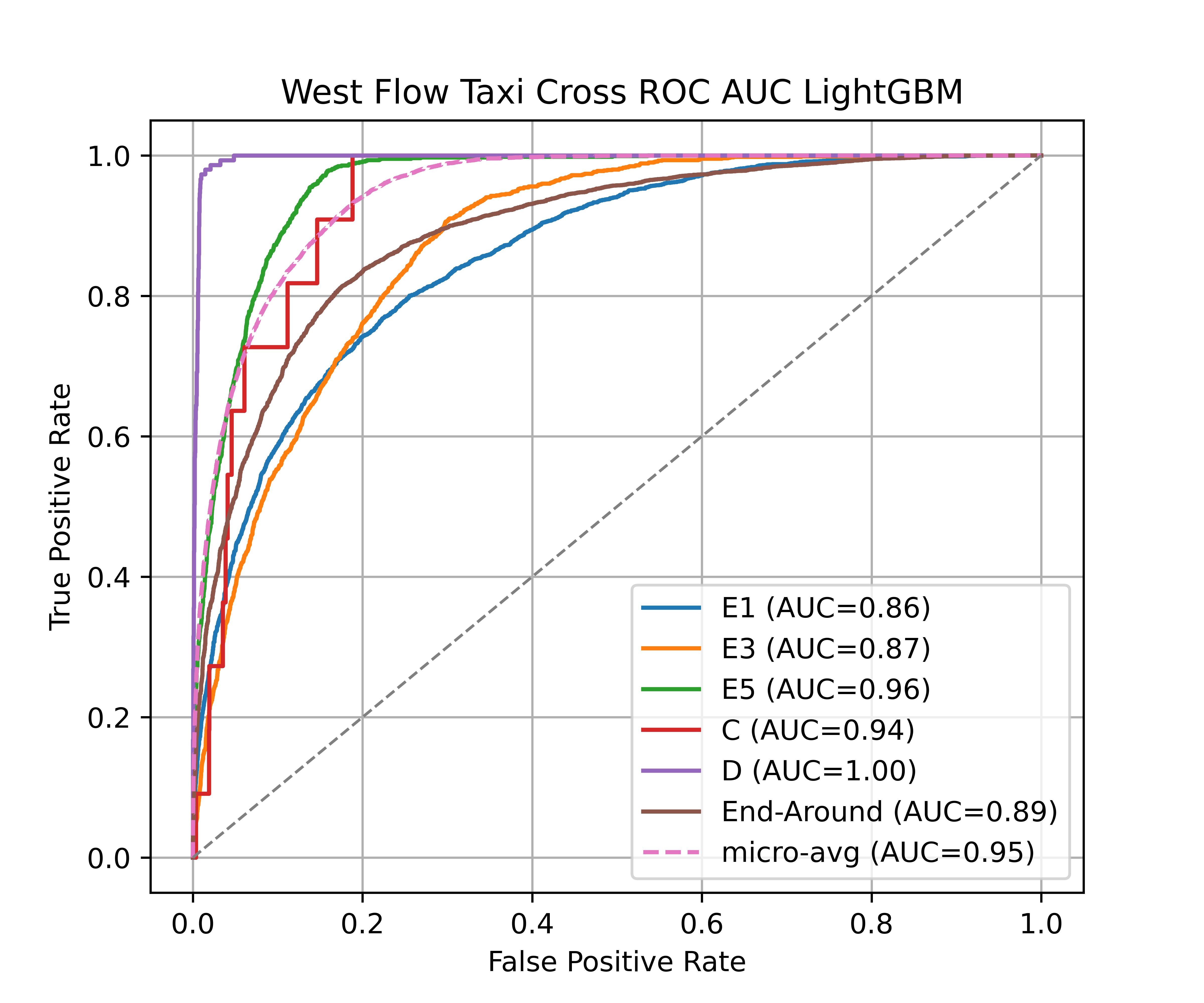}
        \caption{West LightGBM}
        \label{fig:westgbmstep2}
    \end{subfigure}

    \caption{ROC AUC Plots Stage II}
    \label{fig:ROCAUCSTAGE2}
\end{figure}

\begin{figure}[H]
    \centering
    \begin{subfigure}[b]{0.45\textwidth}
        \centering
        \includegraphics[scale=0.45]{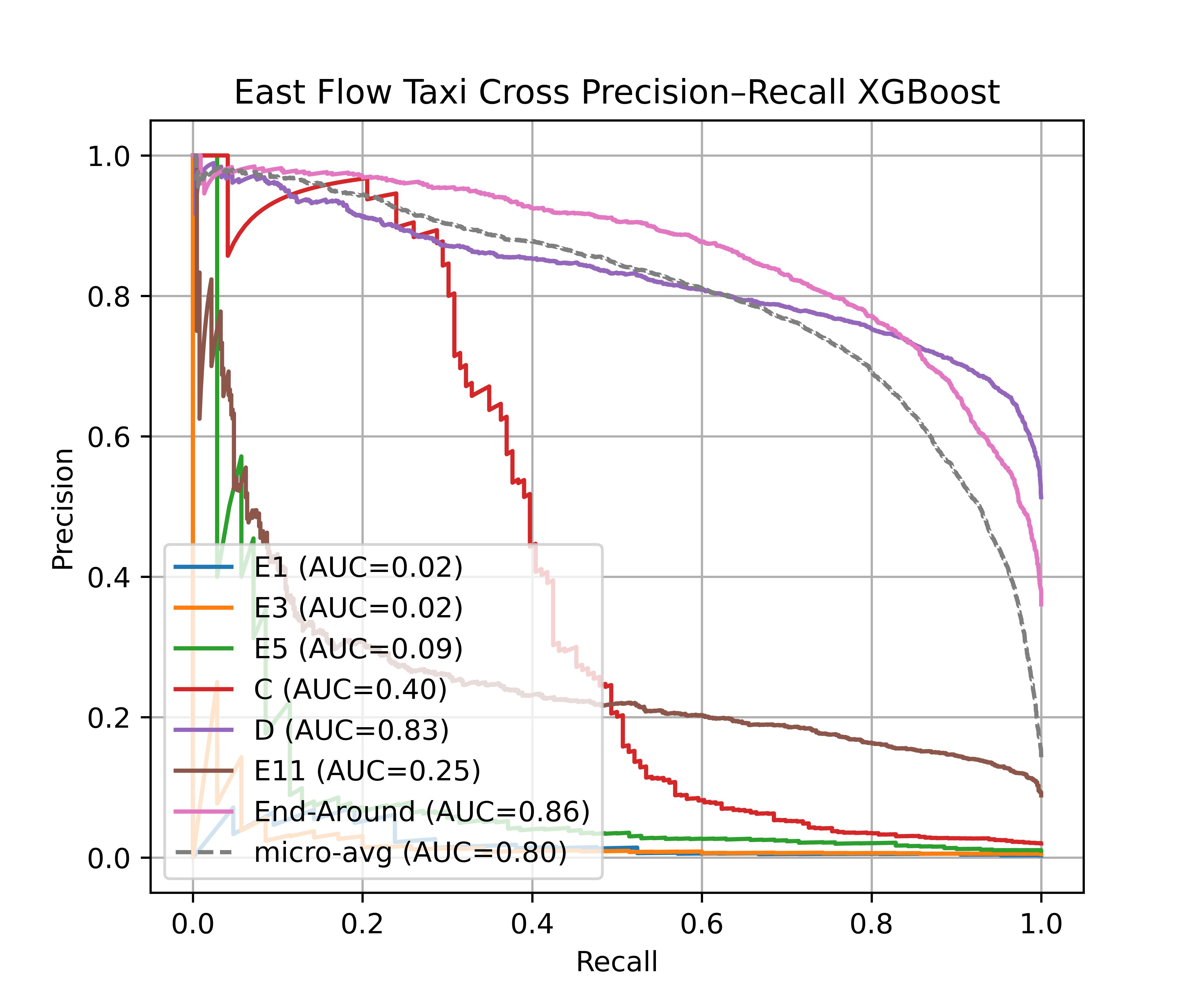}
        \caption{East XGBoost}
        \label{fig:EASTXGSTEP2PR}
    \end{subfigure}
    \begin{subfigure}[b]{0.45\textwidth}
        \centering
        \includegraphics[scale=0.45]{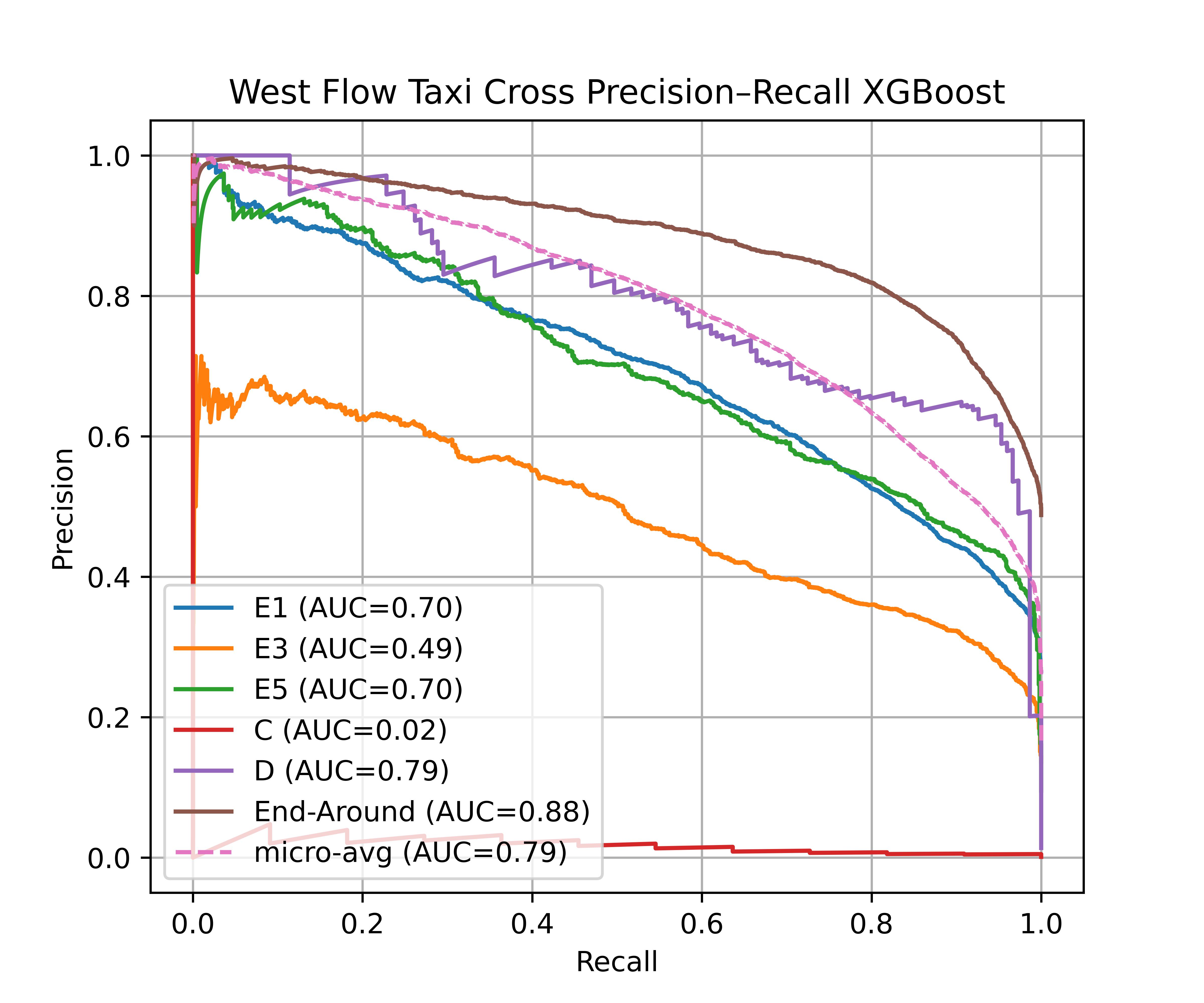}
        \caption{West XGBoost}
        \label{fig:WESTXGSTEP2PR}
    \end{subfigure}
    
    \begin{subfigure}[b]{0.45\textwidth}
        \centering
        \includegraphics[scale=0.45]{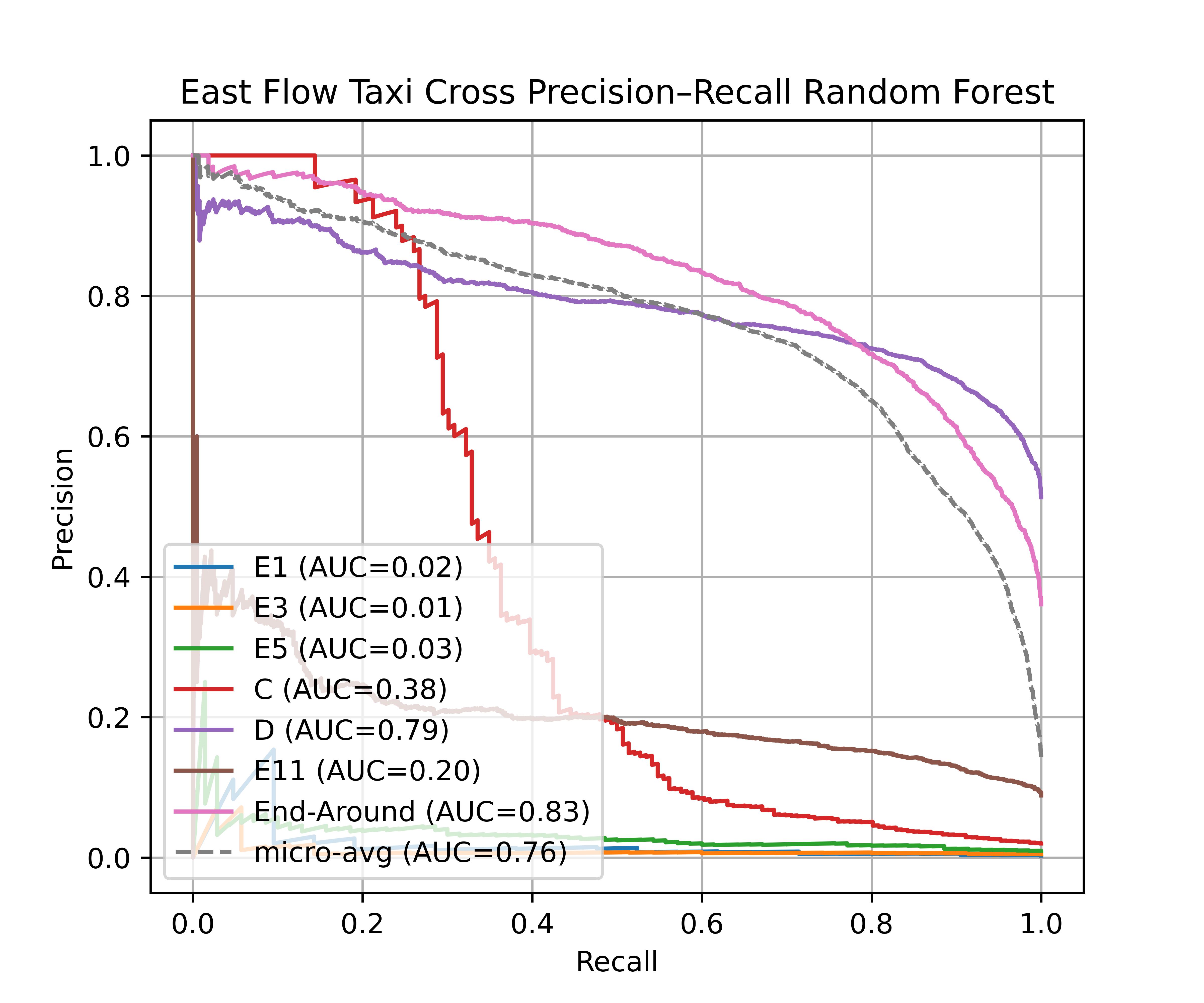}
        \caption{East Random Forest}
        \label{fig:EASTRFSTEP2PR}
    \end{subfigure}
    \begin{subfigure}[b]{0.45\textwidth}
        \centering
        \includegraphics[scale=0.45]{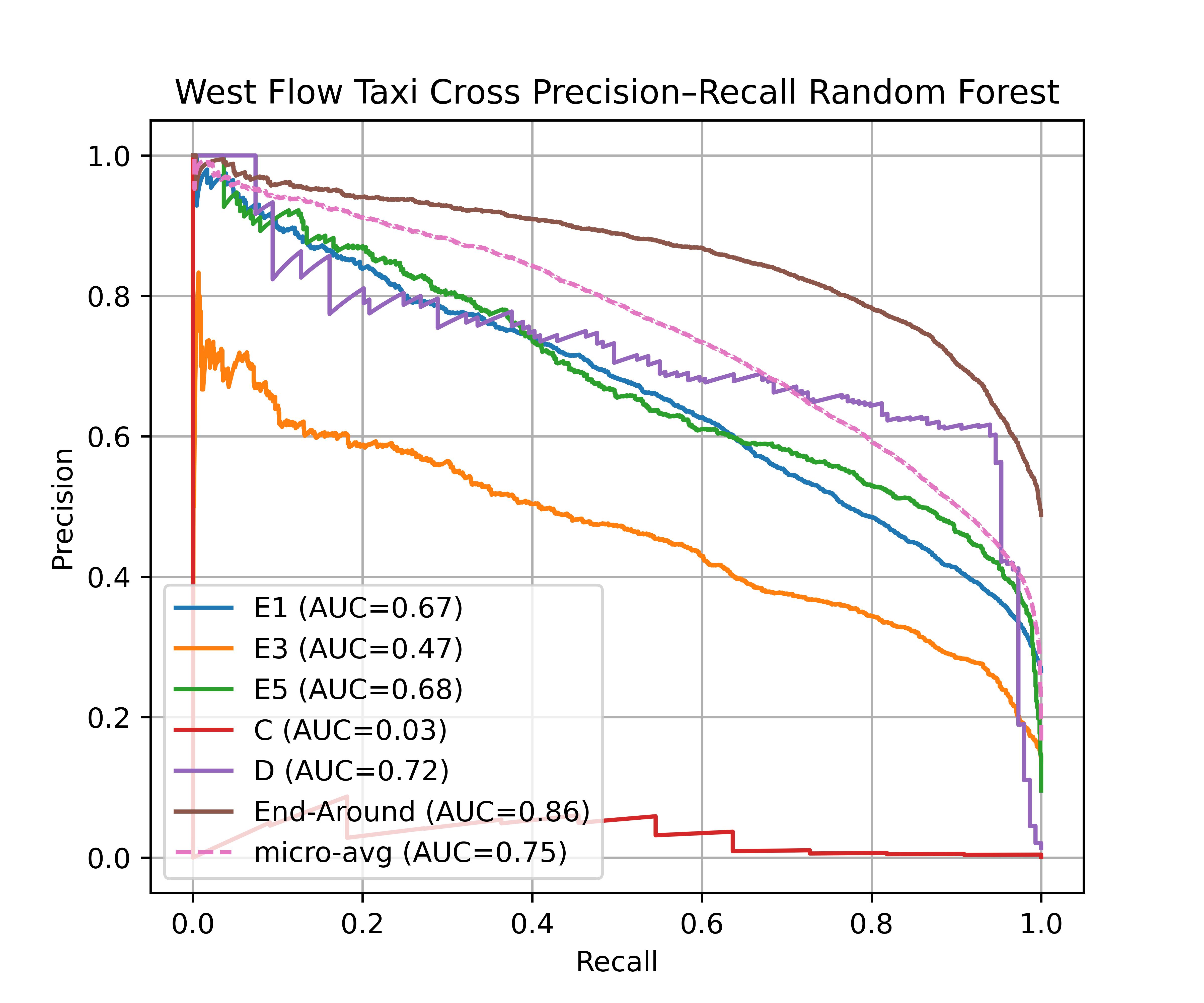}
        \caption{West Random Forest}
        \label{fig:westrfstep2PR}
    \end{subfigure}
    
    \begin{subfigure}[b]{0.45\textwidth}
        \centering
        \includegraphics[scale=0.45]{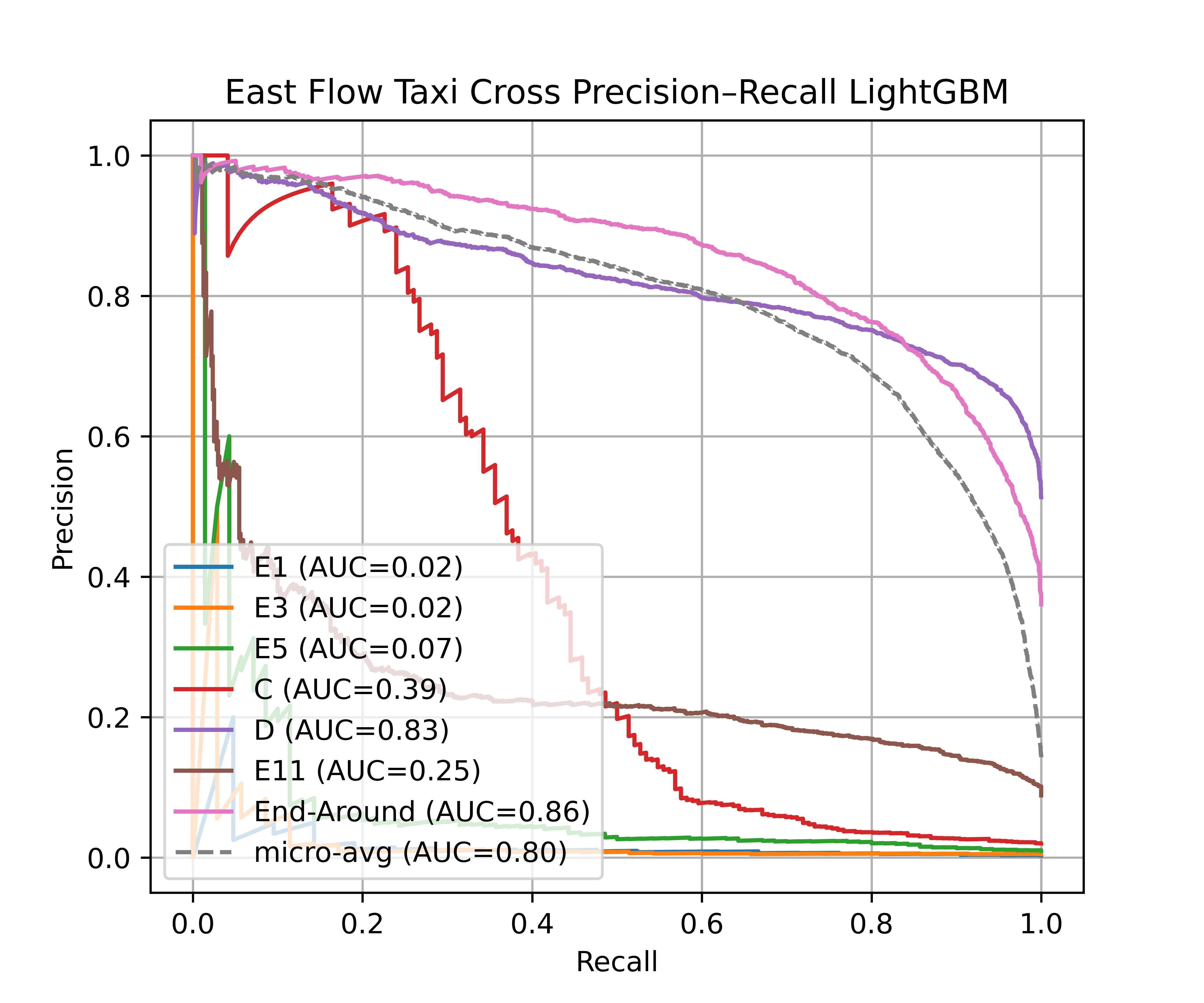}
        \caption{East LightGBM}
        \label{fig:eastgbmstep2PR}
    \end{subfigure}
    \begin{subfigure}[b]{0.45\textwidth}
        \centering
        \includegraphics[scale=0.45]{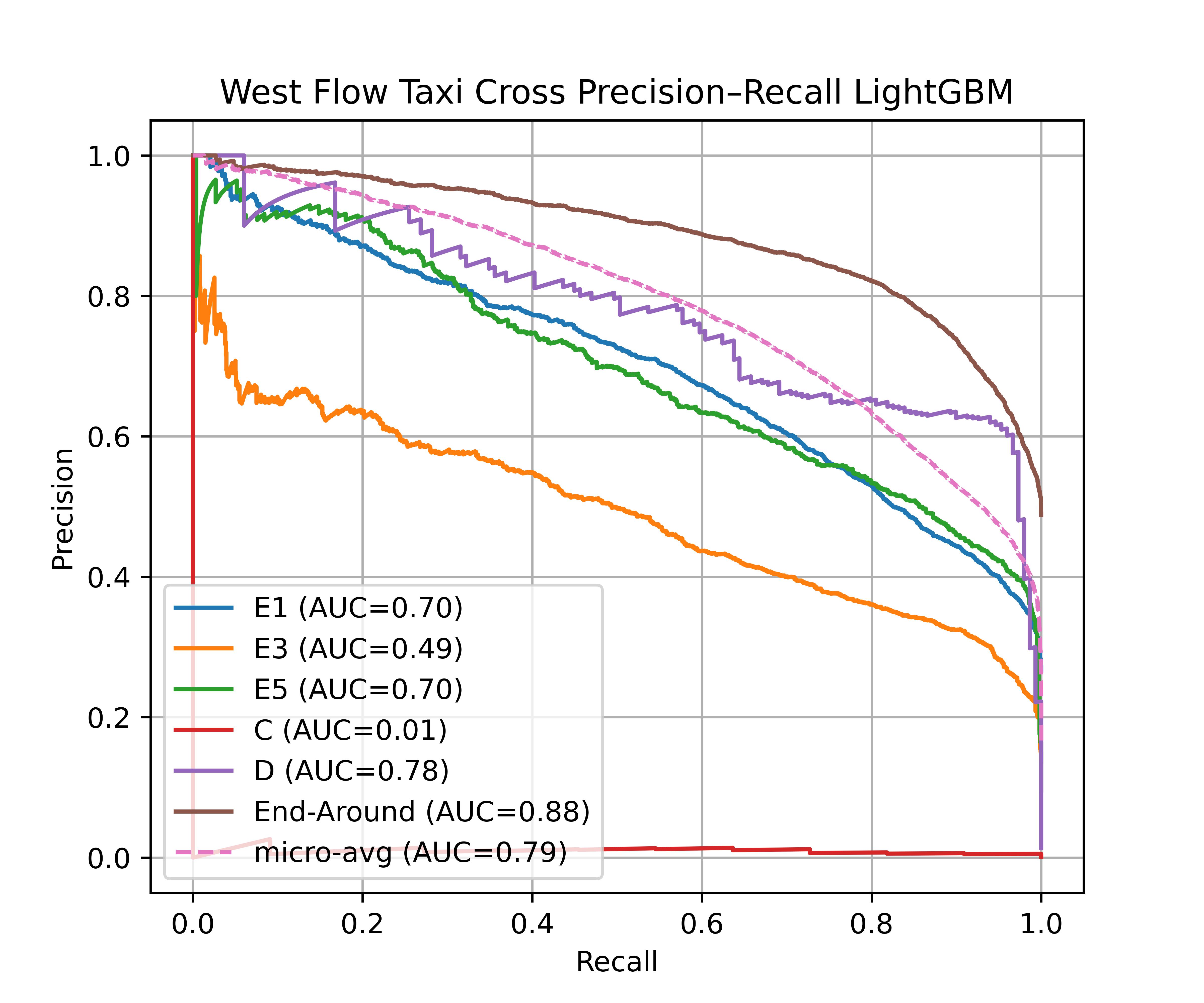}
        \caption{West LightGBM}
        \label{fig:westgbmstep2PR}
    \end{subfigure}

    \caption{Precision Recall Plots Stage II}
    \label{fig:PRSTAGE2}
\end{figure}

\begin{figure}[H]
    \centering
    \begin{subfigure}[b]{0.45\textwidth}
        \centering
        \includegraphics[scale=0.15]{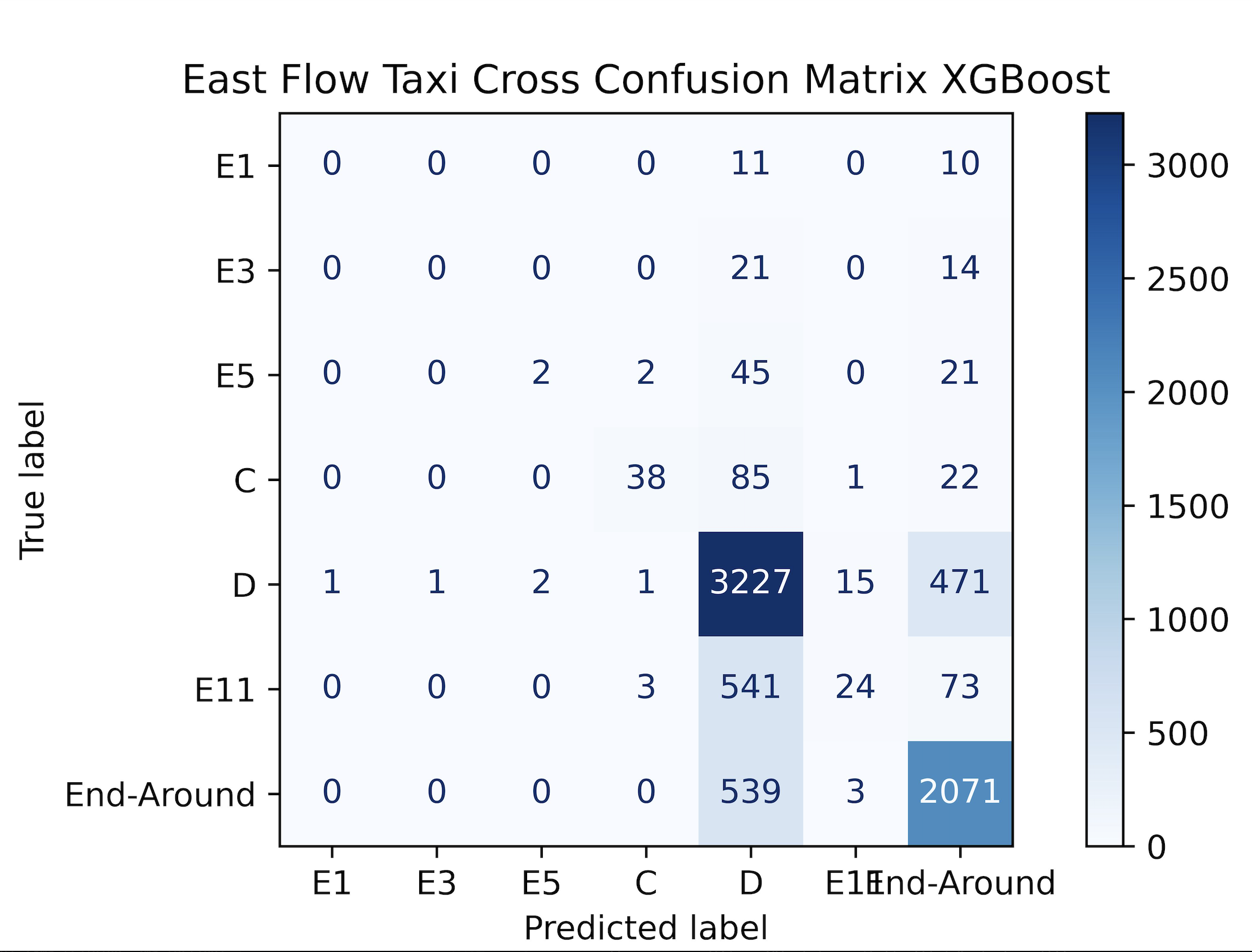}
        \caption{East XGBoost}
        \label{fig:EASTXGSTEP2CM}
    \end{subfigure}
    \begin{subfigure}[b]{0.45\textwidth}
        \centering
        \includegraphics[scale=0.15]{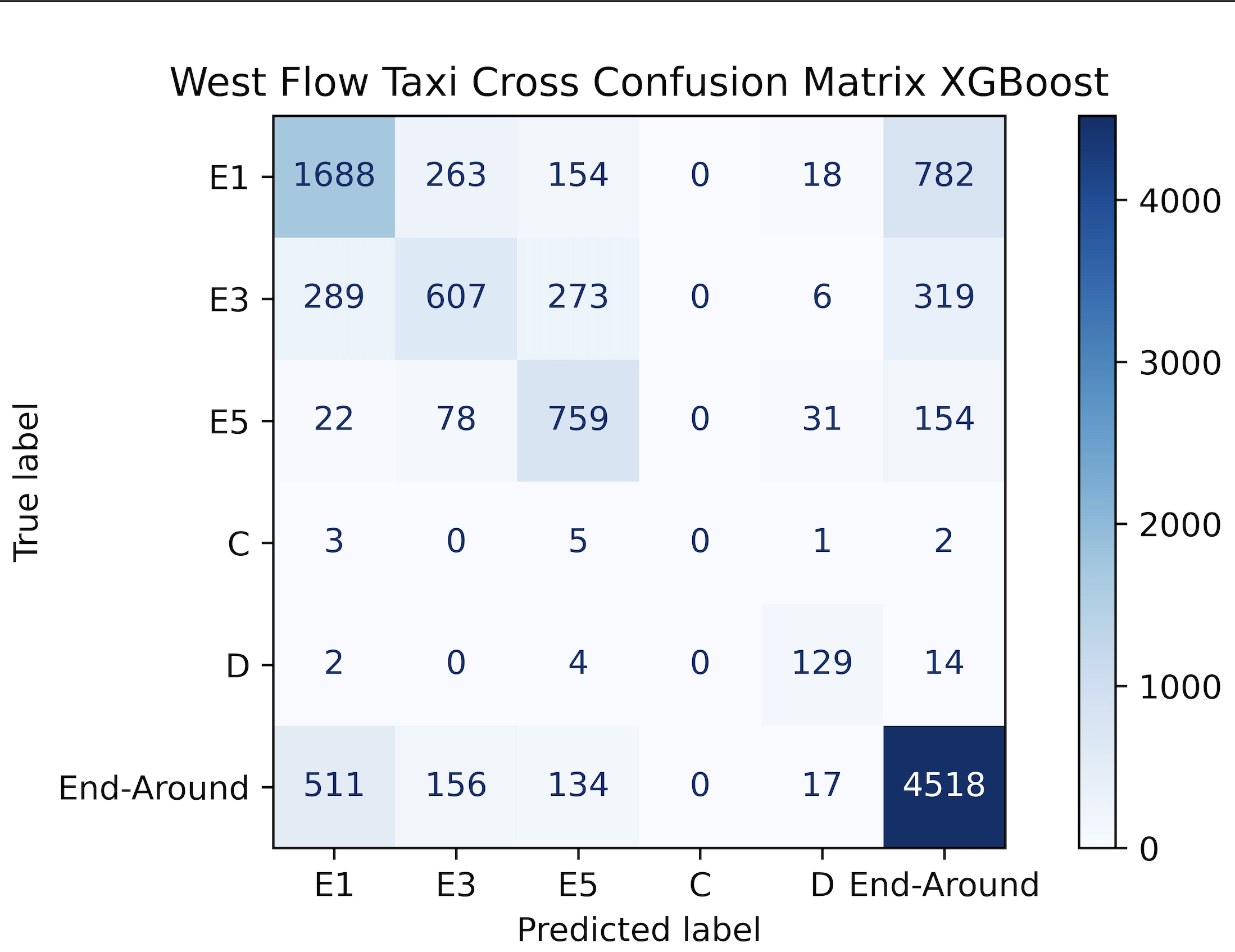}
        \caption{West XGBoost}
        \label{fig:WESTXGSTEP2CM}
    \end{subfigure}
    
    \begin{subfigure}[b]{0.45\textwidth}
        \centering
        \includegraphics[scale=0.15]{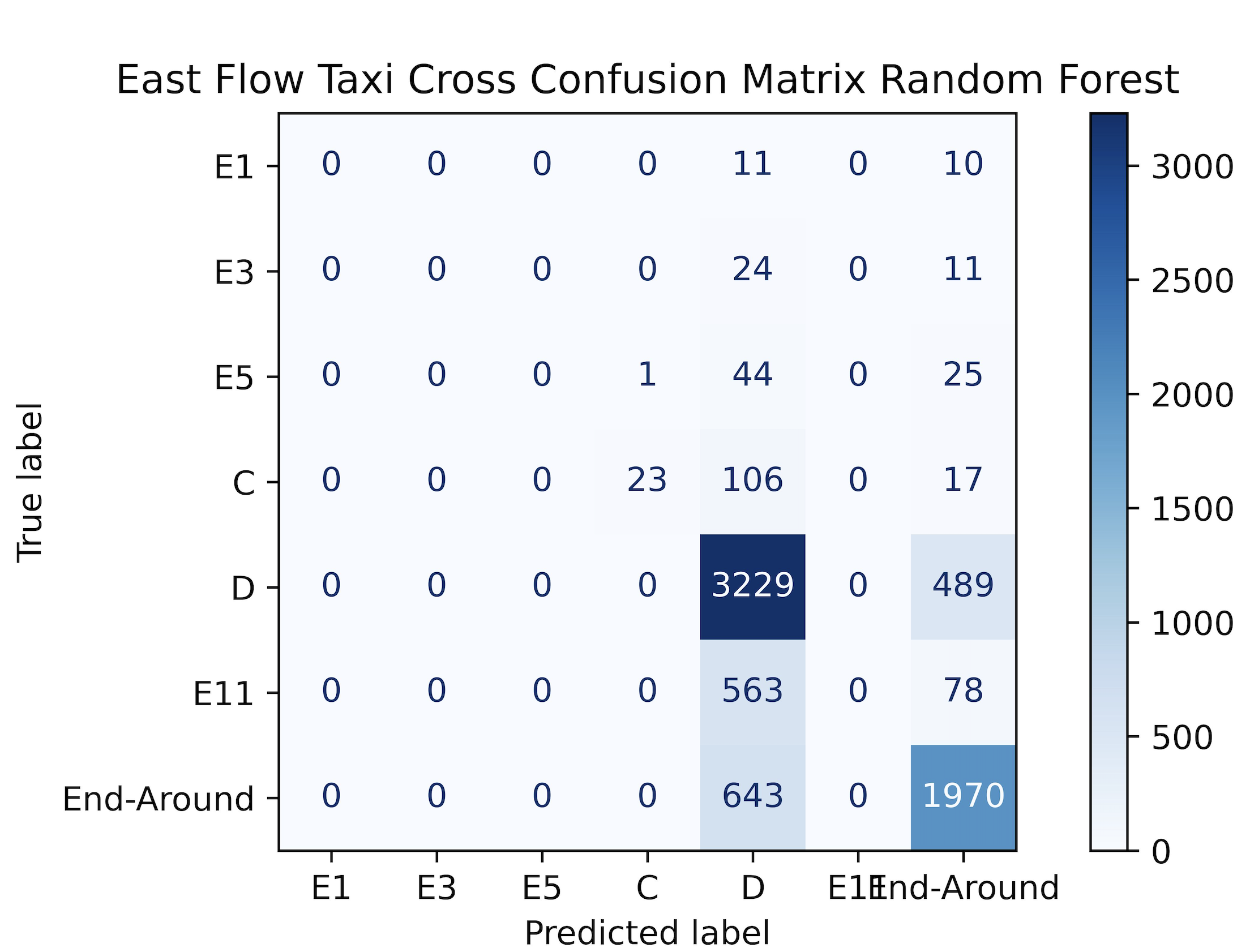}
        \caption{East Random Forest}
        \label{fig:EASTRFSTEP2CM}
    \end{subfigure}
    \begin{subfigure}[b]{0.45\textwidth}
        \centering
        \includegraphics[scale=0.15]{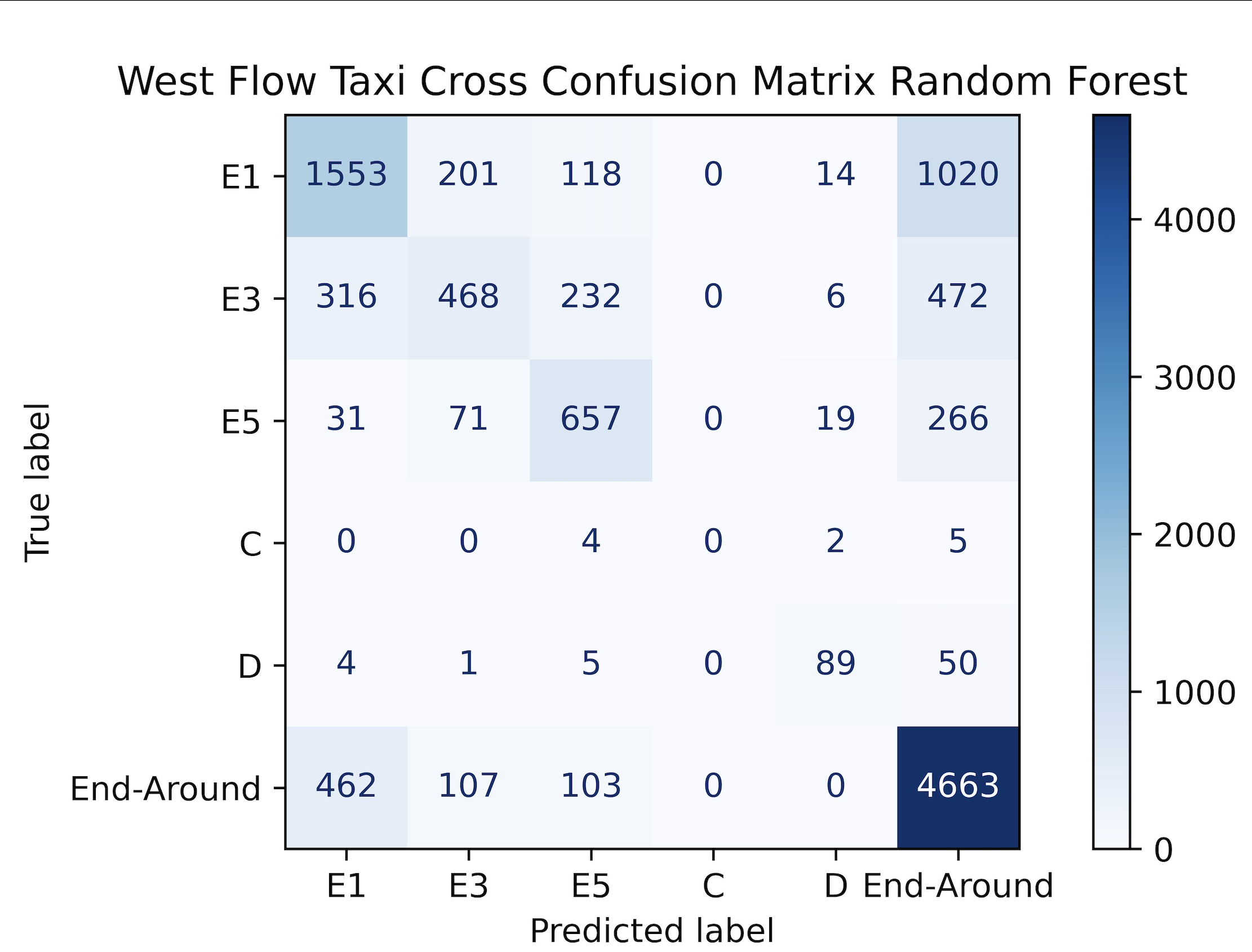}
        \caption{West Random Forest}
        \label{fig:westrfstep2CM}
    \end{subfigure}
    
    \begin{subfigure}[b]{0.45\textwidth}
        \centering
        \includegraphics[scale=0.15]{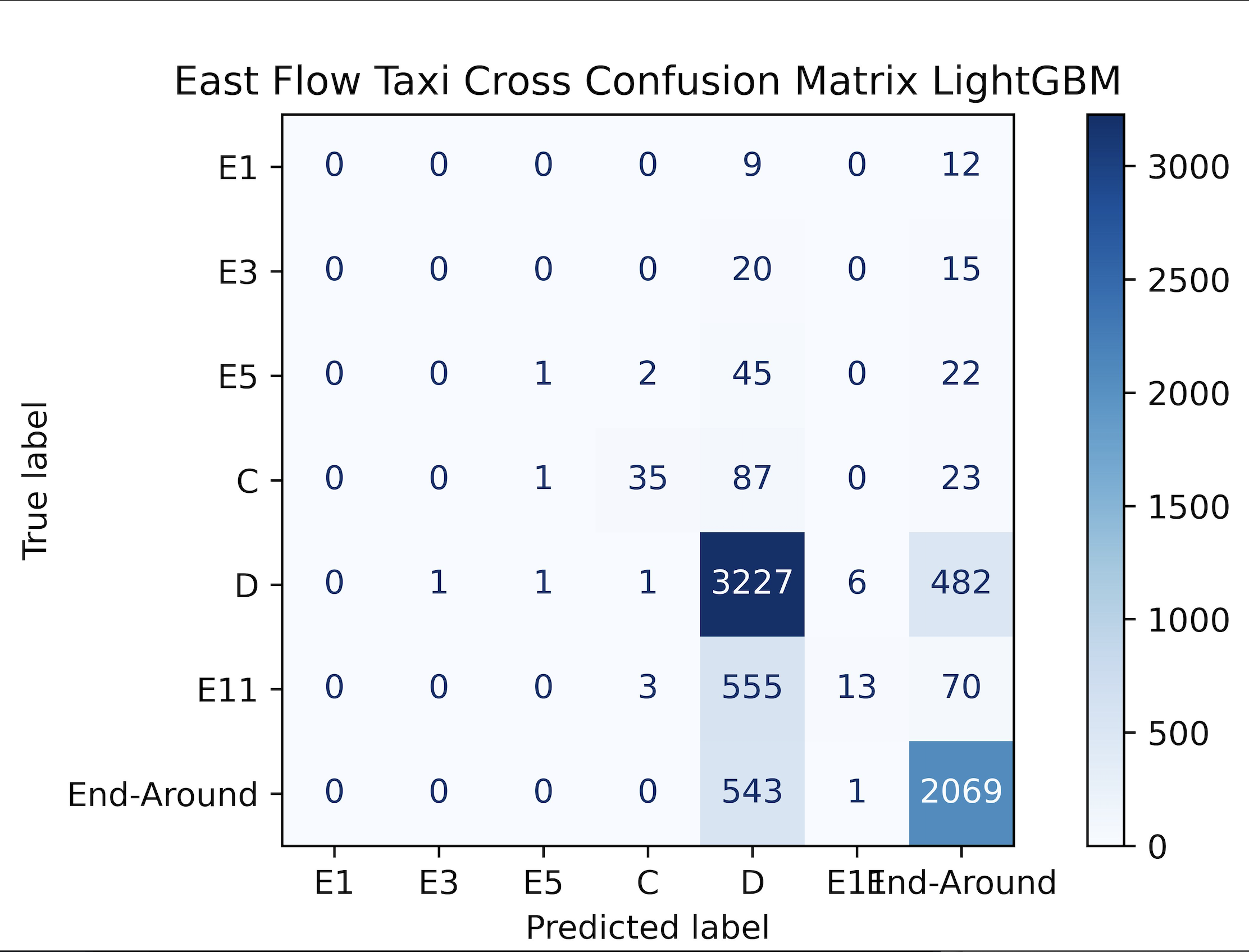}
        \caption{East LightGBM}
        \label{fig:eastgbmstep2CM}
    \end{subfigure}
    \begin{subfigure}[b]{0.45\textwidth}
        \centering
        \includegraphics[scale=0.15]{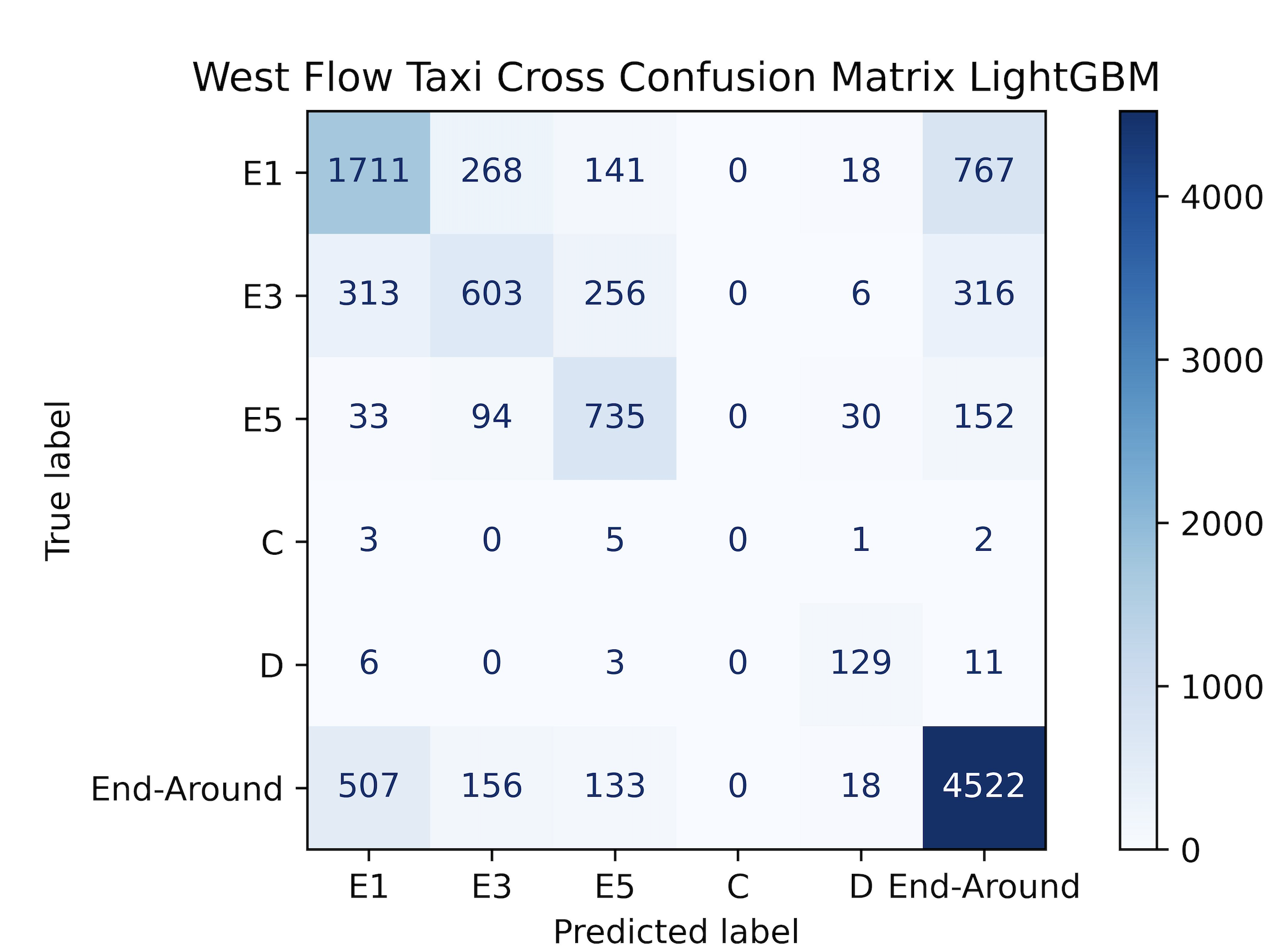}
        \caption{West LightGBM}
        \label{fig:westgbmstep2CM}
    \end{subfigure}

    \caption{Confusion Matrices Stage II}
    \label{fig:CMSTAGE2}
\end{figure}

\Cref{tab:STAGE2_MODEL COMPARISON} shows lower overall accuracies (0.70--0.74) than Stage~I, as expected for a larger label set with heavier skew. XGBoost and LightGBM tie on accuracy in both flows; XGBoost holds a slight macro-F1 edge. ROC-AUCs are respectable but drop relative to Stage~I (\Cref{fig:ROCAUCSTAGE2}). As with exits, macro PR-AUCs (\Cref{fig:PRSTAGE2}) emphasize that high performance is concentrated in the majority classes (e.g., west end-around, east D), while rarer crossing points suffer in precision and recall. Confusion matrices (\Cref{fig:CMSTAGE2}) indicate two recurrent error modes: mode collapse toward the majority option (often end-around) under high departure rate, and near-miss confusions among adjacent crossing points.

\begin{table}[H]
    \centering
    \caption{Class split of Stage~II for West Flow and East Flow arrivals.}
    \label{tab:stage2datasplit}
    \small
    \begin{tabular}{lc|lc}
        \toprule
        \multicolumn{2}{c|}{West Flow} & \multicolumn{2}{c}{East Flow} \\
        \cmidrule(r){1-2} \cmidrule(l){3-4}
        Class & \% & Class & \% \\
        \midrule
        \textbf{E1}         & 0.265 & \textbf{E1}         & 0.002 \\
        \textbf{E3}         & 0.136 & \textbf{E3}         & 0.004 \\
        \textbf{E5}         & 0.095 & \textbf{E5}         & 0.009 \\
        \textbf{C}          & 0.001 & \textbf{C}          & 0.020 \\
        \textbf{D}          & 0.013 & \textbf{D}          & 0.513 \\
        \textbf{End-Around} & 0.487 & \textbf{E11}        & 0.088 \\
                            &       & \textbf{End-Around} & 0.360 \\
        \bottomrule
    \end{tabular}
\end{table}

\begin{table}[H]
    \centering
    \caption{Stage II: Taxi Cross Model Comparison}
    \label{tab:STAGE2_MODEL COMPARISON}
    \begin{tabular}{llcccc}
        \toprule
        & & Precision & Recall & F1-Score & Accuracy \\
        \midrule
        \multirow{3}{*}{West Flow}
            & \textbf{XGBoost}  & 0.535 & 0.571 & 0.546 & 0.704  \\
            & \textbf{Random Forest}  & 0.533 & 0.491 & 0.504 & 0.697 \\
            & \textbf{LightGBM}  & 0.534 & 0.568 & 0.545 & 0.704  \\
        \addlinespace
        \multirow{3}{*}{East Flow}
            & \textbf{XGBoost}  & 0.488 & 0.283 & 0.299 & 0.740  \\
            & \textbf{Random Forest} & 0.345 & 0.254 & 0.257 & 0.720 \\
            & \textbf{LightGBM} & 0.479 & 0.276 & 0.287 & 0.738  \\
        \bottomrule
    \end{tabular}
\end{table}

\subsection{Insights and Feature Attributions}
\begin{figure}[H]
    \centering
    \begin{subfigure}[b]{0.45\textwidth}
        \centering
        \includegraphics[scale=0.30]{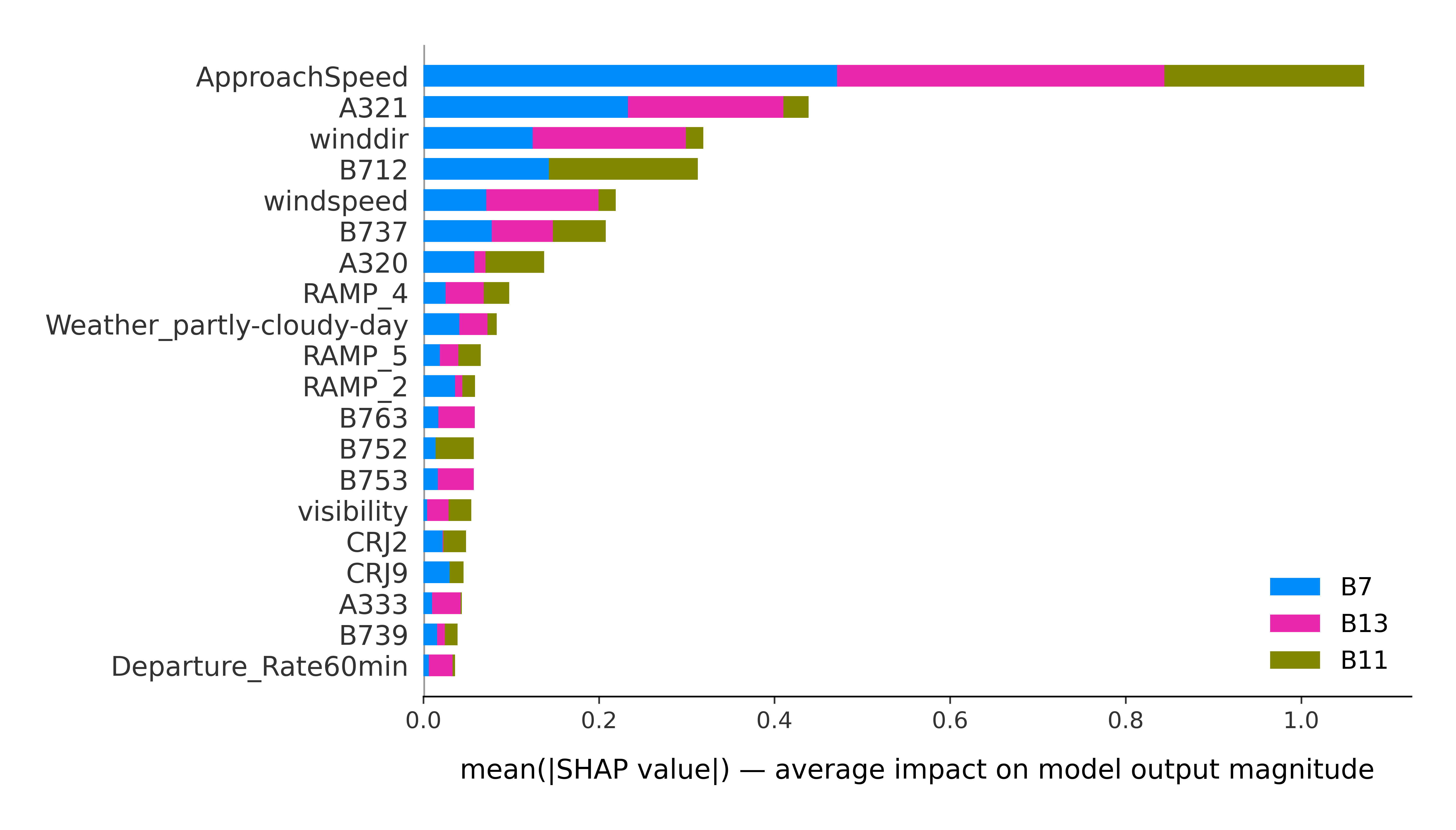}
        \caption{East XGBoost Taxi Exit}
        \label{fig:eaststep1SHAP}
    \end{subfigure}
    \begin{subfigure}[b]{0.45\textwidth}
        \centering
        \includegraphics[scale=0.30]{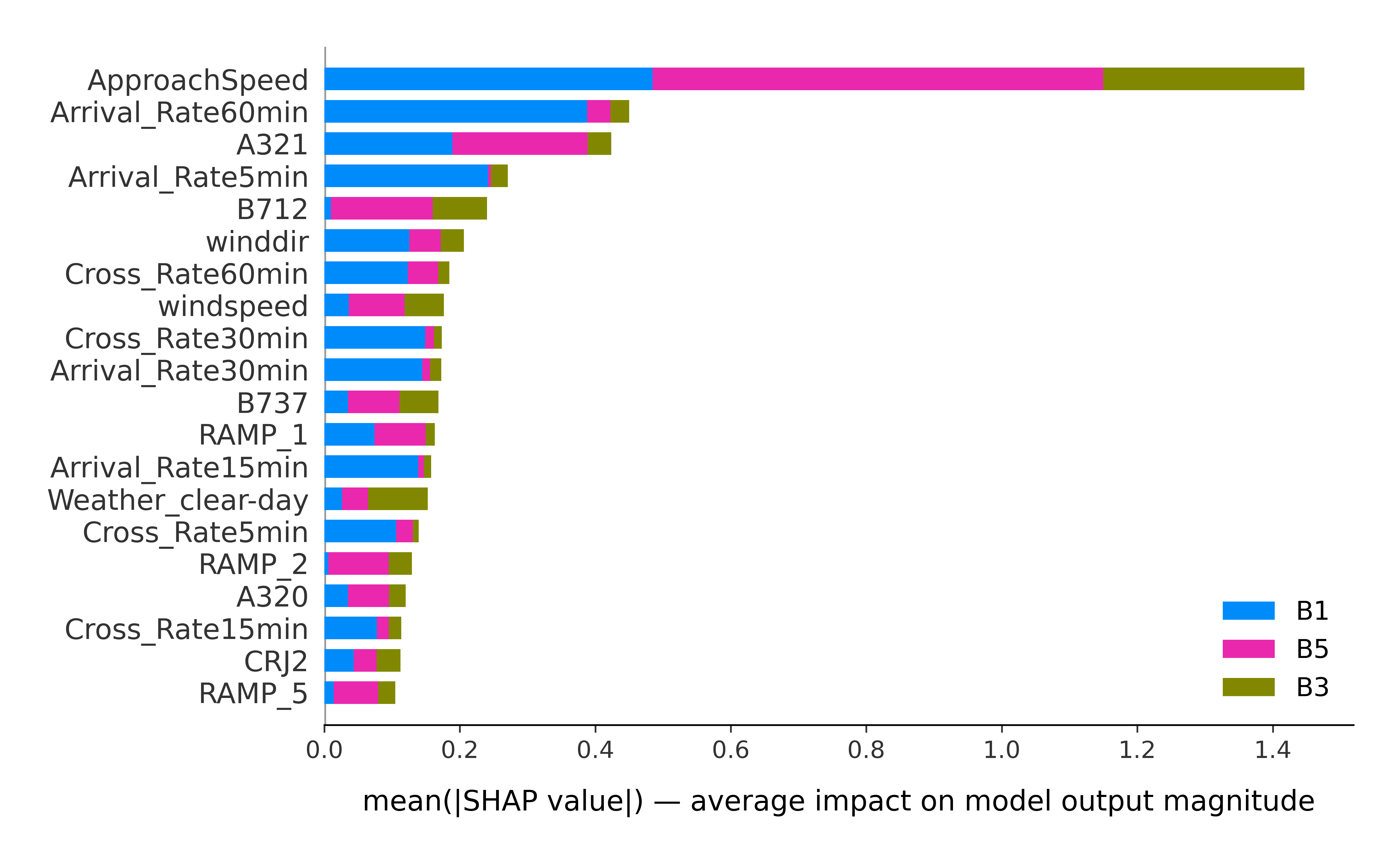}
        \caption{West XGBoost Taxi Exit}
        \label{fig:weststep1SHAP}
    \end{subfigure}
    
    \begin{subfigure}[b]{0.45\textwidth}
        \centering
        \includegraphics[scale=0.30]{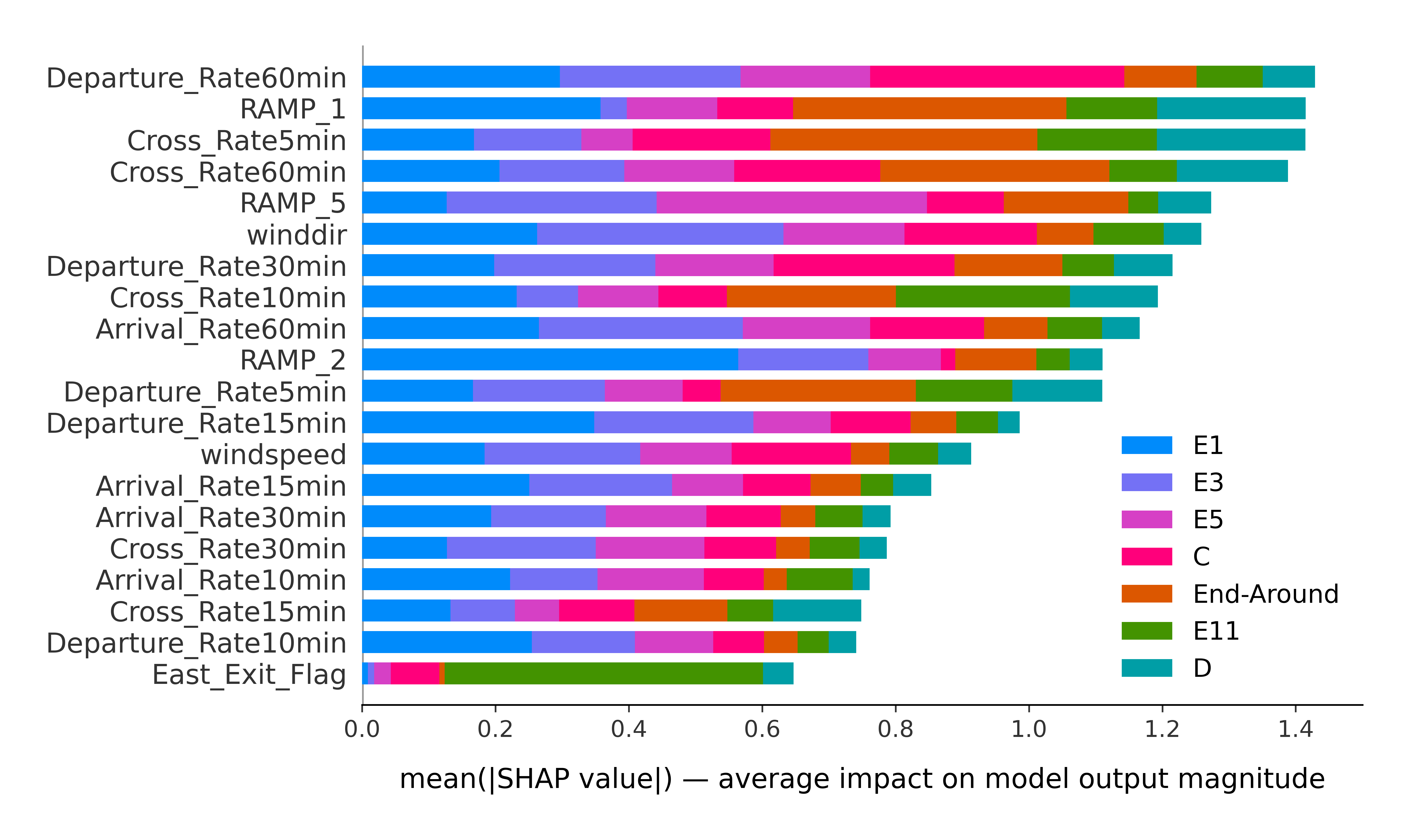}
        \caption{East XGBoost Taxi Cross}
        \label{fig:eaststep2SHAP}
    \end{subfigure}
    \begin{subfigure}[b]{0.45\textwidth}
        \centering
        \includegraphics[scale=0.30]{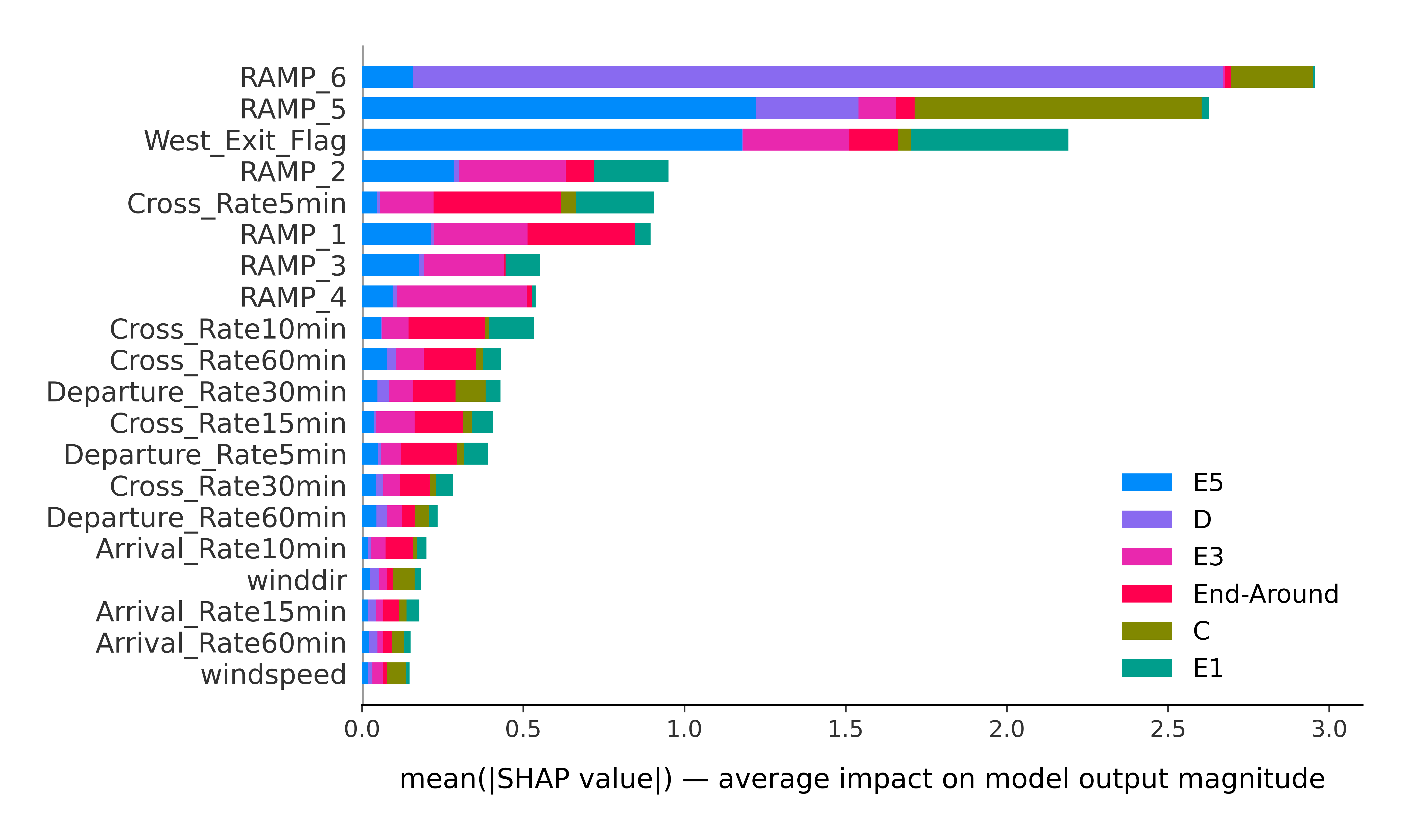}
        \caption{West XGBoost Taxi Cross}
        \label{fig:weststep2SHAP}
    \end{subfigure}

    \caption{SHAP Feature Importance}
    \label{fig:SHAP}
\end{figure}

Focusing on the XGBoost models, \Cref{fig:SHAP} summarizes the leading contributors for both prediction stages. For exit prediction, approach speed is the dominant predictor in both east and west flows. In the east flow, aircraft type, wind conditions, and ramp destination also contribute substantially. In the west flow, arrival rate, aircraft type, wind conditions, and end-around crossing rate are the next most informative features. For crossing decisions in the east flow, the strongest predictors are departure rate, ramp destination, crossing rate, and wind direction; in the west flow, ramp destination, the taken runway exit, and crossing rate contribute most.

SHAP results (\Cref{fig:SHAP}, bottom row) indicate that in the east flow, departure rate, crossing rate, and ramp destination dominate: higher departure throughput suppresses crossing probability and shifts mass to the end-around unless recent crossing activity (i.e., gap availability) is high. In the west flow, the realized runway exit and specific ramp (with longer distances via end-around) are most influential, reflecting the geometric interaction between exit-ramp pairing and route alternatives. These patterns are intuitive: when the end-around imposes a substantial path-length penalty (east for some ramps), crossing is attractive only if departure streams yield viable gaps; where geometry favors early exit to short crossing paths (some west ramp pairs), exit choice and ramp dominate.

The divergence between strong ROC-AUC and modest macro-F1 is a consequence of skewness. ROC is prevalence-insensitive, whereas PR highlights the positive predictive value of rare classes. For operational use, PR-AUC, per-class PR curves, and normalized confusion matrices should be primary diagnostic tools. In addition, reporting hierarchical metrics (e.g., Stage~II binary \{cross vs.\ end-around\} accuracy alongside full multiclass accuracy) distinguishes safety-critical errors (predicting cross when end-around is appropriate under high departure flow) from less costly within-family mislabels (e.g., E3 vs.\ E5).

Several practical implications follow. Approach speed is the predominant factor for anticipating exit choice; coupling speed advisories with exit availability could improve predictability at Stage~I. Stage~II decisions hinge on the interaction of departure and crossing rates with airport geometry: in high-$\lambda^\text{dep}$ regimes, the end-around is the safer, more predictable default, but when recent crossing activity is high and geometry is favorable, crossing becomes competitive. Even when minority-class recall is limited, calibrated probabilities enable cost-aware use. Thresholds can be tuned to suppress false crossing predictions under heavy departure demand while still surfacing crossing opportunities when confidence is high.

\subsection{Broader Method Comparison}
To justify the choice of gradient-boosted tree ensembles, we benchmark nine classification methods spanning linear models, instance-based learning, neural networks, and tree-based ensembles. \Cref{tab:method_s1_combined} and \Cref{tab:method_s2_combined} present the results across all four prediction tasks. Features are standardized for methods that require it (Logistic Regression, SVM, KNN, MLP); tree-based methods operate on raw features.

\begin{table}[H]
\centering
\caption{Stage~1 runway exit classification performance comparison for East Flow (B7/B11/B13) and West Flow (B1/B3/B5). Best values in each column are \textbf{bolded}.}
\label{tab:method_s1_combined}
\small
\resizebox{\textwidth}{!}{%
\begin{tabular}{lccccc|ccccc}
\toprule
& \multicolumn{5}{c|}{East Flow (B7/B11/B13)} & \multicolumn{5}{c}{West Flow (B1/B3/B5)} \\
\cmidrule(r){2-6} \cmidrule(l){7-11}
Method & Accuracy & Macro F1 & MCC & Weighted F1 & ROC AUC & Accuracy & Macro F1 & MCC & Weighted F1 & ROC AUC \\
\midrule
Logistic Regression & 0.884 & 0.371 & 0.174 & 0.847 & 0.820 & 0.849 & 0.449 & 0.354 & 0.825 & 0.807 \\
SVM (Linear)        & 0.885 & 0.326 & 0.072 & 0.834 & ---   & 0.852 & 0.440 & 0.344 & 0.822 & ---   \\
KNN ($k$=5)         & 0.877 & 0.426 & 0.231 & 0.855 & 0.682 & 0.836 & 0.453 & 0.322 & 0.817 & 0.708 \\
MLP                 & 0.885 & 0.357 & 0.150 & 0.843 & 0.793 & 0.858 & 0.477 & 0.419 & 0.840 & 0.850 \\
Decision Tree       & 0.886 & 0.425 & 0.264 & 0.861 & 0.746 & 0.846 & 0.478 & 0.413 & 0.835 & 0.760 \\
\midrule
Random Forest       & 0.887 & 0.342 & 0.138 & 0.840 & 0.828 & 0.854 & 0.423 & 0.335 & 0.817 & 0.845 \\
XGBoost             & 0.894 & 0.422 & 0.304 & 0.866 & \textbf{0.840} & 0.862 & 0.487 & 0.425 & 0.842 & \textbf{0.872} \\
LightGBM            & 0.893 & \textbf{0.438} & 0.305 & 0.866 & 0.838 & \textbf{0.864} & \textbf{0.503} & \textbf{0.436} & \textbf{0.845} & 0.871 \\
CatBoost            & \textbf{0.895} & 0.425 & \textbf{0.315} & \textbf{0.867} & 0.839 & 0.862 & 0.477 & 0.421 & 0.841 & 0.869 \\
\bottomrule
\end{tabular}%
}
\end{table}

\begin{table}[H]
\centering
\caption{Stage~2 crossing/end-around classification performance comparison for East Flow (7 taxiway exits) and West Flow (6 taxiway exits). Best values in each column are \textbf{bolded}.}
\label{tab:method_s2_combined}
\small
\resizebox{\textwidth}{!}{%
\begin{tabular}{lccccc|ccccc}
\toprule
& \multicolumn{5}{c|}{East Flow (7 classes)} & \multicolumn{5}{c}{West Flow (6 classes)} \\
\cmidrule(r){2-6} \cmidrule(l){7-11}
Method & Accuracy & Macro F1 & MCC & Weighted F1 & ROC AUC & Accuracy & Macro F1 & MCC & Weighted F1 & ROC AUC \\
\midrule
Logistic Regression & 0.716 & 0.218 & 0.491 & 0.668 & 0.768 & 0.686 & 0.535 & 0.520 & 0.679 & 0.913 \\
SVM (Linear)        & 0.718 & 0.218 & 0.495 & 0.669 & ---   & 0.688 & 0.532 & 0.519 & 0.676 & ---   \\
KNN ($k$=5)         & 0.673 & 0.232 & 0.418 & 0.641 & 0.624 & 0.631 & 0.489 & 0.444 & 0.630 & 0.796 \\
MLP                 & 0.721 & 0.244 & 0.502 & 0.676 & 0.756 & 0.698 & 0.545 & 0.544 & 0.693 & 0.918 \\
Decision Tree       & 0.712 & 0.263 & 0.486 & 0.670 & 0.713 & 0.687 & 0.535 & 0.526 & 0.681 & 0.854 \\
\midrule
Random Forest       & 0.724 & 0.266 & 0.508 & 0.680 & 0.770 & 0.691 & 0.529 & 0.524 & 0.678 & 0.912 \\
XGBoost             & 0.731 & 0.283 & 0.520 & 0.687 & 0.786 & 0.705 & 0.545 & 0.551 & 0.697 & 0.919 \\
LightGBM            & \textbf{0.737} & \textbf{0.298} & \textbf{0.533} & \textbf{0.695} & \textbf{0.794} & \textbf{0.706} & 0.546 & \textbf{0.552} & \textbf{0.699} & 0.902 \\
CatBoost            & 0.733 & 0.276 & 0.524 & 0.689 & 0.785 & 0.702 & \textbf{0.548} & 0.547 & 0.694 & \textbf{0.920} \\
\bottomrule
\end{tabular}%
}
\end{table}

Gradient-boosted tree ensembles (XGBoost, LightGBM, CatBoost) outperform all other methods across all four tasks and all evaluation metrics. LightGBM and CatBoost rank first or second in macro F1 in every task, with XGBoost close behind. The performance gap is substantial: in Stage~1 East, CatBoost achieves an MCC of 0.315 compared to 0.174 for Logistic Regression and 0.150 for MLP, an 80--110\% relative improvement. Linear models (Logistic Regression, SVM) achieve the lowest macro F1 scores in most tasks, indicating that the decision boundaries are inherently nonlinear, interactions among traffic rates, weather, aircraft types, and operational context cannot be captured by linear separating hyperplanes. MLPs are competitive, performing comparably to Decision Trees and occasionally approaching the boosted models, but never achieve the top rank. This is consistent with the tabular-data literature \citep{grinsztajn2022tree, qin2021neural}, which shows that tree-based methods generally outperform neural networks on structured datasets with moderate feature counts and heterogeneous feature types. Random Forest underperforms relative to the boosted ensembles because bagging is less effective than boosting at concentrating capacity on difficult-to-classify samples, which matters for imbalanced multiclass problems. Their strong performance on tabular data, native handling of mixed feature types, built-in support for class weighting and missing values, and efficient training and inference make them the most appropriate model family for this task.

\subsection{Per-Class Metrics}
To provide full transparency on minority-class performance, \Cref{tab:perclass_s1_combined} and \Cref{tab:perclass_s2_combined} report per-class precision, recall, and F1-score for the top-performing models.

\begin{table}[H]
\centering
\caption{Per-class metrics for Stage~1 East Flow (B7/B11/B13) and West Flow (B1/B3/B5).}
\label{tab:perclass_s1_combined}
\small
\resizebox{\textwidth}{!}{%
\begin{tabular}{llcccc|llcccc}
\toprule
\multicolumn{6}{c|}{East Flow} & \multicolumn{6}{c}{West Flow} \\
\cmidrule(r){1-6} \cmidrule(l){7-12}
Model & Class & Precision & Recall & F1 & Support
& Model & Class & Precision & Recall & F1 & Support \\
\midrule
\multirow{3}{*}{LightGBM}
 & B7  & 0.626 & 0.224 & 0.330 & 700
 & \multirow{3}{*}{LightGBM}
 & B1  & 0.667 & 0.036 & 0.069 & 221 \\
 & B11 & 0.904 & 0.984 & 0.942 & 6411
 &  & B3  & 0.886 & 0.960 & 0.922 & 9115 \\
 & B13 & 0.273 & 0.023 & 0.042 & 133
 &  & B5  & 0.653 & 0.429 & 0.518 & 1603 \\
\midrule
\multirow{3}{*}{CatBoost}
 & B7  & 0.664 & 0.220 & 0.331 & 700
 & \multirow{3}{*}{XGBoost}
 & B1  & 1.000 & 0.018 & 0.036 & 221 \\
 & B11 & 0.903 & 0.988 & 0.944 & 6411
 &  & B3  & 0.884 & 0.962 & 0.921 & 9115 \\
 & B13 & 0.132 & 0.011 & 0.028 & 133
 &  & B5  & 0.650 & 0.414 & 0.506 & 1603 \\
\bottomrule
\end{tabular}%
}
\end{table}

\begin{table}[H]
\centering
\caption{Per-class metrics for Stage~2 East Flow (7 classes) and West Flow (6 classes).}
\label{tab:perclass_s2_combined}
\small
\resizebox{\textwidth}{!}{%
\begin{tabular}{llcccc|llcccc}
\toprule
\multicolumn{6}{c|}{East Flow} & \multicolumn{6}{c}{West Flow} \\
\cmidrule(r){1-6} \cmidrule(l){7-12}
Model & Class & Precision & Recall & F1 & Support
& Model & Class & Precision & Recall & F1 & Support \\
\midrule
\multirow{7}{*}{LightGBM}
 & E1         & 0.565 & 0.033 & 0.040 & 21
 & \multirow{6}{*}{CatBoost}
 & E1         & 0.672 & 0.568 & 0.616 & 2905 \\
 & E3         & 1.000 & 0.029 & 0.056 & 35
 &  & E3         & 0.544 & 0.399 & 0.460 & 1494 \\
 & E5         & 0.500 & 0.014 & 0.028 & 70
 &  & E5         & 0.572 & 0.726 & 0.640 & 1044 \\
 & C          & 0.870 & 0.274 & 0.417 & 146
 &  & C          & 0.000 & 0.000 & 0.000 & 11 \\
 & D          & 0.719 & 0.867 & 0.786 & 3718
 &  & D          & 0.670 & 0.886 & 0.763 & 149 \\
 & E11        & 0.583 & 0.011 & 0.021 & 641
 &  & End-Around & 0.775 & 0.852 & 0.812 & 5336 \\
 & End-Around & 0.765 & 0.791 & 0.778 & 2613
 &  &            &       &       &       &      \\
\bottomrule
\end{tabular}%
}
\end{table}

These per-class results reveal the nature and extent of the class imbalance challenge. Majority classes (B11 in S1 East, B3 in S1 West, D and End-Around in S2) are classified with high precision and recall (F1 $>$ 0.77 across all models), demonstrating that the models learn strong decision boundaries for frequently observed operational patterns. Moderate minority classes (B7 in S1 East, B5 in S1 West, C in S2 East, E3/E5 in S2 West) exhibit high precision but low recall, indicating that when the model does predict these classes, it is usually correct, but it misses many true instances by defaulting to the majority class. Extreme minority classes (B13 in S1 East with 1.8\% prevalence; B1 in S1 West with 2.0\%; E1, E3, E5 in S2 East with $<$1\% each; C in S2 West with 0.1\%) are rarely or never predicted by any model. This is a fundamental limitation: with only 21--133 test samples, these classes lack sufficient representation for reliable model training and evaluation. These findings motivate the imbalance mitigation ablation study presented in the following subsection.

\subsection{Imbalance Mitigation Ablation and Class Overlap Analysis}
To systematically address the class imbalance challenge, we evaluated five additional strategies beyond the baseline class-weighted XGBoost: SMOTE oversampling, CatBoost with SqrtBalanced weights, CatBoost with Balanced weights, XGBoost with balanced sample weights, and calibrated one-vs-rest (OVR) classifiers with threshold tuning. \Cref{tab:s1_ablation_combined} and \Cref{tab:s2_ablation_combined} summarize the results.

\begin{table}[H]
\centering
\caption{Stage~1 runway exit classification ablation results for East Flow (B7/B11/B13) and West Flow (B1/B3/B5). Best values in each column are \textbf{bolded}.}
\label{tab:s1_ablation_combined}
\small
\resizebox{\textwidth}{!}{%
\begin{tabular}{lcccc|cccc}
\toprule
& \multicolumn{4}{c|}{East Flow (B7/B11/B13)} & \multicolumn{4}{c}{West Flow (B1/B3/B5)} \\
\cmidrule(r){2-5} \cmidrule(l){6-9}
Method & Accuracy & Balanced Acc. & Macro F1 & MCC & Accuracy & Balanced Acc. & Macro F1 & MCC \\
\midrule
Baseline XGBoost (original) & 0.758 & 0.650 & 0.494 & 0.275 & \textbf{0.862} & 0.462 & 0.482 & 0.425 \\
SMOTE + XGBoost             & \textbf{0.889} & 0.423 & 0.444 & 0.316 & 0.859 & 0.489 & 0.516 & 0.433 \\
CatBoost (SqrtBalanced)     & 0.868 & 0.503 & 0.505 & 0.367 & 0.844 & 0.593 & 0.595 & \textbf{0.478} \\
CatBoost (Balanced)         & 0.579 & \textbf{0.684} & 0.407 & 0.272 & 0.697 & \textbf{0.720} & 0.510 & 0.392 \\
XGBoost (balanced weights)  & 0.848 & 0.518 & 0.491 & 0.369 & 0.708 & 0.712 & 0.516 & 0.402 \\
OVR Calibrated + Threshold  & 0.862 & 0.566 & \textbf{0.541} & \textbf{0.368} & 0.840 & 0.622 & \textbf{0.597} & 0.463 \\
\bottomrule
\end{tabular}%
}
\end{table}

\begin{table}[H]
\centering
\caption{Stage~2 crossing/end-around classification ablation results for East Flow (7 classes) and West Flow (6 classes). Best values in each column are \textbf{bolded}.}
\label{tab:s2_ablation_combined}
\small
\resizebox{\textwidth}{!}{%
\begin{tabular}{lcccc|cccc}
\toprule
& \multicolumn{4}{c|}{East Flow (7 classes)} & \multicolumn{4}{c}{West Flow (6 classes)} \\
\cmidrule(r){2-5} \cmidrule(l){6-9}
Method & Accuracy & Balanced Acc. & Macro F1 & MCC & Accuracy & Balanced Acc. & Macro F1 & MCC \\
\midrule
Baseline XGBoost (original) & \textbf{0.736} & 0.275 & 0.282 & \textbf{0.531} & \textbf{0.705} & 0.568 & 0.546 & 0.551 \\
SMOTE + XGBoost             & 0.732 & 0.287 & 0.302 & 0.524 & 0.702 & 0.582 & \textbf{0.550} & 0.551 \\
CatBoost (SqrtBalanced)     & 0.726 & 0.296 & \textbf{0.309} & 0.516 & 0.698 & 0.601 & \textbf{0.550} & \textbf{0.554} \\
CatBoost (Balanced)         & 0.412 & \textbf{0.354} & 0.236 & 0.299 & 0.643 & \textbf{0.646} & 0.535 & 0.503 \\
XGBoost (balanced weights)  & 0.511 & 0.369 & 0.278 & 0.363 & 0.667 & 0.603 & 0.537 & 0.530 \\
OVR Calibrated (argmax)     & 0.732 & 0.271 & 0.278 & 0.522 & \textbf{0.705} & 0.576 & 0.548 & 0.551 \\
\bottomrule
\end{tabular}%
}
\end{table}

The ablation reveals a fundamental accuracy--fairness tradeoff. Methods that achieve the highest balanced accuracy (CatBoost Balanced, XGBoost balanced weights) do so at significant cost to overall accuracy and MCC, as they over-predict minority classes; conversely, methods optimized for overall accuracy (baseline XGBoost) severely under-predict minority classes. CatBoost with SqrtBalanced weights and OVR Calibrated + Threshold represent a principled middle ground, improving minority-class recognition while maintaining strong majority-class performance. For Stage~1, macro F1 improved from 0.494 to 0.541 (+9.5\%) in the east flow and from 0.482 to 0.597 (+23.9\%) in the west flow.

SMOTE oversampling provides only modest improvements: pure oversampling yields incremental gains in macro F1 (+1--3\%), suggesting that the difficulty lies not in sample quantity but in limited feature-space separability between classes. The relatively consistent macro ROC~AUC ($\approx 0.83$--$0.84$ for Stage~1, $\approx 0.79$--$0.92$ for Stage~2) across all methods supports this interpretation, as the underlying discriminative power of the feature set appears to be the binding constraint.

Cost-sensitive learning is more effective than resampling. CatBoost with SqrtBalanced weights consistently achieved the best or near-best macro F1 and MCC without requiring synthetic sample generation. Unlike SMOTE, cost-sensitive learning does not alter the feature-space geometry; it adjusts the loss function to penalize minority-class errors more heavily, which better suits the overlapping class structure observed in the data.

Stage~2 East exhibits fundamental feature limitations. The extremely low balanced accuracy ($< 0.37$) across all methods for Stage~2 East reflects the fact that the three smallest crossing classes (E1, E3, E5, accounting for only 1.7\% of observations) are nearly indistinguishable in the available feature space, an inherent data limitation rather than a modeling deficiency.

\textbf{Class overlap analysis.}
To investigate why all strategies yield only modest improvements, we draw on recent findings from the imbalanced classification literature. \citet{vuttipittayamongkol2021class} demonstrate through extensive experiments that class overlap has a higher negative impact on classifier performance than class imbalance alone, and that a linearly separable dataset can be perfectly classified by a typical classification algorithm no matter how skewed the class distribution is. Conversely, when class overlap is present, even a balanced dataset can be difficult for a learning task. \citet{tanha2020boosting} further show in a comprehensive review of 14 boosting algorithms on multi-class imbalanced datasets that CatBoost and gradient-boosted methods are superior to other boosting variants, and that the multi-class imbalanced setting introduces additional challenges beyond the binary case because a class may be a minority one when compared to some other classes, but a majority of the rest of them.

\begin{figure}[H]
\centering
    \begin{subfigure}[b]{\textwidth}
        \centering
        \includegraphics[width=0.75\textwidth]{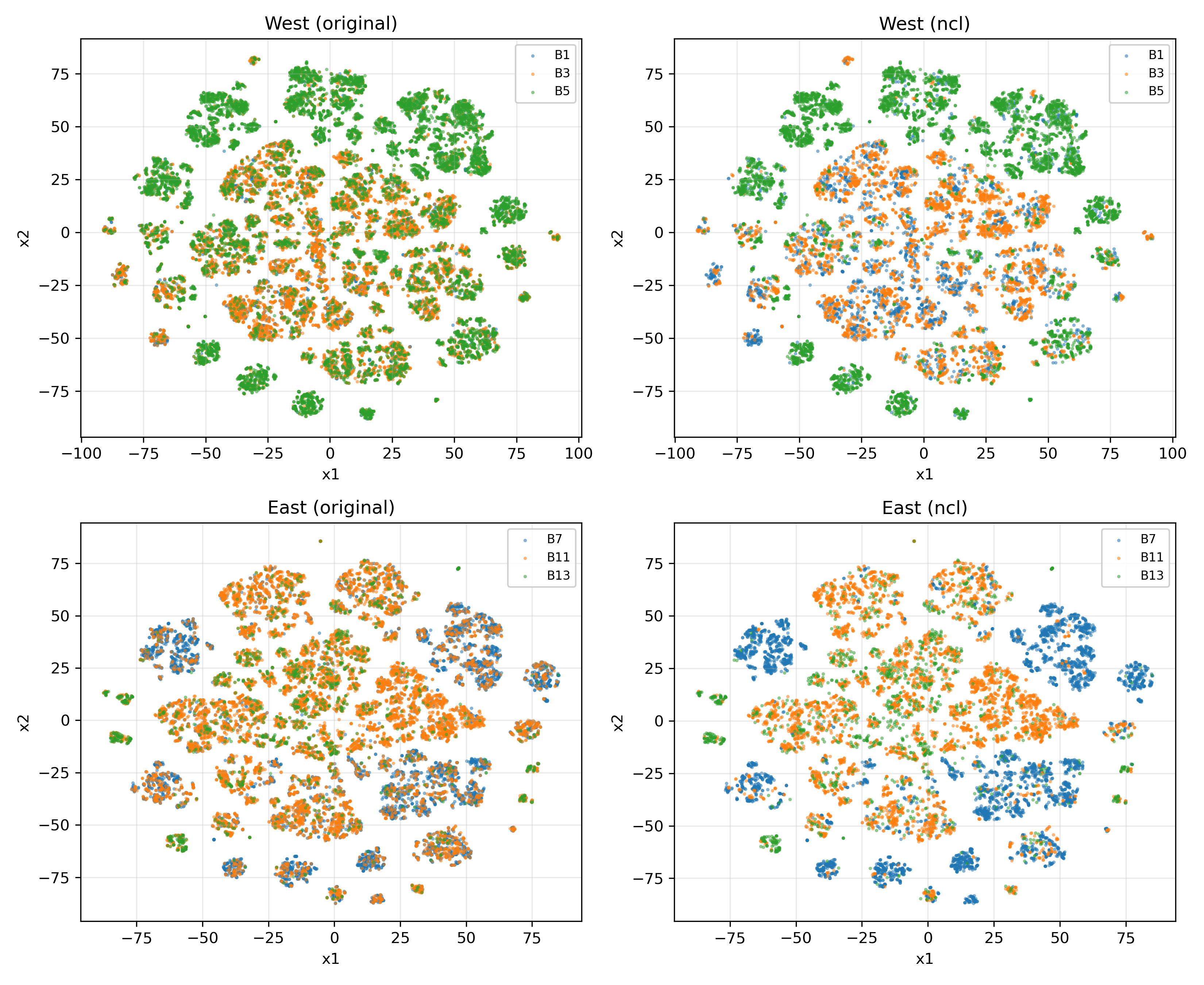}
        \caption{t-SNE \citep{van2008visualizing} visualization of reduced feature-space class overlap for Stage~1 runway exit classification.}
        \label{fig:tsne}
    \end{subfigure}

    \begin{subfigure}[b]{\textwidth}
        \centering
        \includegraphics[width=0.75\textwidth]{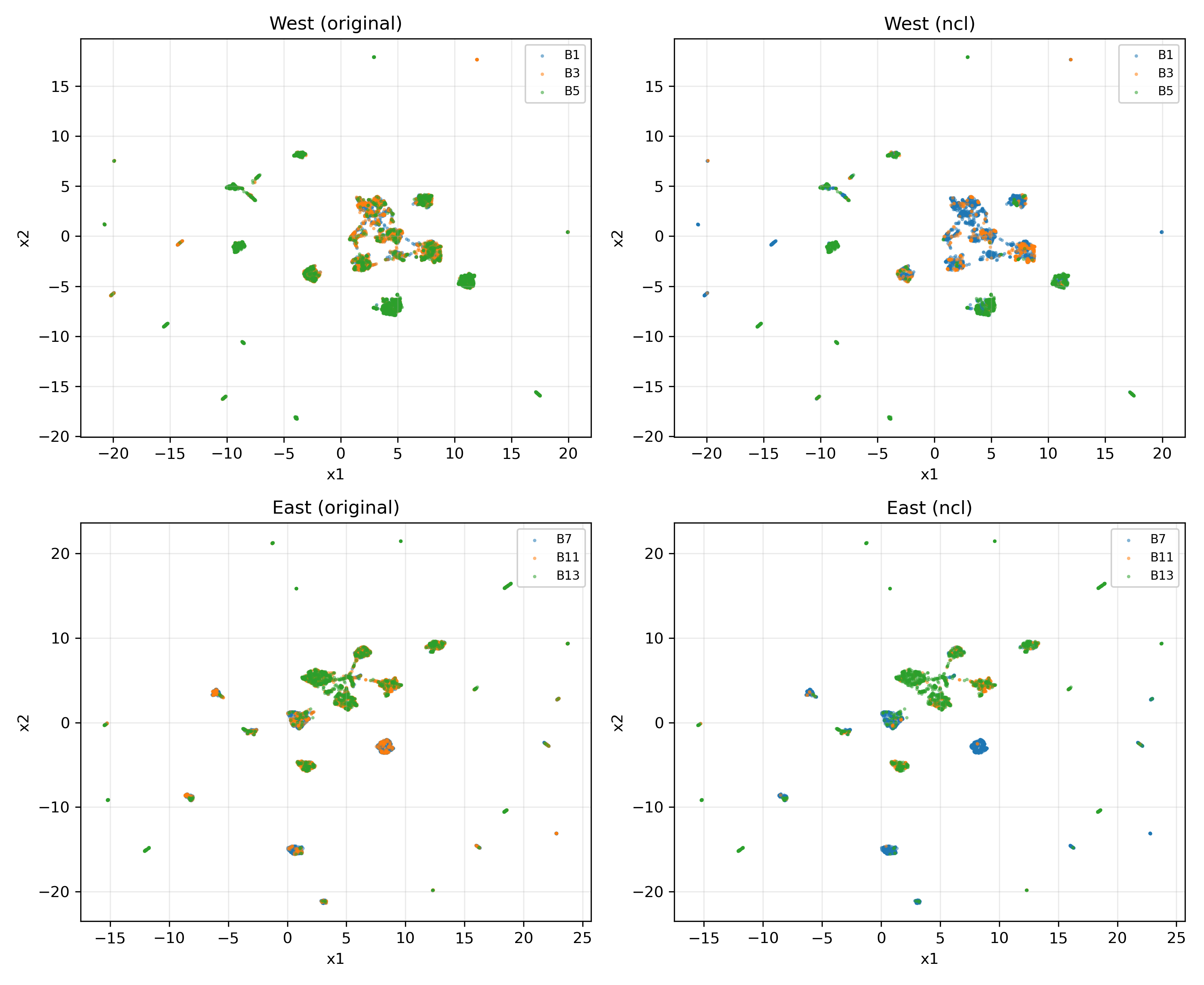}
        \caption{UMAP \citep{mcinnes2018umap} visualization of reduced feature-space class overlap for Stage~1 runway exit classification.}
        \label{fig:umap}
    \end{subfigure}

    \caption{Visualization of the reduced 2D feature-space class overlap for Stage~1 runway exit classification. Left: original data; right: after overlap-aware undersampling. Classes remain heavily intermixed in both cases, confirming that other than class imbalance, class overlap is a primary challenge.}
    \label{fig:class_overlap}
\end{figure}

Motivated by these findings, we conducted a class overlap analysis using multiple dimensionality reduction techniques such as t-SNE \citep{van2008visualizing} and UMAP \citep{mcinnes2018umap} to visualize the feature-space structure of each classification task (\Cref{fig:class_overlap}). The visualizations reveal severe inter-class overlap: minority-class samples (e.g., B7 and B13 in the east flow, B1 in the west flow) are not concentrated in distinct regions of feature space but are instead deeply embedded within the majority-class distribution. This pattern persists across all dimensionality reduction methods and both flow directions, confirming that the difficulty is intrinsic to the feature space, not an artifact of any particular projection. This is consistent with the central finding of \citet{vuttipittayamongkol2021class}: when instances of multiple classes share a common region in the data space, the decision boundary tends to shift towards the negative class leading to misclassification of positive instances near the class boundary.

We further quantified this overlap by computing pairwise Cohen's $d$ effect sizes between minority classes (e.g., B7 vs.\ B13 in Stage~1 East). Most features exhibit small-to-moderate effect sizes ($|d| < 0.3$), indicating that the feature distributions of different exit classes are highly similar. This explains why SMOTE is largely ineffective: as \citet{vuttipittayamongkol2021class} note, traditional resampling methods only aim at getting a more balanced version of the training data and do not factor in the problem of class overlap. The synthetic samples SMOTE creates fall within regions already dominated by the majority class, providing no additional discriminative signal.

We also experimented with overlap-aware undersampling methods (Tomek links, Edited Nearest Neighbors, Condensed Nearest Neighbors, One-Sided Selection, Neighbourhood Cleaning Rule, Instance Hardness Threshold) to selectively remove ambiguous majority-class samples from the overlap region. As shown in the right panels of \Cref{fig:class_overlap}, these methods reduce the total sample count but do not create cleaner class boundaries, further confirming that the overlap is a fundamental property of the feature space rather than a correctable artifact.

Our finding that cost-sensitive boosting (CatBoost SqrtBalanced) outperforms resampling strategies is consistent with both \citet{tanha2020boosting}, who show CatBoost's superiority on multi-class imbalanced datasets, and \citet{vuttipittayamongkol2021class}, who argue that algorithm-level approaches (cost-sensitive learning) are more appropriate than data-level approaches (resampling) when class overlap is the dominant challenge. Cost-sensitive learning adjusts the loss function to penalize minority-class errors more heavily without attempting to alter the feature-space geometry, which is better suited to the overlapping class structure observed in our data.

This finding has an important operational interpretation: under similar traffic, weather, and aircraft conditions, different runway exit and crossing decisions are observed in the historical data. This likely reflects the influence of factors not captured in the ASDE-X and weather features, such as controller preferences, radio instructions \citep{pang2026voice}, pilot familiarity, and real-time conflict resolution, which introduce irreducible stochasticity into the classification task. As noted by \citet{herrema2019machine} and \citet{meijers2019data}, pilot motivation (e.g., gate proximity, airline operational procedures) and ATC instructions via radio are significant drivers of exit choice that cannot be observed from surveillance and weather data alone.

\subsection{Probability Calibration}
\begin{figure}[H]
    \begin{subfigure}[b]{\textwidth}
        \includegraphics[scale=0.5]{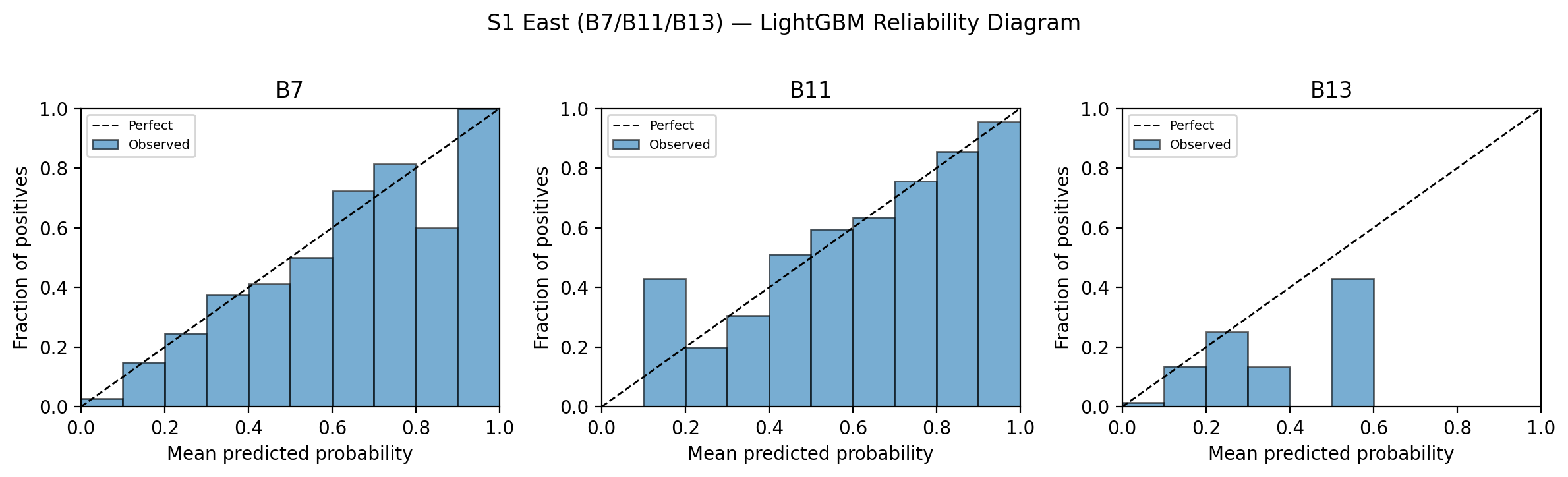}
        \caption{LightGBM S1 East arrival calibration.}
        \label{fig:calis1gbm_east}
    \end{subfigure}

    \begin{subfigure}[b]{\textwidth}
        \includegraphics[scale=0.5]{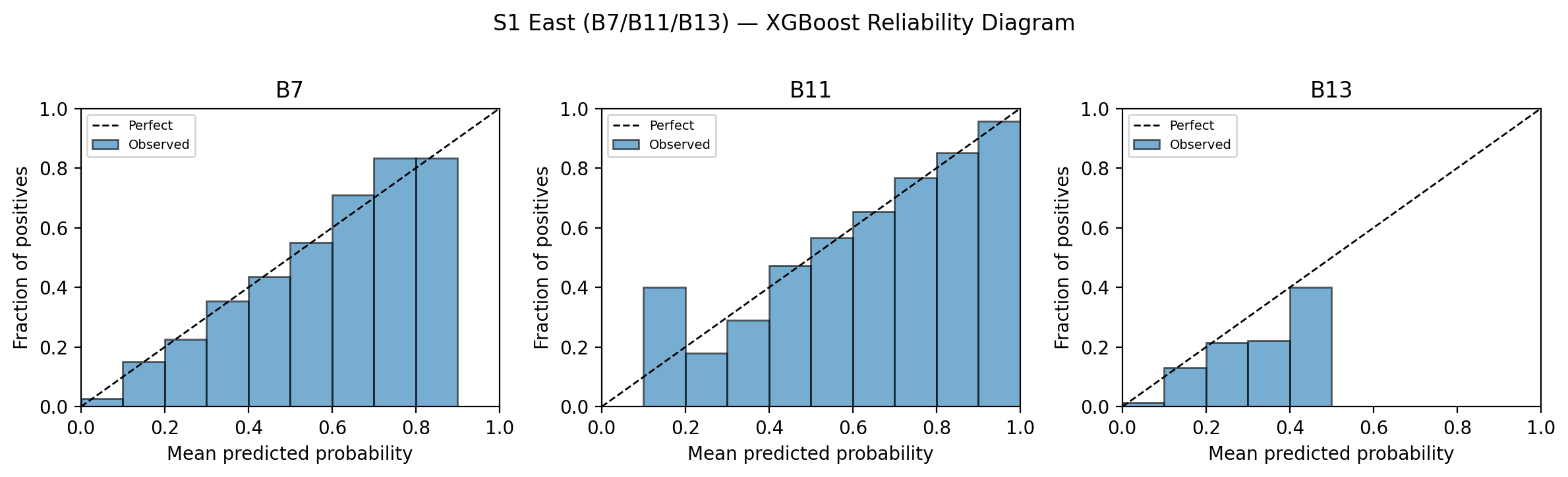}
        \caption{XGBoost S1 East arrival calibration.}
        \label{fig:calis1xgb_east}
    \end{subfigure}

    \begin{subfigure}[b]{\textwidth}
        \includegraphics[scale=0.5]{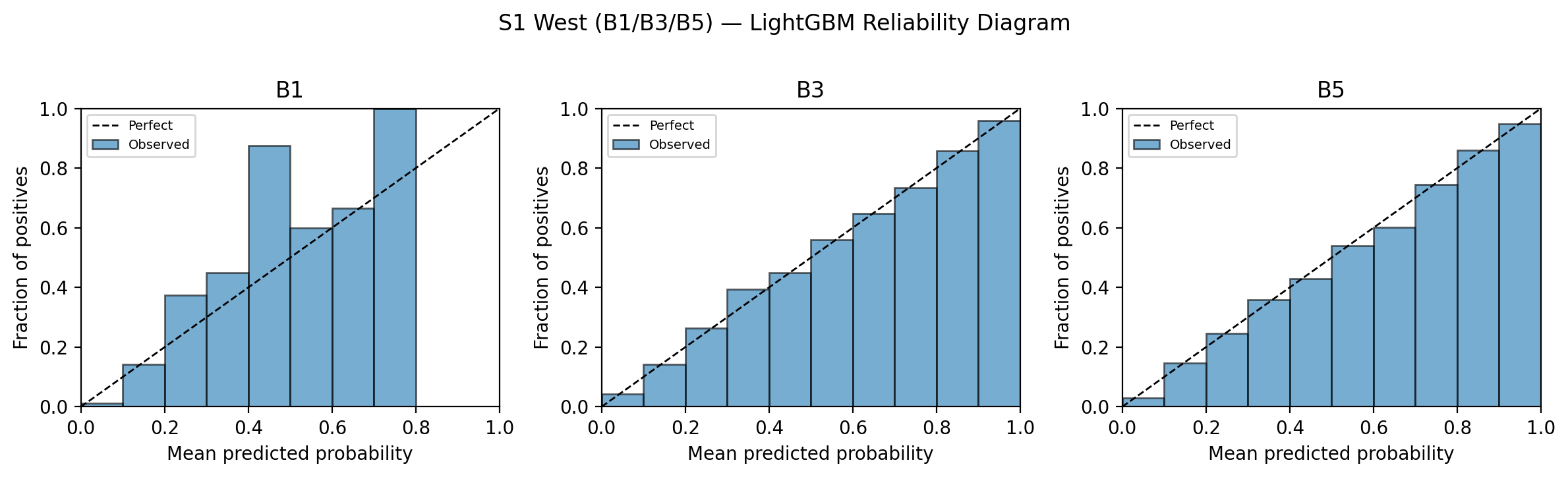}
        \caption{LightGBM S1 West arrival calibration.}
        \label{fig:caligbm_west}
    \end{subfigure}

    \begin{subfigure}[b]{\textwidth}
        \includegraphics[scale=0.5]{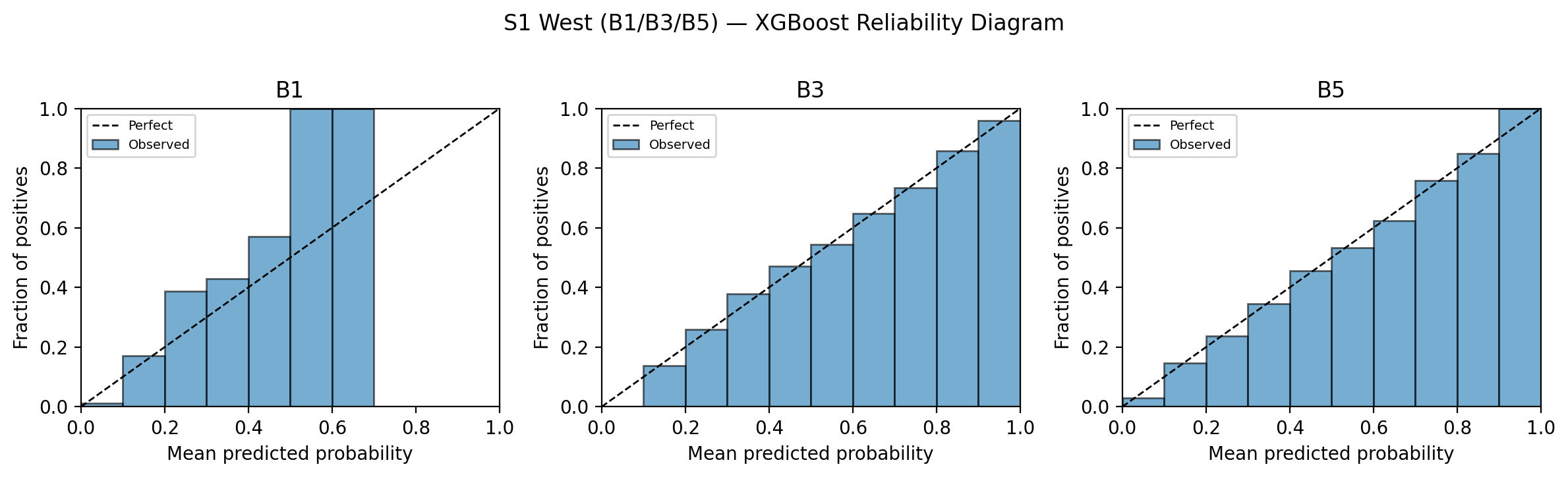}
        \caption{XGBoost S1 West arrival calibration.}
        \label{fig:calixgb_west}
    \end{subfigure}

    \caption{Stage~1 reliability diagrams. Predicted probabilities align closely with observed class frequencies, confirming well-calibrated outputs.}
    \label{fig:cali_s1}
\end{figure}

For a decision-support tool targeting ATCOs, the trustworthiness of predicted probabilities is critical. We report the multiclass Brier score \citep{redelmeier1991assessing} and Expected Calibration Error (ECE) \citep{nixon2019measuring} in \Cref{tab:calibration}. Reliability diagrams are presented in \Cref{fig:cali_s1}.

\begin{table}[H]
\centering
\caption{Probability calibration metrics for all classification tasks. Brier score and ECE are reported as macro-averages across classes. Lower values indicate better calibration.}
\label{tab:calibration}
\small
\begin{tabular}{ll cc}
\toprule
Task & Model & Brier Score & ECE \\
\midrule
\multirow{3}{*}{S1 East (B7/B11/B13)}
 & XGBoost  & 0.0548 & \textbf{0.0036} \\
 & LightGBM & 0.0552 & 0.0058 \\
 & CatBoost & \textbf{0.0545} & 0.0065 \\
\midrule
\multirow{3}{*}{S1 West (B1/B3/B5)}
 & XGBoost  & 0.0667 & 0.0054 \\
 & LightGBM & \textbf{0.0664} & \textbf{0.0037} \\
 & CatBoost & 0.0673 & 0.0075 \\
\midrule
\multirow{3}{*}{S2 East (7 classes)}
 & XGBoost  & 0.0555 & 0.0057 \\
 & LightGBM & \textbf{0.0547} & \textbf{0.0057} \\
 & CatBoost & 0.0559 & 0.0058 \\
\midrule
\multirow{3}{*}{S2 West (6 classes)}
 & XGBoost  & 0.0671 & \textbf{0.0053} \\
 & LightGBM & \textbf{0.0670} & 0.0065 \\
 & CatBoost & 0.0670 & 0.0054 \\
\bottomrule
\end{tabular}
\end{table}

The calibration results are encouraging for the intended decision-support application. The multiclass Brier scores range from 0.054 to 0.067, indicating that the predicted probability distributions are close to the true class outcomes. The macro-averaged ECE is below 0.008 for every model-task combination, with most values below 0.006. This means the predicted probabilities deviate from true frequencies by less than 1 percentage point on average, which is an excellent level of calibration for operational decision support.

All three gradient-boosted models produce comparably well-calibrated probabilities, with no single model consistently dominating. This is consistent with the known property of gradient-boosted trees: because they optimize log-loss (cross-entropy) by default, their softmax outputs are inherently well-calibrated without requiring post-hoc calibration (e.g., Platt scaling or isotonic regression). The reliability diagrams in \Cref{fig:cali_s1} confirm that predicted probabilities align closely with observed class frequencies across the full range of confidence levels.

These results confirm that the predicted probabilities can be meaningfully presented to ATCOs as calibrated confidence estimates rather than hard class predictions alone. In practice, a controller presented with a 70\% probability of crossing at taxiway D can trust that, historically, approximately 70\% of flights with similar characteristics and conditions did cross at D.

\subsection{Summary}
Gradient-boosted tree ensembles deliver consistent gains over all other methods across both stages, confirming the suitability of tree-based models for this structured classification task. Stage~I achieves high accuracy (0.86--0.89) but uneven macro performance due to exit prevalence; Stage~II is harder, with performance governed by geometry and short-horizon runway usage. The nine-method comparison demonstrates that the performance gap between boosted trees and alternative methods (linear models, neural networks) is substantial and consistent across all tasks. The imbalance ablation study shows that cost-sensitive learning (CatBoost SqrtBalanced) provides the best practical balance between minority-class recognition and overall reliability, improving macro F1 by up to 23.9\% over the baseline. The class overlap analysis identifies feature-space inseparability as the fundamental bottleneck limiting minority-class classification, a finding that explains both the modest gains from SMOTE and the inherent ceiling on balanced accuracy. The calibration analysis confirms that the probabilistic outputs are well-calibrated (ECE $<$ 0.008) and can be trusted for operational decision support. SHAP-based explanations align with controller heuristics (i.e., approach/kinematics and wind for exit; departure/cross rates, realized exit, and ramp for routing) supporting transparent use as a decision aid.

Based on the totality of these analyses, we recommend CatBoost with SqrtBalanced class weights as the primary model, as it provides the most favorable balance between minority-class recognition (macro F1) and overall predictive reliability (MCC) across all four tasks, while also producing well-calibrated probabilistic outputs. Taken together, these properties (the staged factorization that mirrors controller workflow, the calibrated probabilities suitable for confidence-aware display, and the SHAP-based attributions tied to operationally meaningful covariates) make the framework well suited for deployment as an ATCO decision-support tool that augments situational awareness without removing the human controller from the decision loop.

\section{Conclusions\label{sec: conclusion}}
In this paper, we presented a two-stage, data-driven framework for classifying (i) the runway exit selected by arriving aircraft and (ii) the subsequent routing decision between crossing the departure runway at a specific point or using the end-around taxiway on the KATL north complex. By aligning the modeling sequence with actual controller workflow, the framework yields probabilistic, interpretable outputs that are intended to be surfaced as controller-support cues at the tower. We emphasize that the framework is positioned as an assistive decision aid that enhances ATCO situational awareness while preserving human judgment and operational responsibility, not as an autonomous replacement for controller decision making.

We conducted a comprehensive model comparison spanning nine classification methods (Logistic Regression, SVM, KNN, MLP, Decision Tree, Random Forest, XGBoost, LightGBM, CatBoost), confirming that gradient-boosted tree ensembles consistently outperform all alternatives across both stages and all evaluation metrics \citep{grinsztajn2022tree}. In Stage~I, accuracies near 0.86--0.89 coexisted with macro-F1 in the 0.40--0.50 range; in Stage~II, accuracies around 0.70--0.74 reflected the larger and more imbalanced label space, with macro-F1 spanning roughly 0.28--0.55. Probability calibration analysis demonstrated that all gradient-boosted models produce well-calibrated probabilistic outputs (ECE $<$ 0.008 across all tasks), confirming their suitability for decision support where trustworthy confidence estimates are paramount.

To address the class imbalance challenge, we conducted a systematic ablation study evaluating SMOTE, cost-sensitive learning (CatBoost SqrtBalanced/Balanced, XGBoost balanced weights), and calibrated one-vs-rest classifiers with threshold tuning. CatBoost with SqrtBalanced weights achieved the best overall balance between minority-class recognition (macro F1) and predictive reliability (MCC). A class overlap analysis using t-SNE \citep{van2008visualizing} and UMAP \citep{mcinnes2018umap} revealed that feature-space inseparability (not merely class imbalance) is the fundamental bottleneck: minority-class samples are deeply embedded within the majority-class distribution, and overlap-aware undersampling methods fail to create cleaner boundaries. This finding has an important operational interpretation: under similar traffic, weather, and aircraft conditions, different routing decisions are observed, likely reflecting unrecorded factors such as controller preferences, radio communications, and real-time conflict resolution.

SHAP analyses provided stable, operationally meaningful explanations: \emph{approach speed} dominated the exit choice in Stage~I, while Stage~II decisions were influenced by \emph{departure rate}, recent crossing activity, and ramp--exit geometry. These patterns agree with and extend prior exit-prediction findings \citep{martinez2018boosted, herrema2019machine, woo2022runway, meijers2019data}.

Several factors constrain performance and generality. Class overlap concentrates accuracy in majority options and limits minority recall despite weighting, resampling, and careful validation. The models are tuned to KATL north-side geometry and procedures, so transfer to other complexes will require retraining with topology-aware features. We also intentionally restricted inputs to information available by the decision time to avoid leakage, which caps attainable recall for rare actions that depend on emerging queue gaps or last-moment clearances.

\subsection{Limitations and Future work}
The class overlap analysis and per-class evaluation identify several concrete directions for future research.

On the feature side, the class overlap analysis shows that the available ASDE-X and weather features do not contain sufficient discriminative signal to distinguish all routing classes, particularly for rare decisions. Incorporating real-time departure queue length and waiting times at each crossing point would capture a factor that directly enters ATCO crossing versus end-around decisions; this requires integrating SWIM TBFM data or reconstructing departure queue dynamics from ASDE-X surface tracks, a non-trivial data engineering effort beyond the present scope. Modeling explicit pairwise conflict probabilities at intersections \citep{sui2023conflict} would align the feature set more closely with controller cognitive processes; our departure and crossing rate features at multiple time horizons are aggregate proxies, and explicit conflict features could improve classification of rare crossing decisions. ATC instructions via radio are a significant unrecorded factor \citep{herrema2019machine, pang2026voice}, and recent advances in speech-to-text for aviation communications offer a path to extract intent features from voice data that could reduce the irreducible stochasticity identified in the class overlap analysis. Additional gap-availability proxies (i.e., time since the last crossing event and time to the next scheduled departure) could also sharpen Stage~II classification by capturing the departure-stream gap structure that ATCOs exploit when authorizing crossings.

On the methodological side, a hierarchical Stage~II classifier that first predicts the binary cross-versus-end-around decision before refining the specific crossing point could improve practical utility by separating the safety-critical routing decision from the within-family crossing-point assignment. Regime-specific decision thresholds (e.g., tuned separately for high versus low departure rate) could suppress false crossing predictions under heavy departure demand while surfacing crossing opportunities when confidence is high. Integrating efficiency labels, linking each routing decision to downstream departure delays, additional taxiing time, and fuel consumption, would move the framework from prediction toward prescription, though such analysis requires either a validated airport simulation calibrated to KATL or comprehensive cost data linking routing decisions to delay propagation.

The models are currently tuned to KATL north-side geometry and procedures; transferring them to other airport complexes will require retraining with topology-aware features. Integrating the two-stage classifier with existing surface-operations tools (e.g., TFDM, SARDA-style advisories) and evaluating controller acceptance in human-in-the-loop studies are important next steps toward operational deployment.

\subsection{Operational Integration within Human--Machine Collaboration and ATCO Training}
We emphasize that the proposed framework is intended as a \emph{controller-support tool}, not as an autonomous decision maker. The classifier surfaces a calibrated, ranked set of likely exit and crossing alternatives together with the associated probabilities and SHAP-based attributions, while the ATCO retains full authority over the final clearance and operational responsibility for separation and safety. This positioning is consistent with the human-in-command philosophy adopted in current ATM modernization roadmaps and with the broader literature on explainable AI for safety-critical systems \citep{xie2021explanation, sutthithatip2022explainable, wang2022artificial, NASA_ARMD_2023}. In an integrated tower environment, the Stage~I and Stage~II outputs can be rendered as advisory annotations on the surface-management display, e.g., highlighting the most likely exit and signalling cross versus end-around with a confidence band, so that controllers can quickly cross-check the system's expectation against the unfolding traffic picture and intervene whenever real-time information (radio readbacks, observed conflicts, supervisor coordination) calls for a different choice.

Several aspects of the framework are deliberately designed to support, rather than displace, the controller's cognitive workflow. First, the staged factorization mirrors the natural temporal ordering of ATCO decisions (exit choice at touchdown, routing choice at exit), which keeps the advisory aligned with the moment the controller is actually committing to each decision and limits the amount of additional information presented at any one time. Second, the well-calibrated probabilistic outputs (ECE $<$~0.008) allow controllers to weigh the advisory in proportion to its confidence rather than treating it as a binary prescription, reducing the risk of automation bias when the prediction is uncertain. Third, the SHAP-based attributions tied to operationally meaningful covariates (approach speed, ramp destination, departure and crossing rates) give controllers a concise, inspectable rationale for each suggestion, which prior human-factors work in ATM identifies as a key enabler of appropriate reliance on AI-based advisories \citep{xie2021explanation, sutthithatip2022explainable}. Finally, the rare-event limitations documented in the class overlap analysis are reported transparently, so that downstream user-interface design can flag low-confidence regimes (e.g., extreme minority crossing classes) where the advisory should be suppressed or shown with explicit uncertainty.

The deployment of such predictive systems also has implications for ATCO training and decision-support workflows. We agree with the reviewers that fully realizing the benefits of human--machine collaboration requires complementary updates to the ATCO training paradigm. Predictive advisories shift part of the controller's cognitive load from forecasting future surface state to assessing, calibrating, and selectively overriding model outputs, which is a qualitatively different skill set than that emphasized in current training syllabi. We anticipate that future training programs will need to (i) familiarize trainees with the meaning of calibrated probabilities and reliability information so that confidence cues can be interpreted correctly under workload; (ii) develop heuristics for recognizing situations in which the model is likely to be unreliable, in particular the rare-routing and high-overlap regimes identified in our class overlap analysis, so that controllers know when to discount the advisory; (iii) provide structured exposure to model failure modes through simulator scenarios, including adversarial or off-nominal traffic patterns, so that trust in the system is calibrated rather than uniformly high or uniformly low; and (iv) reinforce that final responsibility for clearances and separation remains with the human controller, in line with the human-in-command framing adopted throughout this paper. Embedding the predictive framework in tower simulators alongside such curriculum updates would also generate the human-in-the-loop performance data needed to refine both the advisory and the operating procedures around it. We view the co-design of the predictive tool and its supporting training paradigm as a necessary precondition for moving from research prototype to operational decision support, and as an important direction for follow-on work in partnership with operating facilities.

The staged classification framework, with its calibrated probabilities and explainable attributions, is designed to complement existing optimization and rule-based surface tools. Deployed as a controller-support aid rather than an automated directive, and accompanied by an appropriately updated training paradigm, it can improve predictability of exits and crossings and help controllers manage workload as traffic volumes grow while preserving human judgment and operational responsibility \citep{NASA_ARMD_2023}.

\section*{Acknowledgment}
This work was supported by the National Aeronautics and Space Administration (NASA) University Leadership Initiative (ULI) program under project “Autonomous Aerial Cargo Operations at Scale”, via grant No. 80NSSC21M071 to the University of Texas at Austin. Any opinions, findings, conclusions, or recommendations expressed in this material are those of the authors and do not necessarily reflect the views of the project sponsor.

\bibliography{ref}

\end{document}